\documentclass[sigconf,screen]{acmart}
\usepackage{xspace,balance,tabularx,multirow}
\usepackage{flushend}
\usepackage{tikz}
\usepackage{pgfplots}
\pgfplotsset{compat=1.16}
\usetikzlibrary{patterns}
\usepackage{subfig}
\usepackage[ruled, vlined, linesnumbered]{algorithm2e}
\usepackage{xcolor}
\usepackage{colortbl}
\usepackage{bbold}
\SetKwComment{Comment}{$\triangleright$\ }{}
\usepackage{enumitem}
\usepackage{tablefootnote}
\usepackage{upgreek,textgreek}
\usepackage{pifont}%
\usepackage[noabbrev]{cleveref}
\usepackage{titlecaps}
\usepackage{lipsum}

\pgfplotsset{every tick label/.append style={font=\tiny}}

\newlength{\starsize}
\newlength{\starspread}
\tikzset{starsize/.code={\setlength{\starsize}{#1}},
         starspread/.code={\setlength{\starspread}{#1}}}
\tikzset{starsize=1mm,
         starspread=3mm}
\pgfdeclarepatternformonly[\starspread,\starsize]%
  {my fivepointed stars}%
  {\pgfpointorigin}%
  {\pgfqpoint{\starspread}{\starspread}}%
  {\pgfqpoint{\starspread}{\starspread}}%
  {%
   \pgftransformshift{\pgfqpoint{\starsize}{\starsize}}
   \pgfpathmoveto{\pgfqpointpolar{18}{\starsize}}
   \pgfpathlineto{\pgfqpointpolar{162}{\starsize}}
   \pgfpathlineto{\pgfqpointpolar{306}{\starsize}}
   \pgfpathlineto{\pgfqpointpolar{90}{\starsize}}
   \pgfpathlineto{\pgfqpointpolar{234}{\starsize}}
   \pgfpathclose%
   \pgfusepath{fill}
  }

\makeatletter
\newcommand*\bigcdot{\mathpalette\bigcdot@{.5}}
\newcommand*\bigcdot@[2]{\mathbin{\vcenter{\hbox{\scalebox{#2}{$\m@th#1\bullet$}}}}}
\makeatother

\newcommand{\stitle}[1]{\vspace*{0.5em}\noindent{\underline{\bf #1.\/}}}

\def\header{\vspace{1mm} \noindent}

\DeclareMathOperator{\Tr}{\textsf{tr}}

\newcommand{\V}{\mathcal{V}\xspace}
\newcommand{\G}{\mathcal{G}\xspace}
\newcommand{\N}{\mathcal{N}\xspace}

\newcommand{\EDG}{\mathcal{E}\xspace}

\newcommand{\WM}{\mathbf{W}\xspace}
\newcommand{\AM}{\mathbf{A}\xspace}
\newcommand{\NAM}{\hat{\mathbf{A}}\xspace}
\newcommand{\DM}{\mathbf{D}\xspace}
\newcommand{\IM}{\mathbf{I}\xspace}
\newcommand{\SM}{\mathbf{S}\xspace}
\newcommand{\MM}{\mathbf{M}\xspace}
\newcommand{\PM}{\mathbf{P}\xspace}

\newcommand{\KM}{\boldsymbol{K}\xspace}

\newcommand{\XM}{\mathbf{X}\xspace}
\newcommand{\LM}{\mathbf{L}\xspace}
\newcommand{\UM}{\mathbf{U}\xspace}
\newcommand{\VM}{\mathbf{V}\xspace}
\newcommand{\HM}{\mathbf{H}\xspace}
\newcommand{\ZM}{\mathbf{Z}\xspace}

\newcommand{\FM}{\mathbf{F}\xspace}

\newcommand{\QM}{\mathbf{Q}\xspace}

\newcommand{\zm}{\mathbf{z}\xspace}

\newcommand{\eat}[1]{}

\newcommand{\algo}{\texttt{DGF}\xspace}

\def\addlegendimage{\csname pgfplots@addlegendimage\endcsname}

\makeatletter
\newcommand\footnoteref[1]{\protected@xdef\@thefnmark{\ref{#1}}\@footnotemark}
\makeatother

\let\oldnl\nl%
\newcommand{\nonl}{\renewcommand{\nl}{\let\nl\oldnl}}%

\definecolor{myred}{HTML}{fd7f6f}
\definecolor{myred_new}{HTML}{D8D8D8}
\definecolor{myred_new2}{HTML}{D7191C}
\definecolor{myblue}{HTML}{7eb0d5}
\definecolor{mygreen}{HTML}{b2e061}
\definecolor{mypurple}{HTML}{bd7ebe}
\definecolor{myorange}{HTML}{ffb55a}
\definecolor{myyellow}{HTML}{ffee65}
\definecolor{mypurple2}{HTML}{beb9db}
\definecolor{mypink}{HTML}{fdcce5}
\definecolor{mycyan}{HTML}{8bd3c7}

\definecolor{myblue2}{HTML}{115f9a}
\definecolor{myred2}{HTML}{c23728}

\definecolor{mygreen3}{HTML}{70ad47}
\definecolor{myblue3}{HTML}{5959a4}
\definecolor{myred3}{HTML}{b05159}

\AtBeginDocument{%
  \providecommand\BibTeX{{%
    \normalfont B\kern-0.5em{\scshape i\kern-0.25em b}\kern-0.8em\TeX}}}

\setcopyright{acmcopyright}
\copyrightyear{2018}
\acmYear{2018}
\acmDOI{XXXXXXX.XXXXXXX}

\acmPrice{15.00}
\acmISBN{978-1-4503-XXXX-X/18/06}

\acmSubmissionID{472}

\begin{document}

\title{Cross-Contrastive Clustering for Multimodal Attributed Graphs with Dual Graph Filtering}
\subtitle{Technical Report}
\author{Haoran Zheng}
\affiliation{%
  \institution{Hong Kong Baptist University}
  \country{Hong Kong SAR, China}
}
\email{cshrzheng@comp.hkbu.edu.hk}
\orcid{0009-0002-4769-3716}

\author{Renchi Yang}
\authornote{Corresponding Author}
\affiliation{%
  \institution{Hong Kong Baptist University}
  \country{Hong Kong SAR, China}
}
\email{renchi@hkbu.edu.hk}
\orcid{0000-0002-7284-3096}

\author{Hongtao Wang}
\affiliation{%
  \institution{Hong Kong Baptist University}
  \country{Hong Kong SAR, China}
}
\email{cshtwang@comp.hkbu.edu.hk}
\orcid{0009-0002-1279-4357}

\author{Jianliang Xu}
\affiliation{%
  \institution{Hong Kong Baptist University}
  \country{Hong Kong SAR, China}
}
\email{xujl@comp.hkbu.edu.hk}
\orcid{0000-0001-9404-5848}

\settopmatter{printfolios=true}

\renewcommand{\shortauthors}{Zheng et al.}

\begin{abstract}
{\em Multimodal Attributed Graphs} (MMAGs) are an expressive data model for representing the complex interconnections among entities that associate attributes from multiple data modalities (text, images, etc.). Clustering over such data finds numerous practical applications in real scenarios, including social community detection, medical data analytics, etc.
However, as revealed by our empirical studies, existing multi-view clustering solutions largely rely on the high correlation between attributes across various views and overlook the unique characteristics (e.g., low modality-wise correlation and intense feature-wise noise) of multimodal attributes output by large pre-trained language and vision models in MMAGs, leading to suboptimal clustering performance.

Inspired by foregoing empirical observations and our theoretical analyses with graph signal processing, we propose the {\em Dual Graph Filtering} (\algo{}) scheme, which innovatively incorporates a feature-wise denoising component into node representation learning, thereby effectively overcoming the limitations of traditional graph filters adopted in the extant multi-view graph clustering approaches.
On top of that, \algo{} includes a tri-cross contrastive training strategy that employs instance-level contrastive learning across modalities, neighborhoods, and communities for learning robust and discriminative node representations.
Our comprehensive experiments on eight benchmark MMAG datasets exhibit that \algo{} is able to outperform a wide range of state-of-the-art baselines consistently and significantly in terms of clustering quality measured against ground-truth labels.
\end{abstract}



\maketitle

\section{Introduction}

Graphs are ubiquitous data structures used for modeling the interplay among real-world entities. In practical scenarios, they are often endowed with information from heterogeneous modalities, e.g., textual, visual, and acoustic ones, which is termed as the {\em multimodal attributed graph} (hereinafter MMAG)~\cite{yoon2023multimodal,yan2024graph}. For instance, web pages in hyperlink graphs (e.g., Wikipedia~\cite{burns2023suite}) usually embody not only text but also pictures; in social media like TikTok and YouTube~\cite{vedula2017multimodal}, videos are associated with textual descriptions, visual appearance and acoustic signal; In e-commerce platforms such as Amazon~\cite{li2020adversarial}, products like cloths typically feature detailed descriptions, example photos, and demonstration videos. The multimodal attributes accommodate rich information about entities and compensate for noisy/incomplete relationships in MMAGs. However, such useful information is largely under-exploited or inadequately utilized in graph analytics due to technical barriers and unique challenges therein, particularly in clustering tasks, which seek to partition the entities in MMAGs into disjoint groups~\cite{wang2025deep}.

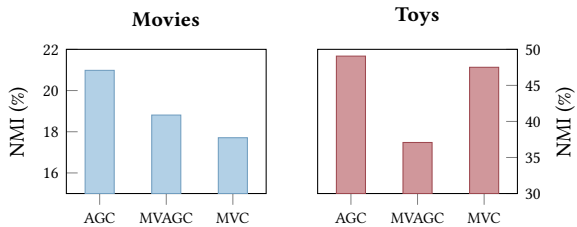
\begin{figure}[!t]
\centering
\begin{small}
\begin{tikzpicture}
\begin{axis}[
    title={\textbf{Movies}},
    ybar,
    width=0.5\columnwidth,
    height=3.5cm,
    bar width=11pt,
    enlarge x limits=0.25,
    ylabel={{NMI (\%)}},
    ymin=15, ymax=22,
    axis lines=box,
    yticklabel style={font=\scriptsize},
    ytick pos=left,
    symbolic x coords={AGC, MVAGC, MVC},
    xtick=data,
    xticklabel style={font=\scriptsize, anchor=north},
    xtick pos=bottom,
]
\addplot[fill=myblue!60, draw=myblue!90!black] coordinates {(AGC, 20.98) (MVAGC, 18.81) (MVC, 17.71)};
\end{axis}
\end{tikzpicture}\hspace{6mm}
\begin{tikzpicture}
\begin{axis}[
    title={\textbf{Toys}},
    ybar,
    width=0.5\columnwidth,
    height=3.5cm,
    bar width=11pt,
    enlarge x limits=0.25,
    ylabel={{NMI (\%)}},
    ymin=30, ymax=50,
    axis lines=box,
    yticklabel pos=right,
    yticklabel style={font=\scriptsize},
    ytick pos=right,
    symbolic x coords={AGC, MVAGC, MVC},
    xtick=data,
    xticklabel style={font=\scriptsize, anchor=north},
    xtick pos=bottom,
]
\addplot[fill=myred3!60, draw=myred3!90!black] coordinates {(AGC, 49.06) (MVAGC, 37.08) (MVC, 47.51)};
\end{axis}
\end{tikzpicture}
\end{small}
\vspace{-3ex}
\caption{Performance comparison of the state-of-the-art AGC, MVAGC, and MVC methods 
(i.e., \texttt{S3GC}~\cite{devvrit2022s3gc}, \texttt{LMGEC}~\cite{fettal2023simultaneous}, \texttt{EMVGC-LG}~\cite{wen2023efficient}) on real MMAGs \textit{Movies}~\cite{yan2024graph} and \textit{Toys}~\cite{yan2024graph}.
}
\label{fig:sota_fail}
\vspace{-3ex}
\end{figure}

A simple treatment for clustering over MMAGs is to directly adopt the well-established {\em multi-view clustering} (MVC)~\cite{zhou2024survey} techniques, which treat the data from diverse sources as multiple view attributes and aim to achieve consensus clustering among all the views. However, this methodology fails to capitalize on the accompanied graph structures, which inspires subsequent research on {\em multi-view attributed graph clustering} (MVAGC)~\cite{li2025clustering}.
Although MVAGC can be readily applied on MMAGs for cluster tasks, the majority of existing approaches are deficient since they are designed for multi-view data wherein the attribute data in distinct views is highly correlated/resembling, as pointed out by our preliminary experiments in \S~\ref{sec:MMAG-char}. 
The reason is that most views are extracted from similar sources, such as color histograms, texture descriptors, and SIFT features from images, or RGB images, depth maps, and infrared images from multiple sensors~\cite{zhao2017multi}, etc. Even more, some of them are artificially generated/constructed from the attributes of base views.
Particularly, as revealed in Fig.~\ref{fig:sota_fail}, compared to a recent model~\texttt{S3GC}~\cite{devvrit2022s3gc} for uni-modal attributed graph clustering (AGC)~\cite{lin2025spectral,li2024versatile}, the state of the arts for MVC and MVAGC, \texttt{EMVGC-LG}~\cite{wen2023efficient} and \texttt{LMGEC}~\cite{fettal2023simultaneous}, exhibit conspicuously inferior clustering performance on real MMAGs \textit{Movies}~\cite{yan2024graph} and \textit{Toys}~\cite{yan2024graph} that encompass uncorrelated attribute data across multiple modalities (see \S~\ref{sec:MMAG-char}).

In light of the significant advancements in multimodal technology (e.g., pre-trained language and vision models)~\cite{yin2024survey}, multimodal attribute data including text, images, and videos in MMAGs can be encoded as high-quality embeddings, fostering the emergence of {\em multimodal graph learning} (MGL)~\cite{peng2024learning,ektefaie2023multimodal}. However, MGL is in its infancy and most works focus on developing benchmarks~\cite{yan2024graph, zhu2024multimodal,zhu2024benchmarking}, synthetic tasks~\cite{shen2024sg, jin2024instructg2i}, and supervised problems including node classification and link prediction through a direct adoption of existing {\em graph neural networks} (GNNs)~\cite{kipf2017semisupervised,veličković2018graph,gasteiger2019combining}.
Very recently, \citet{wang2025deep} combines the Transformer and auto-encoder for MMAG clustering. Despite the advances made, akin to MGL works, they simply adapt the extant model architectures over MMAGs and fall short of algorithmic and model designs dedicated to coping with the unique characteristics of real MMAGs.
Particularly, as uncovered by our empirical studies on MMAGs in \S~\ref{sec:MMAG-char}, aside from the large gaps of multimodal attributes in semantic spaces (i.e., low modality-wise correlation) as aforementioned, multimodal attributes of MMAGs embody considerable noise/outliers at both node and feature domains, rendering existing models suboptimal.

To bridge this gap, this paper presents a novel and effective approach \algo{} that is specially designed for MMAG clustering through two major technical contributions. 
Firstly, theoretically grounded by our analysis pinpointing that prior GNNs essentially apply graph filtering on attributes for node-wise denoising from the perspective of {\em graph signal processing}~\cite{dong2016learning,isufi2024graph}, \algo{} additionally incorporates a {\em feature-domain denoising} component and formulates the representation learning over MMAGs as a {\em dual graph filtering}. The developed truncated dual graph filter can be calculated in almost linear time, and is theoretically proved to be an approximate ideal low-pass filter for multimodal attributes at both node and feature levels.
Furthermore, we devise a {\em tri-cross contrastive training} scheme, aiming at learning robust and discriminative representations for nodes by harnessing the power of instance-level contrastive learning across modalities, topological neighborhoods, and communities. More specifically, a cross-modality contrastive loss that learns a joint embedding space for multimodal attributes is included, thereby mitigating their inconsistency and low correlation in the semantic spaces.
On top of that, to fully capture the relations between nodes in graphs while excluding inaccuracies introduced by spurious links, we design an attribute-aware random walk sampling strategy to realize the cross-neighborhood contrast. 
In the meanwhile, \algo{} implements the cross-community contrast for nodes with positive hard samples~\cite{zhang2024deep}, so as to strengthen the robustness of the clustering results.
Our extensive experiments comparing \algo{} against recent AGC, MVC, and MVAGC baselines on eight real MMAG datasets showcase that our proposed model is consistently and significantly superior to all competitors in terms of five popular metrics for measuring clustering quality.

\section{Related Work}

\subsection{Graph-based Multi-view Clustering}
An important line in {\em multi-view clustering} (MVC) research operates on constructing graphs from feature data to capture its underlying structure. The central idea is to learn a consensus affinity graph from multiple derived graphs, on which a final clustering is performed. The quality of this consensus graph is paramount. To this end, various strategies have been proposed. Anchor-based methods are commonly adopted for scalability, which reduce time costs by modeling relations between the data and a few anchor points~\cite{huang2023fast, liu2024learn, wen2023efficient}. Other approaches focus on improving the fusion process itself, for instance by refining spectral embeddings with global-local mechanisms~\cite{wang2023multi} or by adaptively learning a consensus graph filter to generate more cluster-friendly representations~\cite{zhou2023learnable}. In these methods, the graph is a powerful tool constructed from features, not a primary input modality.

\subsection{Multi-view Attributed Graph Clustering}

{\em Multi-view attributed graph clustering} (MVAGC) can be broadly divided into two settings. The first, which is the focus of this work, involves clustering data with a shared graph structure and multiple attribute matrices.
The other setting {\em multiplex graph clustering} handles multiple graph structures, which is outside the scope of our research.
A prominent line of work leverages {\em graph convolutional networks} (GCNs)~\cite{kipf2017semisupervised}. For instance,~\cite{cheng2021multi} employs a two-pathway GCN to learn view-specific and view-consistent information. Following this direction,~\cite{xia2021self} presents a self-supervised GCN that captures non-linear patterns in a complex space guided by pseudo-labels. Another popular strategy is to construct a unified or consensus graph from multiple views. Methods like~\cite{lin2021graph, lin2021multi} utilize graph filtering to generate smooth representations for consensus graph learning.~\cite{pan2021multi} achieves this through a graph-level contrastive loss. Similarly,~\cite{wang2021consistent} combines a multiple graph auto-encoder with attention and a mutual information maximization module to generate a high-quality consensus graph.
Distinct from these deep learning models,~\cite{fettal2023simultaneous} introduces a linear model that simultaneously learns representations and clusters nodes via a simple weighted propagation scheme.

\subsection{Multimodal Graph Learning and Clustering}

{\em Multimodal graph learning} (MGL)~\cite{ektefaie2023multimodal, peng2024learning} is an emerging technique for integrating diverse data modalities within a unified graph structure.
Existing methods for MGL often adapt established techniques such as graph convolutions, attention, and contrastive learning. These have been successfully applied to tasks such as augmenting {\em Large Language Models} (LLMs) for generation~\cite{yoon2023multimodal}, enhancing text-to-image synthesis~\cite{shen2024sg, jin2024instructg2i}, and establishing comprehensive benchmarks for evaluating multimodal graph algorithms~\cite{yan2024graph, zhu2024multimodal, li2024visiongraph, zhu2024benchmarking}.
Although MGL has seen broad applications, its use for unsupervised clustering of MMAGs remains under-explored. 
A recent effort~\cite{wang2025deep} employs a graph Transformer model for deep multi-modal clustering. 
However, this model relies on the Transformer architecture that suffers from a quadratic time complexity $O(n^2)$ and costly graph structural embeddings, and falls short of exploring the characteristics of MMAGs.

\section{Preliminaries}

\subsection{Symbol and Problem Statement}
\stitle{Graph Terminology} A {\em Multimodal Attributed Graph} (MMAG) is represented by a triplet $\G=(\V,\EDG,\{\XM^{(i)}\}_{i=1}^m)$, where $\V=\{v_1,v_2,\ldots,v_n\}$ denotes a set of $n$ nodes, $\EDG$ is a set of edges connecting nodes in $\V$, and $\{\XM^{(i)}\}_{i=1}^m$ is a set of attribute matrices of $m$ modalities (e.g., textual, visual, or acoustic data) that associate with the nodes. 
In particular, given $j$-th node $v_j$ in $\V$, $\XM^{(i)}_j\in \mathbb{R}^{d_i}$ denotes $v_j$'s attribute vector of $i$-th modality, where $d_i$ is the corresponding attribute dimension.
The adjacency matrix of $\G$ is symbolized by $\AM \in \{0,1\}^{n \times n}$ and the corresponding diagonal degree matrix is denoted by $\DM$, where $\DM_{j,j} = \sum_{k=1}^n \AM_{j,k}$. 
Accordingly, the symmetrically normalized adjacency matrix and Laplacian of $\G$ are represented by $\NAM = \DM^{-1/2}\AM\DM^{-1/2}$ and $\LM = \IM - \NAM$, respectively.
In graph signal processing~\cite{shuman2013emerging,isufi2024graph}, a graph signal $f$ over $\G$ is a function from the node set to the field of real numbers, i.e., $f: \V \rightarrow \mathbb{R}$, and we can represent a graph signal as a vector $\zm\in \mathbb{R}^n$. Matrix $\AM$ or $\NAM$ are often referred to as {\em shift operators} of $\G$.

\stitle{Problem Statement}
Given an MMAG $\G=(\V,\EDG,\{\XM^{(i)}\}_{i=1}^m)$ and a specified number $K$ of clusters, the general, hand-waving goal of clustering over $\G$ is to partition the node set $\V$ into $K$ disjoint clusters $\{C_1,C_2,\ldots,C_K\}$ such that (i) nodes within the same cluster $C_j$ are proximal to each other, while nodes across clusters are distant from each other; and (ii) nodes within the same cluster $C_j$ are similar to each other in terms of attributes of $m$ modalities, while the nodes in different clusters are dissimilar.

\subsection{Graph Signal Denoising and GNNs}\label{sec:GSD}

\begin{definition}[Graph Signal Denoising (GSD)\textnormal{~\cite{ma2021unified,dong2016learning}}]\label{def:GSD}
Given a noisy signal $\ZM\in \mathbb{R}^{n\times d}$ on $\G$, GSD seeks to recover a clean signal $\HM \in \mathbb{R}^{n\times d}$ that is smooth over $\G$, i.e., minimizing the objective:
\begin{equation}
\label{eq:denoising_objective}
\min_{\HM\in \mathbb{R}^{n\times d}}{\|\HM - \ZM\|_{\text{F}}^2 + \alpha\cdot\Tr(\HM^{\top}\LM\HM)},
\end{equation}
where $\alpha\in (0,1)$ stands for a trade-off parameter.
\end{definition}

Definition~\ref{def:GSD} presents the GSD~\cite{ma2021unified,dong2016learning,zheng2025adaptive} problem with Laplacian regularization. Specifically, the first term, $\|\HM-\ZM\|_\text{F}^2$, is a fidelity term that ensures the recovered signal $\HM$ is close to its original $\ZM$. On the contrary, the Laplacian regularizer $\Tr(\HM^{\top}\LM\HM)$ renders the signal at any two adjacent nodes in $\G$ to be similar, i.e., enforces the smoothness of $\HM$ over $\G$. 
The coefficient $\alpha$ is to balance the trade-off between fidelity and smoothness.

As unveiled by recent studies~\cite{ma2021unified,yan2024graph,zhu2021interpreting,wang2025gegennet}, many well-known GNNs including GCN/SGC~\cite{kipf2017semisupervised,wu2019simplifying}, APPNP~\cite{gasteiger2019combining}, and others ~\cite{xu2018representation,chen2020simple} essentially learn node representations $\HM$ by optimizing GSD-based objectives as in Eq.~\eqref{eq:denoising_objective}. Mathematically, the closed-form solution is formulated as $\HM = (\IM + \alpha\cdot\LM)^{-1}\ZM$, which can be further approximately expressed by a polynomial in the graph shift operator via Taylor series expansion or Neumman series:
\begin{equation}\label{eq:graph-filter}
\textstyle \HM = \sum_{t=0}^{T}{\alpha_t\cdot \NAM^{t}} \cdot \ZM,
\end{equation}
where $\{\alpha_t\}_{t=0}^T$ is a set of coefficients. The above operation is known as a {\em graph filter} \cite{isufi2024graph,sandryhaila2013discrete} that outputs a linear combination of $T$ shifted signals on $\G$.

\subsection{Empirical Analyses of MMAGs}\label{sec:MMAG-char}

This section empirically investigates the characteristics of MMAGs from modality, node, and feature-wise perspectives, respectively.

\begin{table}[!t]
\centering
\renewcommand{\arraystretch}{0.8} 
\caption{Averaged ADC of MMAGs and MVGC/MVC datasets.}
\label{tab:dcor_metric}
\vspace{-2ex}
\begin{small}
\begin{tabular}{clcc}
\toprule
{Category} & {Dataset} & {\# Views/Modalities} &  {Avg. ADC} \\
\midrule
\multirow{3}{*}{\parbox{1.5cm}{\centering MVC/MVAGC datasets}} & \textit{Amazon Photos} & 2 & 0.9058 \\
& \textit{CiteSeer} & 2 & 0.8922 \\
& \textit{3sources} & 3 & 0.9066 \\ \midrule
\multirow{3}{*}{MMAG} & \textit{Movies} & 2 & 0.2796 \\
& \textit{Grocery-S} & 2 & 0.2494 \\
& \textit{Reddit-S} & 2 & 0.2412 \\
\bottomrule
\end{tabular}
\end{small}
\vspace{-2ex}
\end{table}

\stitle{Modality-wise Uncorrelatedness}
Firstly, we employ the distance correlation~\cite{szekely2007measuring, szekely2009brownian} to quantify the statistical dependence of node attribute vectors from various modalities/views, formally termed as {\em attribute distance correlation} (ADC) in Definition~\ref{def:ADC}.

\begin{definition}[Attribute Distance Correlation]\label{def:ADC}
Given two attribute matrices $\XM^{(i)} \in \mathbb{R}^{n \times d_i}$ and $\XM^{(j)} \in \mathbb{R}^{n \times d_j}$ of node set $\V$, their attribute distance correlation is defined by
\begin{small}
\begin{equation*}
\textsf{ADC}(\XM^{(i)}, \XM^{(j)}) = \frac{\Delta_{\textnormal{Cov}}(\XM^{(i)}, \XM^{(j)})}{\sqrt{\Delta_{\textnormal{Var}}(\XM^{(i)})\cdot \Delta_{\textnormal{Var}}(\XM^{(j)})}} \in [0,1],
\end{equation*}
\end{small}
where $\Delta_{\textnormal{Cov}}(\XM^{(i)}, \XM^{(j)})$ is the distance covariance between $\XM^{(i)}$ and $\XM^{(j)}$, and $\Delta_{\textnormal{Var}}(\XM^{(i)})$
measures the distance variance of $\XM^{(i)}$.
\end{definition}
Intuitively, a higher (resp. low) $\textsf{ADC}(\XM^{(i)}, \XM^{(j)})$ indicates a strong (resp. weak) correlation between two attribute matrices.
Particularly, $\textsf{ADC}(\XM^{(i)}, \XM^{(j)})=0$ if and only if $\XM^{(i)}$ and $\XM^{(j)}$ are completely statistically independent.

Table~\ref{tab:dcor_metric} reports the averaged ADC values of node attributes from multiple modalities in real MMAGs~\cite{yan2024graph} {\em Movies}, {\em Grocery-S}, and {\em Reddit-S}, and those from multiple views in benchmark datasets {\em Amazon Photos}~\cite{shchur2018pitfalls}, {\em CiteSeer}~\cite{sen2008collective}, and {\em 3sources}~\cite{greene2009matrix, liu2013multi} used in MVAGC~\cite{li2025clustering, lin2021graph, lin2021multi, cheng2021multi, pan2021multi, fettal2023simultaneous, chen2025variational} or MVC~\cite{zhou2024survey, wen2023efficient, zhou2023learnable, wang2023multi} works.
It can be observed that the ADC values of classic multi-view datasets are all more than $0.89$.
Such a high dependence among attribute matrices in multi-view datasets is ascribed to the fact that most of the attribute views are artificially extracted or constructed from same or similar sources, e.g., generated via the Cartesian product of a given set of node attributes. 

In contrast, MMAGs consistently present low ADC values (below $0.28$), i.e., weak modality-wise correlation, which implies highly diverse and distinct nodal attributes across modalities. The reason is that the input attributes in MMAGs are garnered from radically heterogeneous sources (e.g., text, images), and are often derived from distinct models or feature extractors tailored to each modality’s characteristics, e.g., embeddings generated by pre-trained language models (PLMs) and vision models (PVMs). In turn, these embeddings typically reside in different latent spaces due to differences in model architectures and semantics within each modality.

\stitle{Node- and Feature-domain Outlierness}
On the other hand, at both node and feature levels, the input node attributes (embeddings from PLMs/PVMs) of each modality embody noisy and/or task-irrelevant information from the raw textual/visual contents or the models/encoders pre-trained on general-purpose corpora.
For instance, raw images often contain elements unrelated to the clustering task, such as barcodes or logos.
A portion of nodes in practical MMAGs usually contain missing and contaminated textual/visual data, engendering feature-level outliers. 

\begin{table}[!t]
\centering
\renewcommand{\arraystretch}{0.8} 
\caption{Percentages (\%) of nodes/features with at least one outlier in input attribute matrices of real MMAG datasets.}
\label{tab:outliers}
\vspace{-2ex}
\begin{small}
\begin{tabular}{lccc}
\toprule
{Dataset} & Modality & {Node-domain} & {Feature-domain} \\
\midrule
\multirow{2}{*}{\em Grocery-S} & Text  & 100 & 74.93 \\
 & Image & 100 & 74.93 \\ \midrule
\multirow{2}{*}{\em Reddit-S} & Text   & 100 & 89.29 \\
 & Image  & 100 & 89.29 \\
\bottomrule
\end{tabular}
\end{small}
\vspace{-2ex}
\end{table}

\begin{definition}[Outlierness via Z-score\textnormal{~\cite{heckert2003nist, iglewicz1993volume}}]
\label{def:zscore_outlier}
Let $\XM \in \mathbb{R}^{n \times d}$ be a data matrix. For $j$-th column, we denote by $\mu_j$ and $\sigma_j$ its mean and standard deviation, respectively. The Z-score $z_{i,j}$ for each data entry $\XM_{i,j}$ is calculated as: $z_{i,j} = \frac{\XM_{i,j} - \mu_j}{\sigma_j},\ \text{for } \sigma_j \neq 0$.
$\XM_{i,j}$ is said to be an outlier if $|z_{ij}|$ exceeds a predefined threshold $\tau$ and its magnitude quantifies the degree of outlierness.
\end{definition}

Formally, to measure the degree of outlierness in nodes and features, we calculate the averaged Z-score in Definition~\ref{def:zscore_outlier} for rows and columns of attribute matrices $\{\XM^{(i)}\}_{i=1}^m$ in real MMAGs, i.e., {\em Grocery-S} and {\em Reddit-S}.
Following common practice, we set the anomaly threshold $\tau=4$. Table \ref{tab:outliers} reveals a notable presence of outliers at both node and feature levels in each modality of real datasets. For example, on {\em Grocery-S}, all images contain at least one anomaly feature/attribute value; and $89.29\%$ of textual attributes in {\em Reddit-S} have at least an outlier entry in a node. 

\stitle{Limitations of Prior Methods on MMAGs}
Recall that state-of-the-art MVC and MVAGC approaches are built upon GNNs that operate graph filtering as in Eq.~\eqref{eq:graph-filter}.
However, as analyzed in \S~\ref{sec:GSD}, in essence, such graph filters are solely to perform node-domain denoising, which are incompetent for mitigating severe feature-domain noise on real MMAGs. Secondly, most existing MVC/MVAGC works learn representation $\HM$ by aligning the embedding spaces of $\HM$ and the data of each view {\em severally} through optimizing a summed loss, e.g., intra-view contrastive learning~\cite{pan2021multi}. This paradigm excels at multi-view datasets with high correlation across views (see Table~\ref{tab:dcor_metric}), but is ill-suited for MMAGs as it is hard to converge to a consensus and might result in uninformative $\HM$ on MMAGs due to the large gaps in the semantic spaces of multiple modalities.

\section{Methodology}
\begin{figure}[!t]
\centering
\includegraphics[width=1\columnwidth]{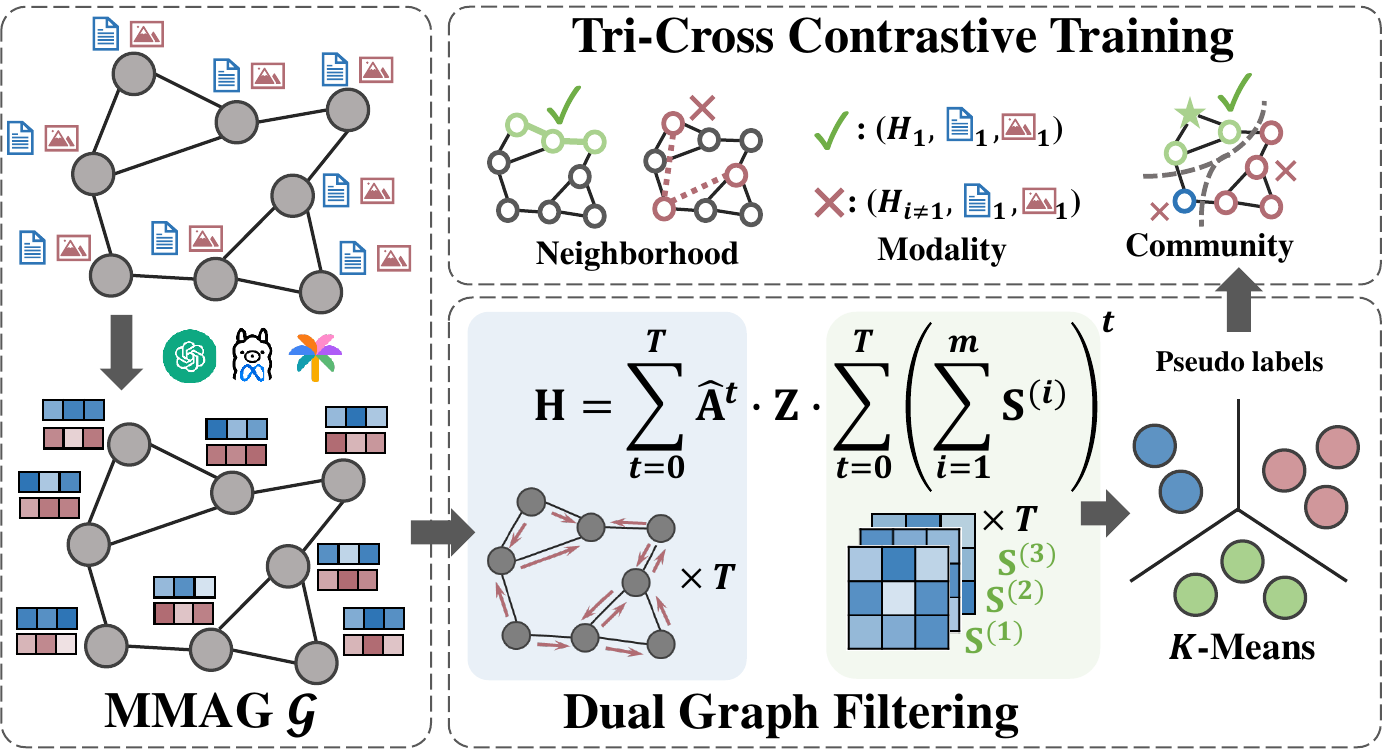}
\vspace{-5ex}
\caption{Overview of \algo{}.}
\label{fig:pipeline}
\vspace{-2ex}
\end{figure}

This section introduces our novel framework \algo{} for clustering over MMAGs.
As overviewed in Fig.~\ref{fig:pipeline}, our training pipeline mainly comprises two constituent components: {\em dual graph filtering} and {\em tri-cross contrastive training}, which aim to construct node representations $\HM$ and clusters, and guide the representation learning with the consideration of semantic similarities of node pairs across modalities, graph neighborhoods, and communities, respectively.

\subsection{Denoising via Dual Graph Filtering}

Firstly, we propose the {\em dual graph filter} for generating $\HM$ to overcome the deficiency of existing graph filters pinpointed in the preceding section.

\stitle{Optimization Objective with FDD}
Instead of focusing on node-domain denoising as in Definition~\ref{def:GSD}, \algo{} additionally introduces a {\em feature-domain denoising} (FDD) term~\cite{shahid2016fast} and formulates the new optimization objective for constructing node representations $\HM \in \mathbb{R}^{n\times d}$ as follows: 
\begin{small}
\begin{equation}\label{eq:obj}
\min_{\HM}{\|\HM-\ZM\|^2_\text{F} + \alpha\cdot\Tr(\HM^{\top}\LM\HM) + \frac{\beta}{m}\cdot\sum_{i=1}^{m}{\Tr(\HM(\IM-\SM^{(i)})\HM^{\top})}}.
\end{equation}
\end{small}
Therein, $\ZM\in \mathbb{R}^{n\times d}$ represents initial node features, which is combined from $m$ modalities:
\begin{equation*}
\ZM = \textsf{combine}(\ZM^{(1)},\ldots,\ZM^{(m)}),
\end{equation*}
where $\textsf{combine}(\cdot)$ is a weighted summation function and $\ZM^{(i)} = \XM^{(i)}\WM^{(i)} \in \mathbb{R}^{n\times d}$ is transformed from input node attributes $\XM^{(i)}$ of $i$-th modality via a linear layer $\WM^{(i)}$.
For each $i$-th modality, $\SM^{(i)}\in \mathbb{R}^{d\times d}$ stands for the shift operator over the affinity graph constructed for its $d$ distinct feature dimensions, and coefficients $\alpha \ge 0$ and $\beta \ge 0$ control the importance of the second and third terms, i.e., node-domain and feature-domain denoising.

Distinct from the node-domain denoising term $\alpha\cdot\Tr(\HM^{\top}\LM\HM)$  that promotes the smoothness of $\HM$'s rows over $\G$ as illustrated in \S~\ref{sec:GSD}, our additional FDD term is to enforce that the corresponding columns in $\HM$ of similar (resp. dissimilar) feature dimensions in each modality of data $\ZM^{(i)}$ should be close (resp. distance), i.e., $\HM$'s columns are smooth over shift operators $\{\SM^{(i)}\}_{i=1}^m$.

\stitle{Constructing $\{\SM^{(i)}\}_{i=1}^m$}
Let $f(\HM)$ be the objective function in Eq.~\eqref{eq:obj}. Importantly, $\IM-\SM^{(i)}$ must be positive semidefinite; otherwise, the optimization problem $f(\HM)$ will be non-convex, and the solution $\frac{\partial f(\HM)}{\partial \HM}$ might converge to a saddle point.

When $\IM-\SM^{(i)}$ is positive semidefinite, with the definition of $\LM=\IM-\NAM$, we can get the closed-form solution of the problem $f(\HM)$ by setting $\frac{\partial f(\HM)}{\partial \HM}=0$, yielding
\begin{small}
\begin{equation}\label{eq:DGF-H-optimal}
\HM = \frac{1}{(\alpha+1)(\beta+1)}\cdot\left(\IM - \frac{\alpha}{\alpha+1}\NAM \right)^{-1}\cdot\ZM \cdot\left(\IM - \frac{\beta}{(\beta+1)m}\sum_{i=1}^m\SM^{(i)} \right)^{-1}.
\end{equation}
\end{small}

\begin{theorem}[\cite{horn2012matrix}]\label{lem:neu}
Let $\MM$ be a matrix whose dominant eigenvalue $\lambda(\MM)$ satisfies $\lambda(\MM)<1$. Then, the inverse $(\IM-\MM)^{-1}$ can be expanded as a Neumann series: $(\IM-\MM)^{-1}=\sum_{\ell=0}^\infty\MM^\ell$.
\end{theorem}

According to Theorem~\ref{lem:neu}, we can sidestep the prohibitive matrix inversion $\left(\IM - \frac{\beta}{(\beta+1)m}\sum_{i=1}^m\SM^{(i)} \right)^{-1}$ and approximate it using a finite series if the dominant eigenvalue of $\frac{\beta}{(\beta+1)m}\sum_{i=1}^m\SM^{(i)}$ is bounded by $1$, meaning that each shift operator $\SM^{(i)}$ should satisfy $\lambda(\SM^{(i)})\le 1$.

Based on the aforementioned requirements: (i) $\IM-\SM^{(i)}$ is positive semidefinite, and (ii) $\lambda\left(\frac{\beta}{(\beta+1)m}\sum_{i=1}^m\SM^{(i)}\right)\le 1$, we propose to build the affinity graph for the $i$-th modality based on the {\em exponential dot product kernel}~\cite{kar2012random} and {\em symmetric softmax function}:
\begin{small}
\begin{equation}\label{eq:Si}
\SM^{(i)}_{j,\ell} = \frac{\exp{\left(\frac{\ZM_{\cdot,j}^{\top}\ZM_{\cdot,\ell}}{\sqrt{n}}\right)}}{\sqrt{\sum_{x=1}^d \exp{\left(\frac{\ZM_{\cdot,j}^{\top}\ZM_{\cdot,x}}{\sqrt{n}}\right)}} \cdot \sqrt{\sum_{x=1}^d \exp{\left(\frac{\ZM_{\cdot,\ell}^{\top}\ZM_{\cdot,x}}{\sqrt{n}}\right)}}},
\end{equation}
\end{small}
where $\sqrt{n}$ is to scale down the values for well‐behaved and stable gradients in model training, and each column in $\ZM^{(i)}$ is $L_2$-normalized before calculating $\SM^{(i)}$.
The following lemma establishes the correctness of the aforementioned construction of $\SM^{(i)}$.
\begin{lemma}\label{lem:S-property}
$\SM^{(i)}\ \forall{1\le i\le m}$ in Eq.~\eqref{eq:Si} is positive semidefinite and the dominant eigenvalue of $\frac{\beta}{(\beta+1)m}\sum_{i=1}^m\SM^{(i)}$ is not greater than $1$.
\end{lemma}

\header
{\em\bf Remark.} It is worth mentioning that $\SM^{(i)}$ can be interpreted as a symmetric version of the well-known {\em attention mechanism}~\cite{vaswani2017attention} applied on feature dimensions of the embeddings.

\stitle{Generating Representations and Clusters}
Building on Theorem~\ref{lem:neu} and Lemma~\ref{lem:S-property}, we can prove the following lemma:
\begin{lemma}\label{lem:dual-filtering}
The closed-form solution to the optimization objective in Eq.~\eqref{eq:obj} can be expressed as an infinite series expansion as follows:
\begin{small}
\begin{equation}\label{eq:DGF-H}
\HM = \frac{1}{(\alpha+1)(\beta+1)}\sum_{t=0}^\infty{\left(\frac{\alpha}{\alpha+1}\NAM\right)^t}\cdot {\ZM}\cdot \sum_{t=0}^\infty{\left(\frac{\beta}{(\beta+1)m}\sum_{i=1}^m{\SM^{(i)}}\right)^t}.
\end{equation}
\end{small}
\end{lemma}

This formula suggests that the final node representations $\HM$ is obtained by simultaneously applying a node-domain and a feature-domain graph filter on both l.h.s and r.h.s. of initial node features $\ZM$. 
Note that $\HM$ can be iteratively approximated using $T$ iterations, resulting the following truncated series:
\begin{small}
\begin{equation}\label{eq:DGF-H-truncated}
\HM = \frac{1}{(\alpha+1)(\beta+1)}\sum_{t=0}^T{\left(\frac{\alpha}{\alpha+1}\NAM\right)^t}{\ZM}\cdot \sum_{t=0}^T{\left(\frac{\beta}{(\beta+1)m}\sum_{i=1}^m{\SM^{(i)}}\right)^t}.
\end{equation}
\end{small}
For the interest of space, we defer the pseudo-code for computing $\HM$ to Appendix~\ref{sec:comp-H}.

Afterwards, a standard $K$-Means~\cite{lloyd1982least, elkan2003using} is applied over rows of $\HM$ to generate $K$ distinct clusters $\{C_1,C_2,\ldots,C_K\}$.

\begin{figure}[!t]
\centering
\includegraphics[width=1\columnwidth]{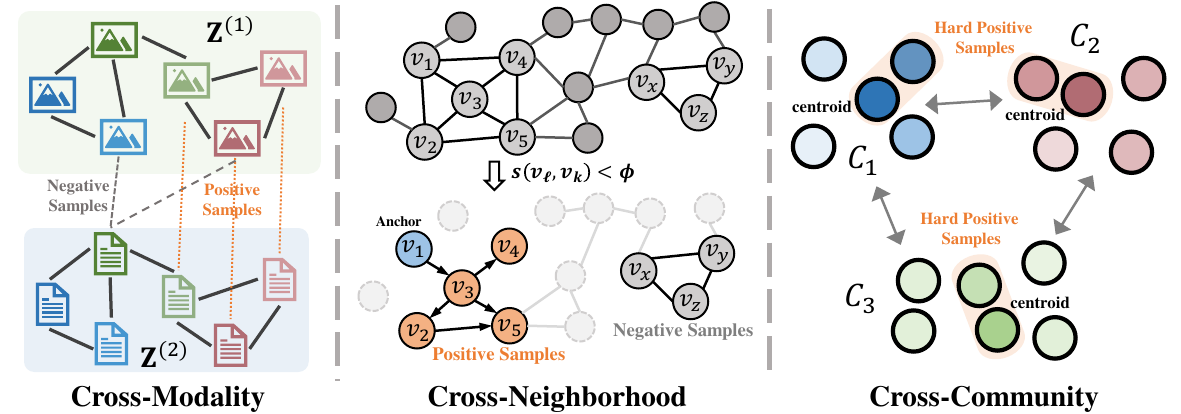}
\vspace{-5ex}
\caption{Illustration of the Tri-Cross Contrastive Training.}
\label{fig:loss}
\vspace{-2ex}
\end{figure}

\subsection{Tri-Cross Contrastive Training}
To update parameters $\{\WM^{(i)}\}_{i=1}^m$ and learn robust and discriminative node representations $\HM$, we resort to instance-level contrastive learning, which is to encourage the agreement of similar node pairs, while discouraging the agreement of the dissimilar ones~\cite{chuang2020debiased,oord2018representation}.
Instead of creating positive/negative pairs by data perturbation or augmentation~\cite{xiao2020should} that is commonly adopted, we propose the {\em tri-cross contrastive training} strategy (as illustrated in Fig.~\ref{fig:loss}) that contrasts nodes across multiple modalities, graph neighborhoods, and communities, i.e., minimizing the joint loss $\mathcal{L}_{\textnormal{mod}}+ \mathcal{L}_{\textnormal{nbr}} + \mathcal{L}_{\textnormal{comm}}$. 

\stitle{Cross-Modality Contrastive Loss $\mathcal{L}_{\textnormal{mod}}$}
Recall that in \S~\ref{sec:MMAG-char}, it is shown that there is a high inconsistency in the semantic spaces of different modalities. In response, \algo{} learns a joint embedding space through a cross-modality contrastive loss $\mathcal{L}_{\textnormal{mod}}$. Given the node features $\ZM^{(0)},\ZM^{(1)},\ldots,\ZM^{(m)}$ of $m+1$ modalities, where $\ZM^{(0)}=\HM$ encodes graph structures, and thus, is deemed as an additional modality, the basic idea of $\mathcal{L}_{\textnormal{mod}}$ is to maximize the inter-modality similarities of the same nodes and meanwhile minimize them for imposter node pairs. More specifically, we define the cross-modalities contrastive loss in Eq.~\eqref{eq:cm_loss_full} as a summation of the {\em masked margin softmax} (MMS) losses~\cite{ilharco2019large,chen2021multimodal} for any two distinct modalities.
 \begin{equation} 
 \label{eq:cm_loss_full} 
 \mathcal{L}_{\textnormal{mod}} = \sum_{i=0}^{m}\sum_{j=0,j\neq i}^m{\textsf{MMS}(\ZM^{(i)}, \ZM^{(j)})}.
 \end{equation}
In particular, $\textsf{MMS}(\ZM^{(i)}, \ZM^{(j)})$ is to discriminate positive pairs \\$(\ZM^{(i)}_\ell, \ZM^{(j)}_\ell)\ \forall{1\le \ell \le n}$ from negative pairs that contrast each $\ZM^{(i)}_\ell$ (resp. $\ZM^{(j)}_\ell$) with other imposter instances $\tilde{\ZM}^{(j)}_k$ (resp. $\tilde{\ZM}^{(i)}_k$) from the other modality, which can be formulated by 
 \begin{equation}\label{eq:cm_pair_l2norm_full}
 \begin{aligned}
\textsf{MMS}(\ZM^{(i)}, \ZM^{(j)}) = -\frac{1}{n} \sum\limits_{\ell=1}^{n} \Bigg[ \log \frac{\displaystyle e^{\ZM^{(i)}_\ell \cdot \ZM^{(j)}_{\ell} - \delta}}{\displaystyle e^{\ZM^{(i)}_\ell \cdot \ZM^{(j)}_{\ell} - \delta} + \textstyle \sum\limits_{\substack{k=1, k\neq \ell}}^{n}e^{\tilde{\ZM}^{(i)}_k \cdot \ZM^{(j)}_{\ell}}} \\
+ \log \frac{\displaystyle e^{\ZM^{(i)}_\ell \cdot \ZM^{(j)}_{\ell} - \delta}}{\displaystyle e^{\ZM^{(i)}_\ell \cdot \ZM^{(j)}_{\ell} - \delta} + \textstyle \sum\limits_{\substack{k=1, k\neq \ell}}^{n}{e^{\ZM^{(i)}_\ell \cdot \tilde{\ZM}^{(j)}_{k}}}} \Bigg],
 \end{aligned}
 \end{equation} 
where $\delta$ is a small margin. In doing so, $\ZM^{(0)},\ZM^{(1)},\ldots,\ZM^{(m)}$ that are even from different modalities can be projected into a common embedding space, facilitating the learning of $\HM$.

\stitle{Cross-Neighborhood Contrastive Loss $\mathcal{L}_{\textnormal{nbr}}$}
To further capture the semantic relations of nodes underlying the graph topology, we introduce the cross-neighborhood contrastive loss $\mathcal{L}_{\textnormal{nbr}}$.
A simple and straightforward way to construct intra-neighborhood pairs (i.e., positive pairs) and inter-neighborhood (i.e., negative pairs) is to sample random walks from each node over the input graph $\G$, as in previous works~\cite{devvrit2022s3gc}.
However, real graphs often inherently contain spurious links, making the sampled neighborhoods noisy.
As a remedy, we propose to leverage the complementary nature between the graph structures and nodal attributes of multiple modalities, and develop an {\em attribute-aware sampling} (AAS) scheme for selecting positive pairs.

Specifically, \algo{} first eliminates spurious edges from the original graph $\G$ by a filtering step, yielding a pruned graph $\G^\prime$. An edge $(v_\ell,v_k)\in \EDG$ is removed if its cross-modal attribute similarity $s(v_\ell,v_k)$ is below a certain threshold $\phi$, i.e.,
\begin{equation}
s(v_\ell,v_k) = \sum_{i<j}{\ZM^{(i)}_\ell \cdot \ZM^{(j)}_k} < \phi.
\end{equation}
To adaptively determine the threshold $\phi$, we randomly sample $|\EDG|$ node pairs $\mathcal{Q}$ from $\G$ and obtain their cross-modal attribute similarities $\mathcal{S}=\{s(v_\ell,v_k)|(v_\ell,v_k)\in \mathcal{Q}\}$. The threshold $\phi$ is then computed by $\phi = \mu(\mathcal{S}) + \sigma(\mathcal{S})$, where $\mu(\cdot)$ and $\sigma(\cdot)$ return the mean and standard deviation, respectively. The rationale is to retain edges whose cross-modal attribute similarities are notably higher than the average similarity in a normal distribution.

Subsequently, for each anchor node $v_i\in \V$, \algo{} simulates random walks starting from $v_i$ on pruned graph $\G^\prime$, and utilize the $\ell$ visited nodes $\mathcal{P}_i = \{p_1,p_2,\ldots,p_\ell\}$ as its neighborhood (i.e., positive samples).
On the contrary, its negative samples $\N_i=\{n_1,n_2,\ldots,n_\ell\}$ are formed by randomly picking $\ell$ nodes from $\V\setminus \mathcal{P}_i$. Based thereon, the cross-neighborhood contrastive loss for each node $v_i\in \V$ is calculated by
\begin{equation} \label{eq:L_nbr_loss_per_node}
\mathcal{L}_{\textnormal{nbr}}^{(i)} = -\log \frac{\sum_{v_j \in \mathcal{P}_i} e^{\HM_i \cdot \HM_j}}{\sum_{v_j \in \mathcal{P}_i} e^{\HM_i \cdot \HM_j} + \sum_{v_k \in \N_i} e^{\HM_i \cdot \HM_{k}}},
\end{equation}
which leads to the combined one by $\mathcal{L}_{\textnormal{nbr}} = \sum_{i=1}^{n} \mathcal{L}_{\textnormal{nbr}}^{(i)}$.

\stitle{Cross-Community Contrastive Loss $\mathcal{L}_{\textnormal{comm}}$}
Lastly, inspired by the cross-prediction scheme~\cite{caron2020unsupervised}, \algo{} includes a cross-community contrastive loss $\mathcal{L}_{\textnormal{comm}}$, which aims at strengthening the cohesiveness of intra-community instances.
Considering that the intermediate clusters might not accurate, i.e., containing misassigned nodes, we implement the contrastive loss using a {\em hard positive sampling} (HPS) strategy for enhanced robustness \cite{zhang2024deep}.

That is, given the latest set of $K$ clusters $\{C_1,C_2,\ldots,C_K\}$ produced by $K$-Means on current $\HM$, we denote $\{c_1,c_2,\ldots,c_k\}$ be their corresponding centroids identified by $K$-Means. Next, a set $\mathcal{H}_k$ of hard samples pertinent to each cluster $C_k$ is obtained by picking the top-$(|C_k|\cdot \theta)$ ($\theta\in (0,1)$) nodes that are closest to the centroid $c_k$ in cluster $C_k$ in terms of the cosine similarity of their embeddings in $\HM$. The cross-community contrastive loss thus can be computed by contrasting hard samples to their respective centroids:
\begin{equation}
\mathcal{L}_{\textnormal{comm}} = -\frac{1}{n} \sum_{k=1}^{K} \log\left( \frac{\sum_{v_j \in \mathcal{H}_{k}} e^{\HM_j \cdot \HM_{c_k}}}{ \sum_{k=1}^{K}{\sum_{v_j \in C_{k}} e^{\HM_j \cdot \HM_{c_k}}}} \right).
\end{equation}

\subsection{Theoretical Analyses}

The following theorem states that the computation of $\HM$ by Eq.~\eqref{eq:DGF-H} in \algo{} is equivalent to a low-pass filtering of node features simultaneously in node and feature domains, which preserves essential low-frequency structures while attenuating high-frequency noise.
\begin{theorem}\label{lem:dual-low-pass}
Both $\sum_{t=0}^\infty{\left(\frac{\alpha}{\alpha+1}\NAM\right)^t}$ and $\sum_{t=0}^\infty{\left(\frac{\beta}{(\beta+1)m}\sum_{i=1}^m{\SM^{(i)}}\right)^t}$ are low-pass filters.
\end{theorem}

Moreover, in Theorem~\ref{lem:approxi}, we establish an approximation guarantee for our $T$-truncated node-domain filter in Eq.~\eqref{eq:DGF-H-truncated}. Note that a result can be easily derived for the feature-domain filter in Eq.~\eqref{eq:DGF-H-truncated} in an analogous way.
\begin{theorem}\label{lem:approxi}
Let $(\IM + \alpha\LM)^{-1}$ be the ideal low-pass filter with spectral response $h(\lambda_k) = (1+\alpha\lambda_k)^{-1}$ for an eigenvalue $\lambda_k$ of $\LM$. The spectral response $h^{(T)}(\lambda_k)$ of $\frac{1}{\alpha+1}\sum_{t=0}^T{\left(\frac{\alpha}{\alpha+1}\NAM\right)^t}$ is an multiplicative approximation of $h(\lambda_k)$: $h^{(T)}(\lambda_k) = h(\lambda_k) \left(1 - \left(\frac{\alpha(1-\lambda_k)}{\alpha+1}\right)^{T+1}\right)$.
\end{theorem}

We further theoretically uncover other implications of our dual graph filtering paradigm through the lens of singular value penalization, penalization in the spectral graph fourier domain, subspace alignment and rotation, respectively. The detailed discussions are provided in Appendix~\ref{sec:additional_theoretical_analyses}.

\stitle{Time Complexity Analysis} 
The computation per epoch in \algo{} model involves several steps: (i) transformation of input node attributes from $m$ modalities to $\ZM$ that takes $O(\sum_{i=1}^m n d_i d)$ time; (ii) the constructions of $\{\SM^{(i)}\}_{i=1}^m$ requiring $O(m n d^2)$ time; (iii) the calculation of $\HM$ by Eq.~\eqref{eq:DGF-H-truncated} in $O(Td^3 + T|\EDG|d)$ time; (iv) running $K$-Means on $\HM$ using $O(Knd)$ time; and (v) the computations of our contrastive losses for a mini-batch of size $b$ that entail $O(nbd)$ time.
Therefore, the overall time complexity is bounded by $O( Td\cdot (d^2 + |\EDG|) + nd\cdot (m d + K + b + \sum_{i=1}^m d_i))$. Since $d$ is often less than $d^{(i)}$, the complexity can be reduced as $O(d\cdot (d^2 + |\EDG|)+ nd\cdot \sum_{i=1}^m d_i)$ when $T, K$, and $b$ are regarded as constants.

\section{Experiments}

\begin{table}[!t]
\centering
\caption{Statistics of experimented MMAG datasets~\cite{yan2024graph}.}
\label{tab:dataset_stats}
\vspace{-2ex}
\begin{footnotesize}
\begin{tabular}{lcccc}
\toprule
{Dataset} & {\# Nodes} & {\# Edges} & {\# Classes} & {Modalities} \\
\midrule
\textit{\textit{Movies}}         & 16,672   & 160,802   & 20    &  Text, Image    \\
\textit{Toys}           & 20,695   & 113,402   & 18    &  Text, Image    \\
\textit{Grocery-S}      & 17,074   & 142,262   & 20     & Text, Image    \\
\textit{Grocery}        & 84,379   & 1,590,082 & 20     &  Text, Image   \\
\textit{Reddit-S}       & 15,894   & 283,080   & 20    &  Text, Image    \\
\textit{Reddit}         & 99,638   & 583,594   & 50     & Text, Image    \\
\textit{Photo}          & 36,731   & 615,958   & 12     & Text, Image    \\
\textit{Arts}           & 58,487   & 812,252   & 11     & Text, Image    \\
\bottomrule
\end{tabular}
\end{footnotesize}
\vspace{-4ex}
\end{table}

This section experimentally evaluates \algo{} against 14 competitors on 8 real MMAGs regarding clustering performance. The source code is available at \url{https://github.com/HaoranZ99/DGF}.

\begin{table*}[!t]
\centering
\caption{Clustering quality (ACC, NMI, F1, ARI, and CS) of \algo{} and baselines on real MMAGs. `-' denotes failure due to out of memory. The best (resp. runner-up) is highlighted in {darker} (resp. {light}) blue.
}
\label{tab:clustering_results}
\vspace{-2ex}
\setlength{\tabcolsep}{3pt} \renewcommand{\arraystretch}{0.8}
\begin{footnotesize}
\begin{tabular}{@{}ll|cccccccccccccc|cc@{}}
\toprule
{Dataset} & {\bf } & \texttt{$K$-Means} & \texttt{MVGRL} & \texttt{S3GC} & \texttt{DMoN} & \texttt{Dink-Net} & \texttt{EMVGC-LG} & \texttt{MCLGF} & \texttt{GLSEF} & \texttt{FastMICE} & \texttt{AEVC} & \texttt{MCGC} & \texttt{MGCCN} & \texttt{LMGEC} & \texttt{VGMGC} & \algo{} &  Improv. \\ \midrule
\multirow{5}{*}{\textit{Movies}} & ACC & 16.92 & 22.88 & 24.12 & {24.21} & 16.41 & 18.31 & \cellcolor{cyan!10}{24.86} & 19.34 & 19.6 & 17.35 & 22.28 & 23.37 & 22.64 & 22.41 & \cellcolor{cyan!30}{26.02} & +1.16 \\
& NMI & 13.54 & 19.43 & \cellcolor{cyan!10}{20.98} & 20.37 & 12.05 & 17.51 & 17.11 & 12.28 & 16.18 & 15.68 & 8.04 & 18.63 & 18.81 & 16.49 & \cellcolor{cyan!30}{22.45} & +1.47 \\
& F1 & 13.13 & 15.3 & 18.17 & 18.33 & 13.47 & 12.01 & \cellcolor{cyan!10}{19.31} & 13.77 & 13.06 & 11.52 & 9.76 & 17.57 & 17.95 & 17.66 & \cellcolor{cyan!30}{20.91} & +1.6 \\
& ARI & 3.87 & 5.77 & 7.06 & \cellcolor{cyan!10}{7.62} & 3.24 & 4.04 & 2.78 & 3.98 & 5.32 & 3.65 & 1.31 & 7.09 & 6.79 & 6.11 & \cellcolor{cyan!30}{8.12} & +0.5 \\
& CS & 12.16 & 18.12 & 19.01 & 18.31 & 10.88 & \cellcolor{cyan!10}{19.81} & 16.48 & 11.36 & 14.49 & 17.77 & 16.36 & 16.72 & 16.84 & 14.85 & \cellcolor{cyan!30}{20.15} & +0.34 \\ \midrule
\multirow{5}{*}{\textit{Toys}} & ACC & 24.62 & 35.27 & \cellcolor{cyan!10}{50.84} & 40.28 & 19.6 & 50.07 & 39.88 & 15.35 & 39.97 & 43.48 & 12.19 & 40.71 & 42.94 & 24.81 & \cellcolor{cyan!30}{55.75} & +4.91 \\
& NMI & 21.45 & 34.58 & \cellcolor{cyan!10}{49.06} & 42.84 & 15.24 & 47.51 & 38.67 & 9.51 & 34.56 & 40.42 & 6.54 & 34.65 & 37.08 & 21.31 & \cellcolor{cyan!30}{52.37} & +3.31 \\
& F1 & 23.29 & 25.41 & \cellcolor{cyan!10}{44.26} & 37.02 & 18.34 & 36.42 & 28.55 & 12.29 & 26.71 & 31.05 & 5.09 & 37.82 & 38.93 & 23.37 & \cellcolor{cyan!30}{50.79} & +6.53 \\
& ARI & 10.1 & 18.44 & \cellcolor{cyan!10}{34.89} & 26.54 & 5.94 & 32.01 & 21.59 & 1.09 & 21.48 & 26.19 & -0.2 & 21.25 & 23.8 & 9.19 & \cellcolor{cyan!30}{38.53} & +3.64 \\
& CS & 21.11 & 36.35 & \cellcolor{cyan!10}{48.4} & 42.15 & 15.29 & \cellcolor{cyan!10}{48.4} & 36.19 & 13.23 & 34.09 & 41.07 & 28.82 & 33.96 & 36.46 & 20.96 & \cellcolor{cyan!30}{51.47} & +3.07 \\ \midrule
\multirow{5}{*}{\textit{Grocery-S}} & ACC & 20.9 & 39.48 & 41.81 & 42.71 & 21.64 & 43.31 & \cellcolor{cyan!10}{44.41} & 20.22 & 29.05 & 38.69 & 17.8 & 32.56 & 33.87 & 25.15 & \cellcolor{cyan!30}{53.41} & +9 \\
& NMI & 15.63 & 37.56 & 44.93 & 44.75 & 13.99 & \cellcolor{cyan!10}{51.57} & 42.06 & 12.34 & 24.71 & 41.36 & 8.98 & 29.3 & 34.23 & 21.19 & \cellcolor{cyan!30}{54} & +2.43 \\
& F1 & 19.53 & 28.9 & 34.33 & 33.37 & 17.32 & 33.06 & \cellcolor{cyan!10}{37.79} & 14.33 & 19.21 & 30.08 & 8.43 & 26.66 & 30.36 & 23.56 & \cellcolor{cyan!30}{45.42} & +7.63 \\
& ARI & 6.76 & 21.14 & 27.94 & 31.15 & 3.24 & 27.61 & \cellcolor{cyan!10}{32.05} & 4.2 & 13.8 & 25.51 & 1.04 & 18.11 & 18.63 & 9.25 & \cellcolor{cyan!30}{41.36} & +9.31 \\
& CS & 15.18 & 39.5 & 43.84 & 43.36 & 13.94 & \cellcolor{cyan!10}{51.95} & 40.4 & 14.84 & 23.91 & 42.94 & 25.13 & 28.73 & 33.3 & 20.85 & \cellcolor{cyan!30}{52.39} & +0.44 \\ \midrule
\multirow{5}{*}{\textit{Grocery}} & ACC & 16.73 & \cellcolor{cyan!10}{37.26} & 31.6 & 35.13 & 17.71 & 36.26 & - & 18.05 & 20.81 & 34.34 & - & - & 30.45 & - & \cellcolor{cyan!30}{38.52} & +1.26 \\
& NMI & 13.84 & 33.86 & 40.59 & 42.76 & 12.2 & \cellcolor{cyan!10}{45.35} & - & 0.93 & 19.9 & 37.14 & - & - & 33.55 & - & \cellcolor{cyan!30}{46.67} & +1.32 \\
& F1 & 12.3 & 15.32 & 22.04 & 25.79 & 12.37 & \cellcolor{cyan!10}{28.93} & - & 17.95 & 17.69 & 28.01 & - & - & 23.96 & - & \cellcolor{cyan!30}{33.62} & +4.69 \\
& ARI & 6.77 & 19.46 & 23.86 & \cellcolor{cyan!10}{28.15} & 5.85 & 22.32 & - & 0.58 & 11.04 & 22.2 & - & - & 17.95 & - & \cellcolor{cyan!30}{31.89} & +3.74 \\
& CS & 12.7 & 36.54 & 37.47 & 39.53 & 11.33 & \cellcolor{cyan!10}{42.05} & - & 1.17 & 18.26 & 40.8 & - & - & 31.01 & - & \cellcolor{cyan!30}{43.05} & +1 \\ \midrule
\multirow{5}{*}{\textit{Reddit-S}} & ACC & 36.05 & 71.02 & 72.24 & \cellcolor{cyan!10}{79.13} & 21.98 & 75.34 & 75.95 & 20.55 & 75.58 & 61.19 & 9.59 & 71.44 & 64.89 & 60.72 & \cellcolor{cyan!30}{89.66} & +10.53 \\
& NMI & 40.81 & 76.52 & 71.66 & \cellcolor{cyan!10}{80.36} & 20.78 & 78.07 & 78.6 & 22.31 & 77.51 & 60.76 & 4.28 & 74.55 & 66.84 & 65.65 & \cellcolor{cyan!30}{88.24} & +7.88 \\
& F1 & 32.3 & 63.01 & 62.63 & 64.4 & 19.56 & 71.44 & \cellcolor{cyan!10}{72.04} & 14.9 & 71.73 & 50.54 & 3.04 & 63.4 & 54.27 & 53.34 & \cellcolor{cyan!30}{81.98} & +9.94 \\
& ARI & 24.33 & 65.96 & 64.86 & \cellcolor{cyan!10}{77.72} & 9.09 & 69.63 & 70.11 & 5.32 & 69.95 & 47.5 & -0.08 & 66.05 & 58.02 & 51.16 & \cellcolor{cyan!30}{88.07} & +10.35 \\
& CS & 40.39 & 75.79 & 70.74 & \cellcolor{cyan!10}{81.2} & 20.67 & 78.65 & 77.38 & 29.62 & 76.87 & 61.54 & 33.35 & 73.76 & 66.32 & 65.09 & \cellcolor{cyan!30}{88.02} & +6.82 \\ \midrule
\multirow{5}{*}{\textit{Reddit}} & ACC & 27.07 & - & 59.7 & 45.31 & 13.71 & 58.88 & - & 11.61 & \cellcolor{cyan!10}{61.17} & 42.03 & - & - & 42.03 & - & \cellcolor{cyan!30}{62.87} & +1.7 \\
& NMI & 44.12 & - & 62.92 & 56.81 & 19.8 & 67.14 & - & 16.2 & \cellcolor{cyan!10}{67.86} & 49.8 & - & - & 50.46 & - & \cellcolor{cyan!30}{74.25} & +6.39 \\
& F1 & 26.83 & - & \cellcolor{cyan!10}{56.32} & 39.46 & 13.8 & 48.5 & - & 5.48 & 54.2 & 30.06 & - & - & 40.85 & - & \cellcolor{cyan!30}{57.54} & +1.22 \\
& ARI & 17.85 & - & 48.43 & 36.95 & 4.64 & 47.33 & - & 1.75 & \cellcolor{cyan!10}{53.14} & 28.49 & - & - & 29.49 & - & \cellcolor{cyan!30}{55.64} & +2.5 \\
& CS & 44.14 & - & 62.89 & 57.22 & 20.36 & 66.93 & - & 21.61 & \cellcolor{cyan!10}{68.36} & 49.76 & - & - & 50.45 & - & \cellcolor{cyan!30}{74.99} & +6.63 \\ \midrule
\multirow{5}{*}{\textit{Photo}} & ACC & 41.75 & 36.44 & 43.32 & \cellcolor{cyan!10}{56.65} & 33.58 & 50.78 & 45.93 & 24.12 & 50 & 37.79 & 43.45 & - & 49.44 & - & \cellcolor{cyan!30}{60.59} & +3.94 \\
& NMI & 34.74 & 26.8 & 46.13 & \cellcolor{cyan!10}{50.67} & 22.43 & 50.19 & 42.48 & 1.91 & 46.72 & 34.93 & 10.52 & - & 46.08 & - & \cellcolor{cyan!30}{57.06} & +6.39 \\
& F1 & 38.65 & 21.6 & 39.7 & \cellcolor{cyan!10}{49.69} & 32.23 & 32.82 & 36.52 & 23.15 & 33.96 & 25.51 & 14.45 & - & 46.91 & - & \cellcolor{cyan!30}{61.25} & +11.56 \\
& ARI & 16.85 & 7.08 & 19.11 & \cellcolor{cyan!10}{25.79} & 6.94 & 23.32 & 19.25 & -0.59 & 24.33 & 14.07 & 2.82 & - & 22.74 & - & \cellcolor{cyan!30}{32.39} & +6.6 \\
& CS & 31.77 & 31.13 & 42.05 & \cellcolor{cyan!10}{46.47} & 21.15 & 46.36 & 40.35 & 2.03 & 42.28 & 38.77 & 25.86 & - & 42.05 & - & \cellcolor{cyan!30}{52.11} & +5.64 \\ \midrule
\multirow{5}{*}{\textit{Arts}} & ACC & 23.99 & \cellcolor{cyan!10}{53.4} & 51.12 & 49.36 & 31.18 & 41.05 & - & 21.2 & 32.48 & 30.51 & - & - & 33.93 & - & \cellcolor{cyan!30}{55.25} & +1.85 \\
& NMI & 11.9 & 38.72 & \cellcolor{cyan!10}{48.66} & 44.18 & 19.36 & 38.96 & - & 0.37 & 25.38 & 28.33 & - & - & 26.63 & - & \cellcolor{cyan!30}{50.64} & +1.98 \\
& F1 & 17 & 21.68 & \cellcolor{cyan!10}{35.64} & 35.46 & 23.88 & 32.17 & - & 22.81 & 25.31 & 26 & - & - & 24.34 & - & \cellcolor{cyan!30}{42.28} & +6.64 \\
& ARI & 6.08 & 32.06 & \cellcolor{cyan!10}{35.96} & 32.66 & 14.81 & 20.04 & - & -0.04 & 14.47 & 15.38 & - & - & 15.71 & - & \cellcolor{cyan!30}{38.81} & +2.85 \\
& CS & 10.87 & \cellcolor{cyan!30}{46.62} & 44.82 & 40.49 & 17.97 & 41.83 & - & 0.44 & 23.1 & 31.56 & - & - & 24.32 & - & \cellcolor{cyan!30}{46.62} & +0 \\ 
\bottomrule
\end{tabular}%
\end{footnotesize}
\end{table*}

\subsection{Experimental Setup}
\stitle{Datasets}
We evaluate our model on eight multimodal graphs from the Multimodal Attribute Graph Benchmark (MAGB)~\cite{yan2024graph}: {\textit{Movies}}, {\textit{Toys}}, {\textit{Grocery-S}}, {\textit{Grocery}}, {\textit{Photo}}, and {\textit{Arts}} from Amazon~\cite{ni2019justifying}; and {\textit{Reddit-S}} and {\textit{Reddit}} from RedCaps~\cite{desai2021redcaps}. Each node in these graphs is associated with textual and visual attributes. Table~\ref{tab:dataset_stats} presents the core statistics for each dataset. 

\stitle{Baselines}
We compare our method against a comprehensive set of baselines, which can be categorized as follows:
\begin{itemize}[leftmargin=*]
\item \textbf{AGC}: \texttt{MVGRL}~\cite{hassani2020contrastive}, \texttt{S3GC}~\cite{devvrit2022s3gc}, \texttt{DMoN}~\cite{tsitsulin2023graph}, and \texttt{Dink-Net}~\cite{liu2023dink}.

\item \textbf{MVC}: \texttt{EMVGC-LG}~\cite{wen2023efficient}, \texttt{MCLGF}~\cite{zhou2023learnable}, \texttt{GLSEF}~\cite{wang2023multi}, \texttt{FastMICE}~\cite{huang2023fast}, and \texttt{AEVC}~\cite{liu2024learn}.

\item \textbf{MVAGC}: \texttt{MCGC}~\cite{pan2021multi}, \texttt{MGCCN}~\cite{liu2022multilayer}, \texttt{LMGEC}~\cite{fettal2023simultaneous}, and \texttt{VGMGC}~\cite{chen2025variational}.
\end{itemize}
For all methods, we feed the learned node embeddings into a $K$-Means classifier to obtain the final cluster assignments. 
To ensure a fair comparison in the self-supervised setting, the supervised components of methods like \texttt{MVGRL} were omitted. 
Due to space constraints, details regarding datasets, baselines, settings, and evaluation metrics are deferred to Appendix~\ref{sec:additional-exp}.

\subsection{Clustering Performance Evaluation}
To validate the effectiveness of our proposed method, we conduct a comprehensive comparison against fourteen state-of-the-art baseline methods for multi-modal graph clustering. The evaluation is performed on eight diverse datasets, and we report five standard metrics following the experimental setup of \cite{devvrit2022s3gc, hassani2020contrastive}: Accuracy (Acc.), Normalized Mutual Information (NMI), F1-Score (F1), Adjusted Rand Index (ARI), and a Completeness Score (CS). The detailed results are presented in Table~\ref{tab:clustering_results}.

As shown in Table~\ref{tab:clustering_results}, our method, \algo{}, consistently achieves superior performance across all datasets and evaluation metrics. The improvements over the best-performing baseline are often substantial. For instance, on {\textit{Reddit-S}}, our method yields remarkable gains, improving Accuracy by $+10.53\%$ and ARI by $+10.35\%$ over the strongest competitor. Similarly, on {\textit{Grocery-S}}, we observe an absolute improvement of +9.0\% in Accuracy and $+9.31\%$ in ARI. The performance gains are consistent across all metrics, such as on the {\textit{Toys}} dataset, where \algo{} surpasses the best baseline by $+4.91\%$ in Accuracy, $+3.31\%$ in NMI, and $+6.53\%$ in F1-score. Even on datasets where the margin is smaller, like {\textit{Movies}}, our method establishes a new state-of-the-art across the board.

A key advantage of our method is its scalability and efficiency when handling large-scale graphs, a challenge where many existing methods falter. As indicated in Table~\ref{tab:clustering_results}, several competitive baselines (e.g., \texttt{MVGRL}, \texttt{MCLGF}, \texttt{MGCCN}, \texttt{VGMGC}) ran into Out of Memory (OOM) errors on the larger {\textit{Grocery}}, {\textit{Reddit}}, {\textit{Photo}}, and {\textit{Arts}} datasets. In contrast, our method successfully processes all datasets without issue, delivering top-tier performance. On the large {\textit{Reddit}} dataset, for example, \algo{} achieves an NMI of $74.25\%$, which is a significant $+6.39\%$ improvement over the best result from viable baselines. Similarly, on {\textit{Photo}}, our method boosts the F1-score to $61.25\%$, a $+11.56\%$ absolute improvement over the next best method.

\begin{table}[!t] 
 \centering 
 \caption{Ablation Study.} 
 \label{tab:ablation} 
 \vspace{-2ex}
 \setlength{\tabcolsep}{2pt} \renewcommand{\arraystretch}{0.8}
 \begin{footnotesize}
 \begin{tabular}{@{}ll|cccccc|c@{}} 
 \toprule 
 \multirow{2}{*}{Dataset} & & \multicolumn{6}{c|}{w/o} &  \multirow{2}{*}{\algo{}} \\ 
& & FDD & $\mathcal{L}_{\textnormal{mod}}$  & $\mathcal{L}_{\textnormal{nbr}}$  & AAS & $\mathcal{L}_{\textnormal{comm}}$  & HPS  & \\ \midrule 
\multirow{3}{*}{\em Movies}	&	ACC	&	25.58	&	25.67	&	23.84	&	\cellcolor{cyan!10}{25.84}	&	25.37	&	25.65	&	\cellcolor{cyan!30}{26.02}	\\
	&	NMI	&	22.37	&	21.59	&	20.62	&	\cellcolor{cyan!30}{22.47}	&	22.26	&	22.23	&	\cellcolor{cyan!10}{22.45}	\\
	&	F1	&	\cellcolor{cyan!30}{21.2}	&	\cellcolor{cyan!10}{20.94}	&	18.68	&	20.46	&	20.23	&	20.72	&	20.91	\\ \midrule
\multirow{3}{*}{\em Toys}	&	ACC	&	51.66	&	52.87	&	47.34	&	\cellcolor{cyan!10}{54.16}	&	53.67	&	53.82	&	\cellcolor{cyan!30}{55.75}	\\
	&	NMI	&	50.22	&	51.73	&	48.56	&	51.34	&	51.46	&	\cellcolor{cyan!10}{51.89}	&	\cellcolor{cyan!30}{52.37}	\\
	&	F1	&	48.28	&	48.43	&	43.28	&	\cellcolor{cyan!10}{50.68}	&	48.42	&	47.53	&	\cellcolor{cyan!30}{50.79}	\\ \midrule
\multirow{3}{*}{\em Grocery-S}	&	ACC	&	52.09	&	\cellcolor{cyan!10}{52.59}	&	51.51	&	50.77	&	51.96	&	52	&	\cellcolor{cyan!30}{53.41}	\\
	&	NMI	&	52.57	&	\cellcolor{cyan!10}{54.12}	&	\cellcolor{cyan!30}{54.88}	&	51.01	&	53.36	&	53.32	&	{54}	\\
	&	F1	&	44.54	&	\cellcolor{cyan!30}{45.5}	&	44.74	&	43.97	&	44.12	&	43.12	&	\cellcolor{cyan!10}{45.42}	\\ \midrule
\multirow{3}{*}{\em Grocery}	&	ACC	&	37.86	&	\cellcolor{cyan!10}{38.38}	&	34.89	&	35.7	&	37.36	&	35.73	&	\cellcolor{cyan!30}{38.52}	\\
	&	NMI	&	45.58	&	46.14	&	44.16	&	44.23	&	\cellcolor{cyan!10}{46.49}	&	46.12	&	\cellcolor{cyan!30}{46.67}	\\
	&	F1	&	33.11	&	\cellcolor{cyan!10}{32.4}	&	28.03	&	30.21	&	31.81	&	28.99	&	\cellcolor{cyan!30}{33.62}	\\ \midrule
\multirow{3}{*}{\em Reddit-S}	&	ACC	&	85.26	&	83.18	&	74.69	&	88.83	&	\cellcolor{cyan!10}{89.16}	&	83.42	&	\cellcolor{cyan!30}{89.66}	\\
	&	NMI	&	85.25	&	85.01	&	82.09	&	\cellcolor{cyan!30}{88.43}	&	87.46	&	86.08	&	\cellcolor{cyan!10}{88.24}	\\
	&	F1	&	75.46	&	70.5	&	63.29	&	77.99	&	\cellcolor{cyan!10}{81.42}	&	70.39	&	\cellcolor{cyan!30}{81.98}	\\ \midrule
\multirow{3}{*}{\em Reddit}	&	ACC	&	61.04	&	55.4	&	49.27	&	59.76	&	\cellcolor{cyan!10}{61.79}	&	56.13	&	\cellcolor{cyan!30}{62.87}	\\
	&	NMI	&	66.54	&	65.86	&	65.8	&	\cellcolor{cyan!10}{74.08}	&	73.5	&	72.38	&	\cellcolor{cyan!30}{74.25}	\\
	&	F1	&	\cellcolor{cyan!10}{56.61}	&	53.32	&	45	&	54.5	&	56.23	&	51.85	&	\cellcolor{cyan!30}{57.54}	\\ \midrule
\multirow{3}{*}{\em Photo}	&	ACC	&	59.28	&	59.23	&	58.41	&	57.38	&	\cellcolor{cyan!10}{59.34}	&	58.26	&	\cellcolor{cyan!30}{60.59}	\\
	&	NMI	&	55.5	&	55.03	&	56.64	&	53.64	&	56.44	&	\cellcolor{cyan!30}{57.14}	&	\cellcolor{cyan!10}{57.06}	\\
	&	F1	&	\cellcolor{cyan!10}{60.64}	&	56.58	&	58.35	&	57.4	&	59.92	&	56.48	&	\cellcolor{cyan!30}{61.25}	\\ \midrule
\multirow{3}{*}{\em Arts}	&	ACC	&	52.48	&	45.59	&	40.88	&	53.62	&	\cellcolor{cyan!10}{55.23}	&	51.19	&	\cellcolor{cyan!30}{55.25}	\\
	&	NMI	&	47.32	&	44.58	&	37.41	&	48.51	&	\cellcolor{cyan!30}{50.64}	&	48.62	&	\cellcolor{cyan!30}{50.64}	\\
	&	F1	&	42.07	&	39.36	&	34.05	&	40.54	&	\cellcolor{cyan!30} 42.3	&	\cellcolor{cyan!30} 42.3	&	42.28	\\
\bottomrule 
 \end{tabular}%
 \end{footnotesize}
 \vspace{-2ex}
 \end{table}

\subsection{Ablation Study}

\begin{figure}[!t]
    \centering
    \subfloat{
    \includegraphics[width=0.42\columnwidth]{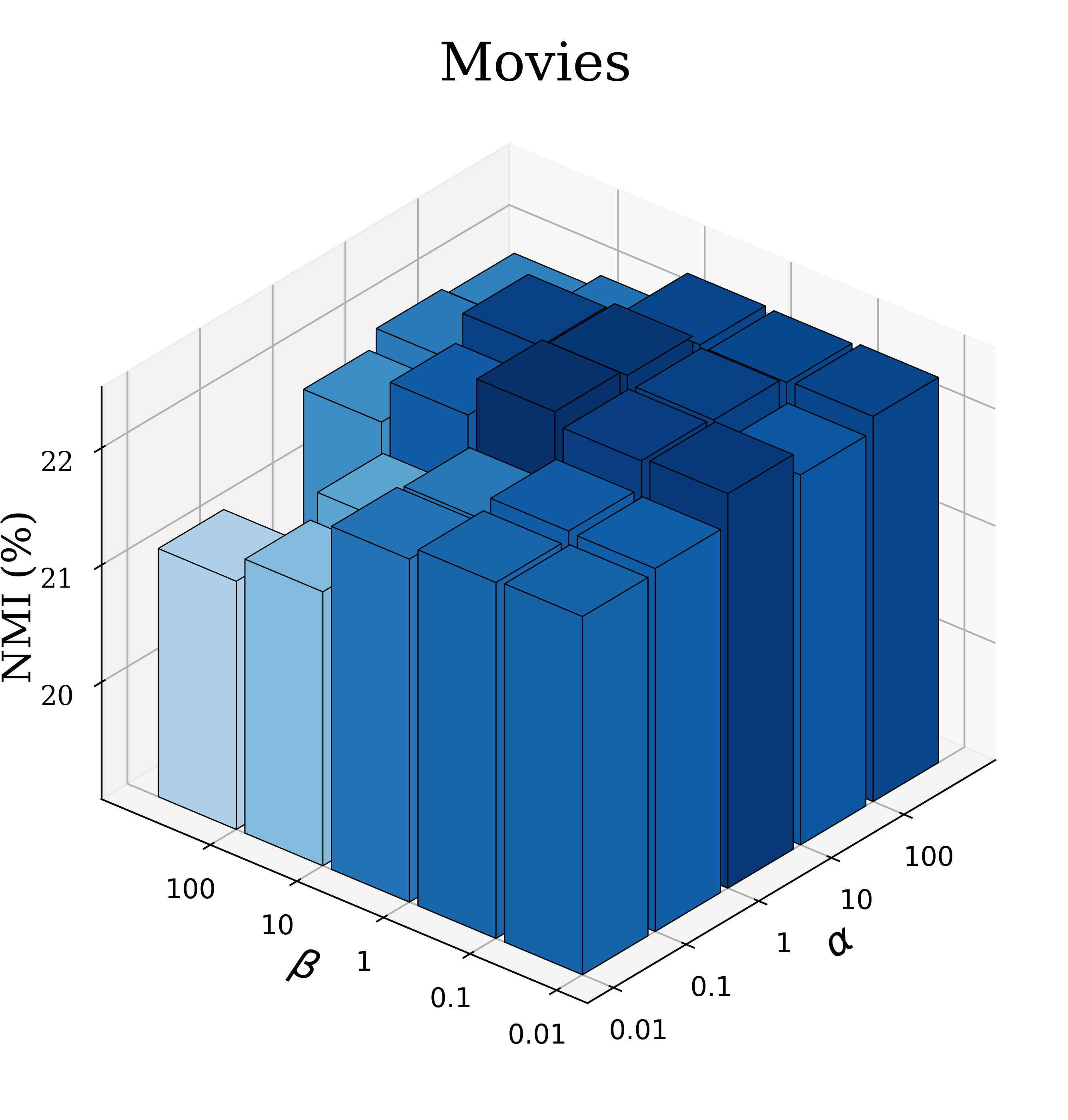}
    }
    \hspace{2mm}
    \subfloat{
    \includegraphics[width=0.42\columnwidth]{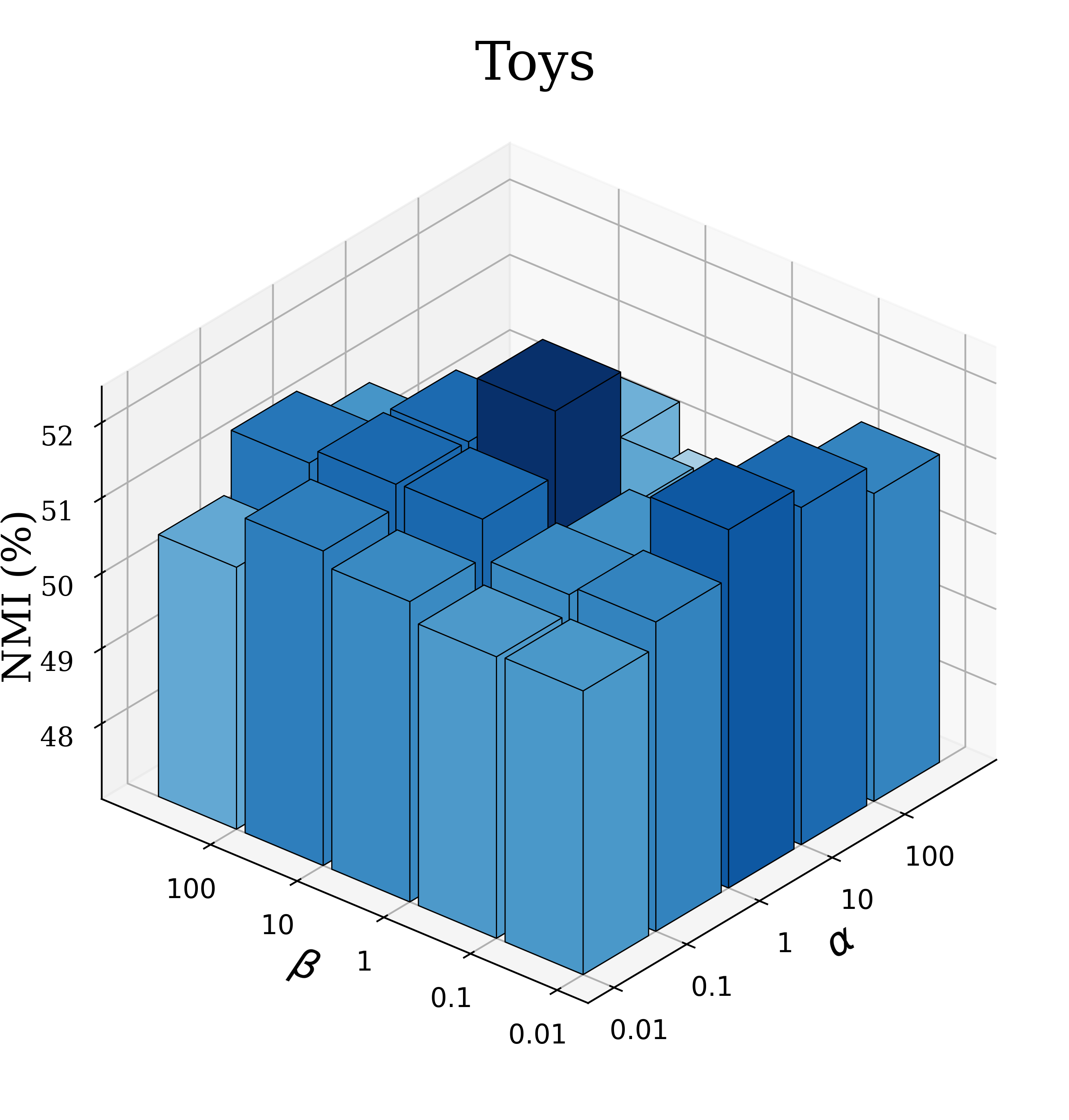}
    }
    \vspace{-16pt}
    \subfloat{
    \includegraphics[width=0.42\columnwidth]{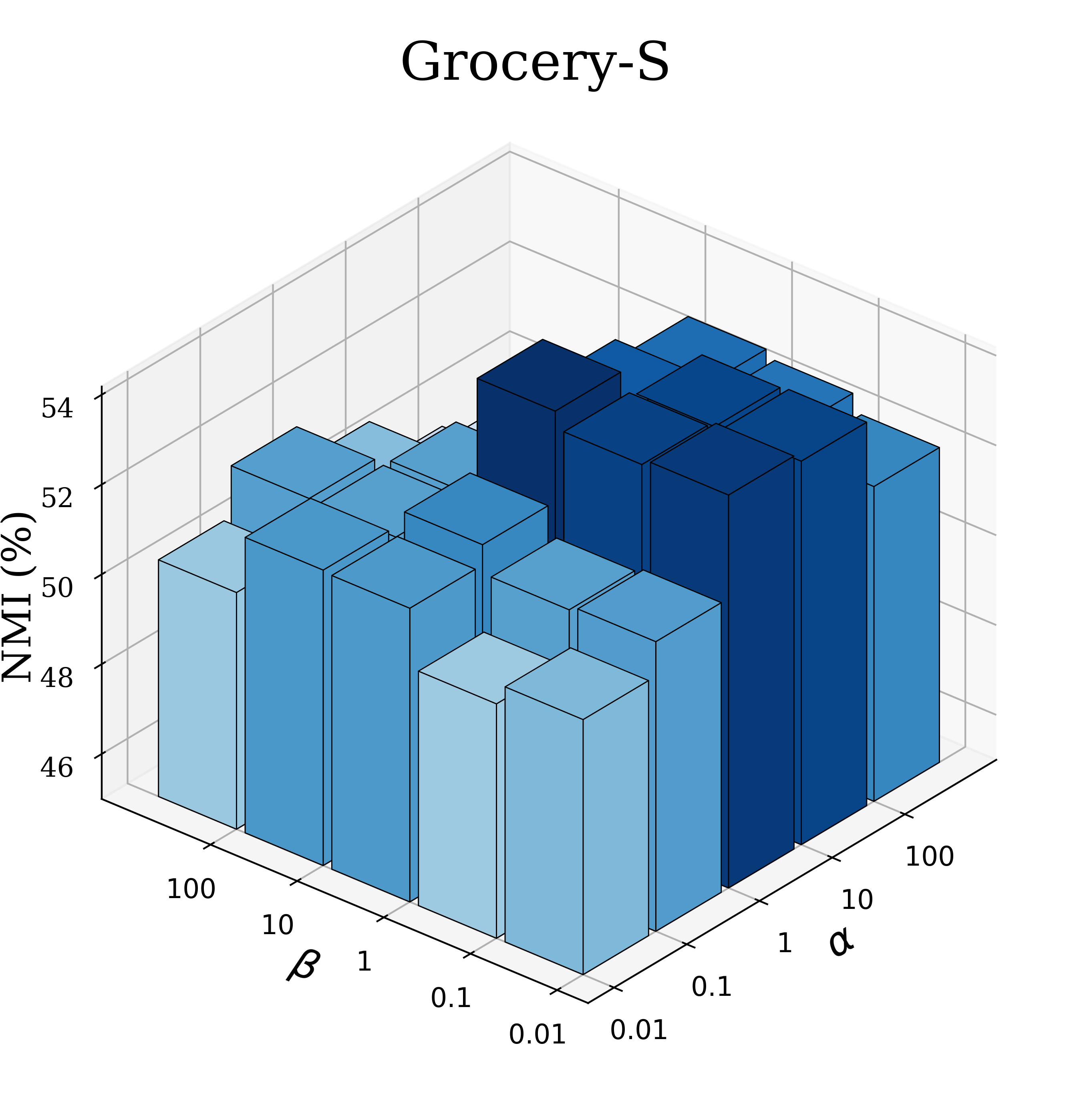}
    }
    \hspace{2mm}
    \subfloat{
    \includegraphics[width=0.42\columnwidth]{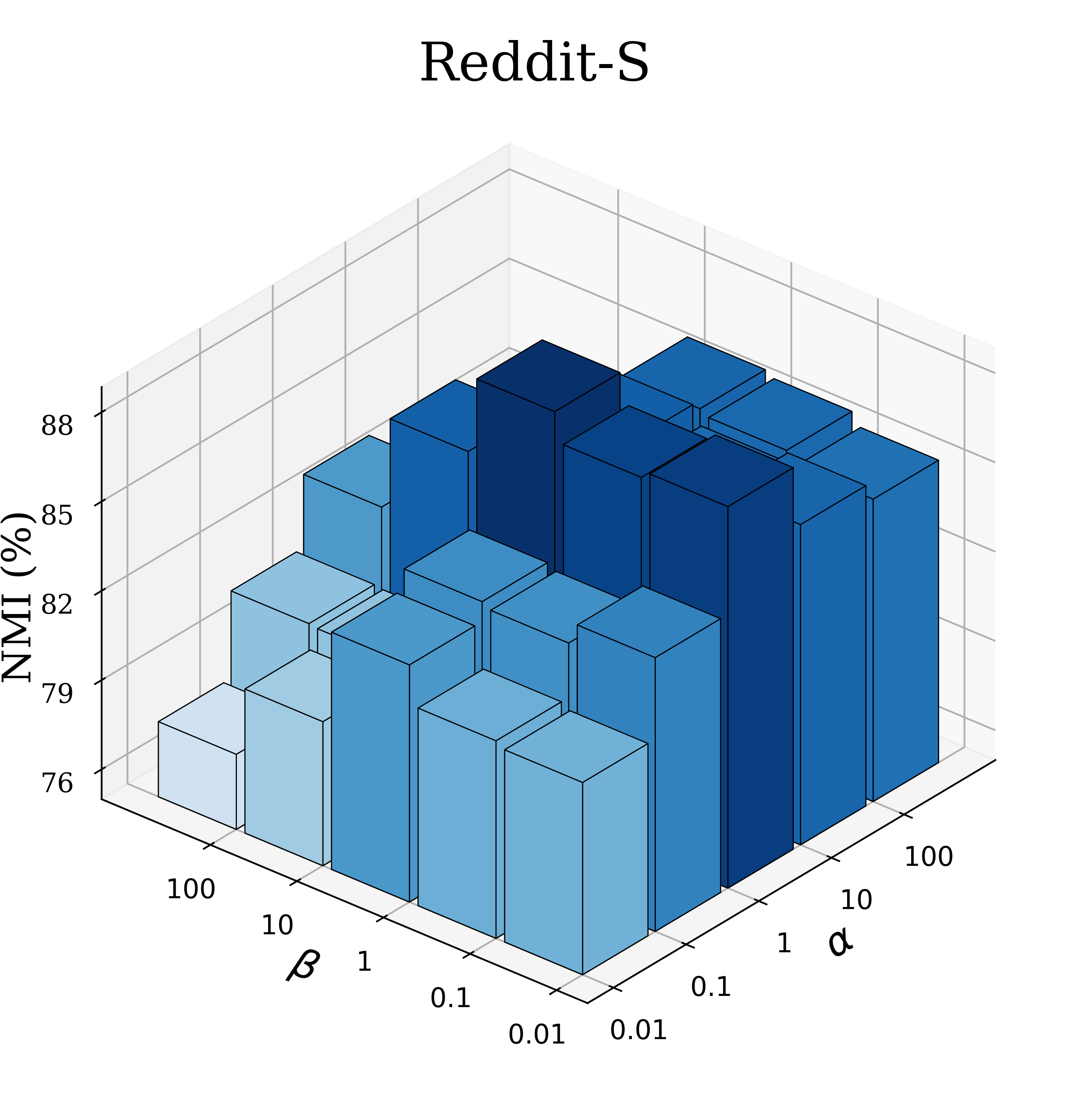}
    }
    \vspace{-16pt}
    \caption{Parameter analysis of $\alpha$ and $\beta$.}
    \vspace{-2ex}
    \label{fig:params_alpha_beta}
\end{figure}

\begin{figure}[ht]
\centering
\begin{small}
\begin{tikzpicture}
    \begin{axis}[
        hide axis,
        xmin=0, xmax=5, ymin=0, ymax=1,
        legend columns=5,
        legend style={
            at={(0.5,1.0)},
            anchor=south,
            draw=none,
            font=\small,
            /tikz/every even column/.append style={column sep=0.5cm}
        }
    ]
    \addlegendimage{line width=0.25mm,mark size=2.5pt,color=myblue2, mark=*}
    \addlegendentry{ACC}
    
   \addlegendimage{line width=0.25mm,mark size=2pt,color=cyan, mark=square*}
    \addlegendentry{NMI}

    \addlegendimage{line width=0.25mm,mark size=2.5pt,color=myorange, mark=triangle*}
    \addlegendentry{F1}

    \addlegendimage{line width=0.25mm,mark size=2.5pt,color=myred, mark=diamond*}
    \addlegendentry{ARI}

    \addlegendimage{line width=0.25mm,mark size=2.5pt,color=mygreen3, mark=pentagon*}
    \addlegendentry{CS}
    \end{axis}
\end{tikzpicture}
\begin{tikzpicture}
    \begin{axis}[
        title={Movies},
        xmin=5, xmax=30,
        xtick={5,10,15,20,25,30},
        grid=major,
        width=0.26\textwidth,
        height=0.20\textwidth,
        clip=false,
    ]
    \addplot[line width=0.25mm,mark size=2.5pt,myblue2, mark=*] table[x=num_layers, y expr=\thisrow{acc}*100] {
        num_layers acc
        5 0.2543785988483685 
        10 0.260196737
        15 0.25923704414587334
        20 0.258637236084453 
        25 0.258637236084453 
        30 0.2567178502879079
    };
    \addplot[line width=0.25mm,mark size=2pt,cyan, mark=square*] table[x=num_layers, y expr=\thisrow{nmi}*100] {
        num_layers nmi
        5 0.2226210990866596 
        10 0.2245119529
        15 0.22468909246702976
        20 0.22324190751892437 
        25 0.2243746961469595 
        30 0.22307256422139463
    };
    \addplot[line width=0.25mm,mark size=2.5pt,myorange, mark=triangle*] table[x=num_layers, y expr=\thisrow{f1}*100] {
        num_layers f1
        5 0.2061297734310096 
        10 0.2091348041
        15 0.20828297568617443
        20 0.2097772737616786 
        25 0.20912018543189137 
        30 0.20706942698170305
    };
    \addplot[line width=0.25mm,mark size=2.5pt,myred, mark=diamond*] table[x=num_layers, y expr=\thisrow{adj}*100] {
        num_layers adj
        5 0.07880887330310965 
        10 0.08120155651
        15 0.07968829365983726
        20 0.07853456044150718 
        25 0.07931646290435763 
        30 0.07917887674712477
    };
    \addplot[line width=0.25mm,mark size=2.5pt,mygreen3, mark=pentagon*] table[x=num_layers, y expr=\thisrow{cs}*100] {
        num_layers cs
        5 0.1997134198521514 
        10 0.2015199527
        15 0.2021912231879379
        20 0.20042143387561914 
        25 0.20136440647902884 
        30 0.20005110862207953
    };
    \end{axis}
\end{tikzpicture}\hspace{4mm}
\begin{tikzpicture}
    \begin{axis}[
        title={Toys},
        xmin=5, xmax=30,
        xtick={5,10,15,20,25,30},
        grid=major,
        width=0.26\textwidth,
        height=0.20\textwidth,
        clip=false,
    ]
    \addplot[line width=0.25mm,mark size=2.5pt,myblue2, mark=*] table[x=num_layers, y expr=\thisrow{acc}*100] {
        num_layers acc
        5 0.5371345735684948 
        10 0.5575259725
        15 0.5297898043005557
        20 0.5298381251510027 
        25 0.5414834501087219 
        30 0.5325440927760329
    };
    \addplot[line width=0.25mm,mark size=2pt,cyan, mark=square*] table[x=num_layers, y expr=\thisrow{nmi}*100] {
        num_layers nmi
        5 0.515355895305687 
        10 0.5236729838
        15 0.5104766123553826
        20 0.5164832591816017 
        25 0.5134563501252121 
        30 0.5182841401898511
    };
    \addplot[line width=0.25mm,mark size=2.5pt,myorange, mark=triangle*] table[x=num_layers, y expr=\thisrow{f1}*100] {
        num_layers f1
        5 0.4806526643425074 
        10 0.5078897502
        15 0.476162556131466
        20 0.484993752551134 
        25 0.4976291559828561 
        30 0.4846702453328573
    };
    \addplot[line width=0.25mm,mark size=2.5pt,myred, mark=diamond*] table[x=num_layers, y expr=\thisrow{adj}*100] {
        num_layers adj
        5 0.3685676739834738 
        10 0.3853123111
        15 0.3631960656692874
        20 0.37047641128061876 
        25 0.3684155210353081 
        30 0.3707948891738041
    };
    \addplot[line width=0.25mm,mark size=2.5pt,mygreen3, mark=pentagon*] table[x=num_layers, y expr=\thisrow{cs}*100] {
        num_layers cs
        5 0.5062739936891818 
        10 0.514716908
        15 0.501507738395152
        20 0.5058568983269038 
        25 0.5038981723868193 
        30 0.5082334124323064
    };
    \end{axis}
\end{tikzpicture}\hspace{4mm}

\begin{tikzpicture}
    \begin{axis}[
        title={Grocery-S},
        xmin=5, xmax=30,
        xtick={5,10,15,20,25,30},
        grid=major,
        width=0.26\textwidth,
        height=0.20\textwidth,
        clip=false,
    ]
    \addplot[line width=0.25mm,mark size=2.5pt,myblue2, mark=*] table[x=num_layers, y expr=\thisrow{acc}*100] {
        num_layers acc
        5 0.5319198781773457 
        10 0.534145484362188
        15 0.5327984069345203
        20 0.5292257233220101 
        25 0.5119479910975753 
        30 0.5235445706922807
    };
    \addplot[line width=0.25mm,mark size=2pt,cyan, mark=square*] table[x=num_layers, y expr=\thisrow{nmi}*100] {
        num_layers nmi
        5 0.5363577518265122 
        10 0.54004622602218
        15 0.541434406912608
        20 0.5389470808277252 
        25 0.5407651032430291 
        30 0.5317545271539393
    };
    \addplot[line width=0.25mm,mark size=2.5pt,myorange, mark=triangle*] table[x=num_layers, y expr=\thisrow{f1}*100] {
        num_layers f1
        5 0.45434614729762446 
        10 0.454163672917032
        15 0.4532906481974467
        20 0.4518677133159786 
        25 0.44068463760704935 
        30 0.4358018255409032
    };
    \addplot[line width=0.25mm,mark size=2.5pt,myred, mark=diamond*] table[x=num_layers, y expr=\thisrow{adj}*100] {
        num_layers adj
        5 0.408043969887663 
        10 0.413648146128375
        15 0.41528280019166547
        20 0.4086834110493479 
        25 0.3979516547218991 
        30 0.4040084305014184
    };
    \addplot[line width=0.25mm,mark size=2.5pt,mygreen3, mark=pentagon*] table[x=num_layers, y expr=\thisrow{cs}*100] {
        num_layers cs
        5 0.5202507349380657 
        10 0.523924444713753
        15 0.525142063072234
        20 0.5228826037812914 
        25 0.5244680020839725 
        30 0.516787505352015
    };
    \end{axis}
\end{tikzpicture}\hspace{4mm}
\begin{tikzpicture}
    \begin{axis}[
        title={Reddit-S},
        xmin=5, xmax=30,
        xtick={5,10,15,20,25,30},
        grid=major,
        width=0.26\textwidth,
        height=0.20\textwidth,
        clip=false,
    ]
    \addplot[line width=0.25mm,mark size=2.5pt,myblue2, mark=*] table[x=num_layers, y expr=\thisrow{acc}*100] {
        num_layers acc
        5 0.8830376242607273 
        10 0.8966276582358123
        15 0.8717125959481565
        20 0.8750471876179691 
        25 0.8847363785076129 
        30 0.877815527872153
    };
    \addplot[line width=0.25mm,mark size=2pt,cyan, mark=square*] table[x=num_layers, y expr=\thisrow{nmi}*100] {
        num_layers nmi
        5 0.8794615092704682 
        10 0.8824435667422431
        15 0.8711615668590622
        20 0.8720804271320921 
        25 0.8809761666298724 
        30 0.8728014945338846
    };
    \addplot[line width=0.25mm,mark size=2.5pt,myorange, mark=triangle*] table[x=num_layers, y expr=\thisrow{f1}*100] {
        num_layers f1
        5 0.7740850109009976 
        10 0.8197597706231031
        15 0.7659736430636975
        20 0.7693682904978052 
        25 0.7779567657362805 
        30 0.7757264493345452
    };
    \addplot[line width=0.25mm,mark size=2.5pt,myred, mark=diamond*] table[x=num_layers, y expr=\thisrow{adj}*100] {
        num_layers adj
        5 0.8748895930335339 
        10 0.8806704829824539
        15 0.8524900740725456
        20 0.8619279256315884 
        25 0.8669314997693558 
        30 0.862698576180516
    };
    \addplot[line width=0.25mm,mark size=2.5pt,mygreen3, mark=pentagon*] table[x=num_layers, y expr=\thisrow{cs}*100] {
        num_layers cs
        5 0.8778501730587533 
        10 0.8802166212082612
        15 0.8682201186523378
        20 0.8701600457338478 
        25 0.879855197959224 
        30 0.8711981696096006
    };
    \end{axis}
\end{tikzpicture}\hspace{4mm}
\end{small}
\vspace{-2ex}
\caption{Parameter analysis of $T$.}
\label{fig:params_t}
\end{figure}
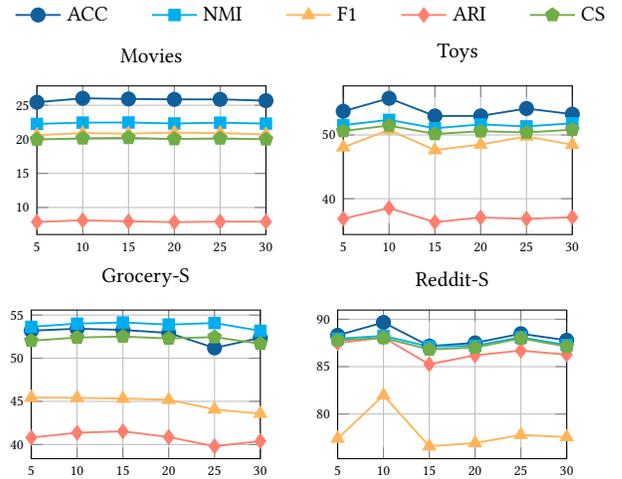

To validate the effectiveness of our key components, we conduct a comprehensive ablation study, with results summarized in Table~\ref{tab:ablation}. The consistent performance degradation across all variants underscores the integral contribution of each component to our full model, \algo{}. We first evaluated the impact of our proposed feature-domain denoising (FDD) term. Removing this term from the optimization objective (\textbf{w/o FDD}) leads to a notable performance drop, e.g., the ACC on \textit{Toys} decreases from $55.75\%$ to $51.66\%$, which demonstrates that explicitly regularizing noise in the feature space is crucial for learning robust representations.

Next, we analyze the contributions of our contrastive objectives. Removing the cross-modal contrastive loss (\textbf{w/o $\mathcal{L}_{\textnormal{mod}}$}) impairs performance, confirming that aligning representations from different modalities into a common space is essential for effective fusion. The most critical component is evidently the attribute-aware neighbor contrastive loss; its removal (\textbf{w/o $\mathcal{L}_{\textnormal{nbr}}$}) causes the most substantial performance degradation across all datasets (e.g., ACC drops from $62.87\%$ to $49.27\%$ on \textit{Reddit}), highlighting its central role in learning discriminative embeddings. To further validate its design, removing only the attribute-aware sampling strategy (\textbf{w/o AAS}) also hurts performance, demonstrating that leveraging attribute similarity for positive sampling is superior to using topological proximity alone.
Finally, we assess the community contrastive loss. Removing it (\textbf{w/o $\mathcal{L}_{\textnormal{comm}}$}) degrades results, confirming its utility in refining the final cluster structure by enhancing inter-cluster separation. Similarly, isolating the effect of hard positive sampling (\textbf{w/o HPS}) by ablating its mechanism for filtering false positives from pseudo-clusters also lowers performance. This proves the value of this targeted sampling strategy in promoting compactness within clusters.

\subsection{Parameter Analysis}

We now analyze the sensitivity of \algo{} to its primary hyperparameters: the balancing coefficients $\alpha$ and $\beta$, and the number $T$ of iterations. Due to space limits, the related analysis for the sampling threshold $\theta$ is deferred to Appendix~\ref{sec:appendix_params}.

\stitle{Analyses of $\alpha$ and $\beta$}
We first investigate the impact of $\alpha$ and $\beta$, which balance the contributions of the primary loss terms. We vary both parameters within the range of $\{0.01, 0.1, 1, 10, 100\}$. As shown in Figure~\ref{fig:params_alpha_beta}, the model is relatively stable but consistently achieves peak NMI scores across all tested datasets when $\alpha$ and $\beta$ are balanced at $1.0$. Extreme values for either parameter, which cause one loss term to dominate the other, tend to degrade performance. This highlights the importance of balancing their respective contributions. Thus, we set $\alpha=1.0$ and $\beta=1.0$ in all experiments.

\stitle{Analysis of $T$}
Next, we examine the effect of the number of message passing layers, $T$, by varying it from 5 to 30. The results in Figure~\ref{fig:params_t} show a consistent trend across the datasets. Performance generally improves as $T$ increases from a small value, peaks around $T=10$, and then either slightly declines or plateaus. This pattern suggests that $T=10$ provides a sufficient receptive field to aggregate neighborhood information effectively. Fewer layers may be inadequate for capturing long-range dependencies, while more layers risk introducing the over-smoothing issue, which makes node representations less distinguishable. Therefore, we use $T=10$ as the default.

\section{Conclusion}
In this paper, we tackled the problem of clustering on MMAGs, where traditional multi-view methods fall short due to the low correlation and high noise levels characteristic of attributes from large pre-trained models. We proposed \algo{}, a novel scheme that introduces a dual graph filtering mechanism to denoise attributes at both the node and feature levels. This is complemented by a tri-cross contrastive training strategy that effectively learns robust and discriminative representations. Extensive experiments validated that \algo{} significantly and consistently outperforms a wide array of state-of-the-art baselines on eight real-world datasets. For future work, we aim to investigate the applicability of \algo{} to other downstream multimodal graph learning tasks.

\begin{acks}
This work is partially supported by the National Natural Science Foundation of China (No. 62302414), the Hong Kong RGC ECS grant (No. 22202623), YCRG (No. C2003-23Y), RGC RIF R1015-23, the Huawei Gift Fund, and Guangdong and Hong Kong Universities ``1+1+1'' Joint Research Collaboration Scheme, project No.: 2025A0505000002.
\end{acks}

\balance

\bibliographystyle{ACM-Reference-Format}
\bibliography{main}

\appendix
\section{Notations}\label{sec:notations}
\begin{table}[ht]
\centering
\caption{Frequently used notations.}
\label{tab:notations}
\vspace{-2ex}
\begin{small}
\begin{tabular}{l|p{0.8\columnwidth}}
\toprule
{Symbol} & {Description} \\
\midrule
$\V, \EDG$ & The set of nodes and edges, respectively. \\
$\XM^{(i)}, \ZM^{(i)}$ & The initial node attribute matrix and the projected node embedding matrix for the $i$-th modality, respectively. \\
$\HM$ & The final, fused node representations learned. \\
$d_i, d$ & The dimensionality of the original node attribute embeddings for the $i$-th modality and hidden representations, respectively. \\
$\LM$ & Symmetric normalized Laplacian of $\G$. \\
$\SM^{(i)}$ & Symmetric normalized feature similarity matrix for the $i$-th modality. \\
$\alpha, \beta$ & Balancing hyperparameters for the objective function in Eq.~\eqref{eq:obj}. \\
$T$ & The number of message passing layers. \\
$\theta$ & The hard positive sampling threshold in the community contrastive loss. \\
$K$ & The number of clusters for the clustering task. \\
\bottomrule
\end{tabular}
\end{small}
\end{table}

\section{Iterative Computation of $\HM$}\label{sec:comp-H}

\begin{algorithm}[b]
\caption{Computing node representations $\HM$}
\label{alg:compute_H}
\SetAlgoLined
\KwIn{
    Initial features $\{\XM^{(i)}\}_{i=1}^m$, Layers $\{\WM^{(i)}\}_{i=1}^m$, Adjacency matrix $\NAM$,
    Hyperparameters $\alpha, \beta$, Truncation depth $T$;
}
\KwOut{Final node representations $\HM$;}

\tcp{1. Compute initial matrices}
$\ZM \leftarrow \textsf{combine}(\{\XM^{(i)}\WM^{(i)}\}_{i=1}^m)$\;
$\SM_{\text{sum}} \leftarrow \mathbf{0}_{d \times d}$\;
\For{$i = 1$ \KwTo $m$}{
    $\SM^{(i)} \leftarrow \text{SymmetricSoftmax}(\ZM^{(i)})$ \tcp*{Using Eq.~\eqref{eq:Si}}
    $\SM_{\text{sum}} \leftarrow \SM_{\text{sum}} + \SM^{(i)}$\;
}
$\bar{\SM} \leftarrow \frac{1}{m} \SM_{\text{sum}}$\;

\tcp{2. Compute filter polynomials}
$\FM_L \leftarrow \sum_{t=0}^T \left(\frac{\alpha}{\alpha+1}\NAM\right)^t$\;
$\FM_R \leftarrow \sum_{t=0}^T \left(\frac{\beta}{\beta+1}\bar{\SM}\right)^t$\;

\tcp{3. Apply filters to get final representations}
$\HM \leftarrow \frac{1}{(\alpha+1)(\beta+1)} \cdot \FM_L \cdot \ZM \cdot \FM_R$\;
\Return{$\HM$}\;
\end{algorithm}

Algorithm~\ref{alg:compute_H} outlines the computation of the final node representations $\HM$.
First, the algorithm initializes the combined feature matrix $\ZM$ (line 1) and the average feature-domain shift operator $\bar{\SM}$ (lines 2-7), where the latter is derived by averaging the modality-specific shift matrices $\SM^{(i)}$ from Eq.~\eqref{eq:Si}. Next, it constructs the node-domain filter $\FM_L$ and the feature-domain filter $\FM_R$ as truncated polynomial expansions (lines 8-9). Finally, the output representation $\HM$ is generated by applying these filters to $\ZM$ (i.e., $\FM_L \ZM \FM_R$), followed by a normalization step (line 10), before the result is returned (line 11).

\section{Additional Experimental Details}\label{sec:additional-exp}
All experiments are conducted on a Linux machine with an NVIDIA Ampere A100 GPU (80GB memory), AMD EPYC 7513 CPUs (2.6 GHz), and 1TB RAM.

\begin{table*}[h]
\centering
\caption{Additional details of the MMAG datasets used.}
\label{tab:dataset_additional}
\begin{small}
\begin{tabular}{lllllll}
\toprule
\multirow{2}{*}{Dataset} & \multicolumn{3}{c}{Text Feature} & \multicolumn{3}{c}{Image Feature} \\ 
\cmidrule(lr){2-4} \cmidrule(lr){5-7} 
& PLM used & \# Dim. & Name & PVM used & \# Dim. & Name \\ \midrule
\textit{Movies}         & Gemma~\cite{team2024gemma}       & 3,072        & Movies\_gemma\_7b\_256\_mean.npy & CLIP ViT~\cite{radford2021learning}       & 768        & Movies\_openai\_clip-vit-large-patch14.npy \\
\textit{Toys}           & Gemma       & 3,072        & Toys\_gemma\_7b\_256\_mean.npy & CLIP ViT       & 768        & Toys\_openai\_clip-vit-large-patch14.npy \\
\textit{Grocery-S}      & Gemma       & 3,072        & GroceryS\_gemma\_7b\_256\_mean.npy & CLIP ViT       & 768        & GroceryS\_openai\_clip-vit-large-patch14.npy \\
\textit{Grocery}        & Gemma       & 3,072        & Grocery\_gemma\_7b\_256\_mean.npy & CLIP ViT       & 768        & Grocery\_openai\_clip-vit-large-patch14.npy \\
\textit{Reddit-S}       & Gemma       & 3,072        & RedditS\_gemma\_7b\_64\_mean.npy & CLIP ViT       & 768        & RedditS\_openai\_clip-vit-large-patch14.npy \\
\textit{Reddit}         & Gemma       & 3,072        & Reddit\_gemma\_7b\_100\_mean.npy & CLIP ViT       & 768        & Reddit\_openai\_clip-vit-large-patch14.npy \\
\textit{Photo}          & Gemma       & 3,072        & Photo\_gemma\_7b\_256\_mean.npy & CLIP ViT       & 768        & Photo\_openai\_clip-vit-large-patch14.npy \\
\textit{Arts}           & Gemma       & 3,072        & Arts\_gemma\_7b\_256\_mean.npy & CLIP ViT       & 768        & Arts\_openai\_clip-vit-large-patch14.npy \\
\bottomrule
\end{tabular}
\end{small}
\end{table*}
\subsection{Additional Dataset Details}\label{sec:additional-datasets}

\stitle{Dataset Descriptions and Preprocessing}
The eight datasets used in our study originate from two main sources. The six e-commerce networks ({\textit{Movies}}, {\textit{Toys}}, {\textit{Grocery-S}}, {\textit{Grocery}}, {\textit{Photo}}, and {\textit{Arts}}) are constructed from the Amazon dataset~\cite{ni2019justifying}. In these graphs, nodes represent products, and an edge connects two nodes if the corresponding products are frequently co-purchased or co-viewed. Node labels are derived from product categories. The two social networks ({\textit{Reddit-S}} and {\textit{Reddit}}) are built from the RedCaps dataset~\cite{desai2021redcaps}. Here, nodes represent individual posts, and an edge links two posts if the same user has commented on both. Node labels correspond to the subreddit from which the post originated. For all graphs, we standardized the structure by converting them to undirected graphs and removing any self-loops.
For node features, we use pre-extracted textual embeddings and visual embeddings provided by MAGB. All feature vectors were $L_2$ normalized, with any `NaN' values imputed using the feature mean. 

\stitle{Node Feature Details}
We use pre-extracted node features from the Multimodal Attribute Graph Benchmark (MAGB)~\cite{yan2024graph}. MAGB provides a rich repository of features generated by various state-of-the-art Pre-trained Language Models (PLMs) and Pre-trained Vision Models (PVMs). The available PLMs include MiniLM~\cite{wang2020minilmv2}, Llama 2~\cite{touvron2023llama}, Mistral~\cite{jiang2023mistral7b}, and Gemma~\cite{team2024gemma}, while the PVMs include CLIP ViT~\cite{radford2021learning}, ConvNeXt~\cite{woo2023convnext}, SwinV2~\cite{liu2022swin}, and DinoV2 ViT~\cite{oquab2023dinov2}.

Based on the strong baseline performance reported in the MAGB paper~\cite{yan2024graph}, we selected features processed by Gemma for the textual modality and CLIP ViT for the visual modality. Table~\ref{tab:dataset_additional} provides specific details about these features. All data and features are accessible from \url{https://huggingface.co/Sherirto/MAG}.

\subsection{Baseline Descriptions}\label{sec:addtional-baselines}
\begin{table}[ht]
\centering
\caption{Links to code of baseline methods.}
\label{tab:links_code}
\begin{small}
\begin{tabular}{l|p{0.7\columnwidth}}
\toprule
{Method} & {Link to code} \\
\midrule
MVGRL         & \url{https://github.com/kavehhassani/mvgrl}         \\
S3GC          & \url{https://github.com/devvrit/S3GC}         \\
DMoN          & \url{https://github.com/google-research/google-research/tree/master/graph_embedding/dmon}         \\
Dink-Net      & \url{https://github.com/yueliu1999/Dink-Net}         \\
EMVGC-LG      & \url{https://github.com/wenyiwy99/EMVGC-LG}         \\
MCLGF         & \url{http://doctor-nobody.github.io/codes/MCLGF.zip}         \\
GLSEF         & \url{https://github.com/pengleiwangcn/GLSEF_code}         \\
FastMICE      & \url{https://github.com/huangdonghere/FastMICE}         \\
AEVC          & \url{https://github.com/Tracesource/AEVC}         \\
MCGC          & \url{https://github.com/Panern/MCGC}         \\
MGCCN         & \url{https://github.com/Lliang97/Multilayer-Clutering-Network}         \\
LMGEC         & \url{https://github.com/chakib401/LMGEC}         \\
VGMGC         & \url{https://github.com/cjpcool/VGMGC}         \\
\bottomrule
\end{tabular}
\end{small}
\end{table}

We briefly describe the core technical contribution of each baseline method used in our comparison. We also include $K$-Means as a fundamental clustering algorithm applied to the datasets' features.

\begin{itemize}[leftmargin=*]
    \item \texttt{MVGRL}~\cite{hassani2020contrastive}: Contrasts local (first-order neighbors) and global (graph diffusion) structural views to learn node representations.

    \item \texttt{S3GC}~\cite{devvrit2022s3gc}: A scalable approach using contrastive learning with a lightweight GNN encoder for large graphs.
    
    \item \texttt{DMoN}~\cite{tsitsulin2023graph}: Learns soft cluster assignments by directly optimizing a differentiable modularity objective.

    \item \texttt{Dink-Net}~\cite{liu2023dink}: An end-to-end framework unifying representation learning and clustering with an adversarial loss.

    \item \texttt{EMVGC-LG}~\cite{wen2023efficient}: An anchor-based method that jointly optimizes anchor construction and graph learning across views.

    \item \texttt{MCLGF}~\cite{zhou2023learnable}: Learns a shared graph filter across all views to produce cluster-friendly representations.

    \item \texttt{GLSEF}~\cite{wang2023multi}: Preserves local topology on the Grassmann manifold while allocating global weights to different views.

    \item \texttt{FastMICE}~\cite{huang2023fast}: Constructs and partitions view-sharing bipartite graphs using a hybrid early-late fusion strategy.

    \item \texttt{AEVC}~\cite{liu2024learn}: A plug-and-play anchor enhancement module that refines anchor quality via a learned view-graph.

    \item \texttt{MCGC}~\cite{pan2021multi}: Constructs a consensus graph by applying a graph filter regularized by a graph-level contrastive loss.

    \item \texttt{MGCCN}~\cite{liu2022multilayer}: An autoencoder using attention for layer-wise embeddings and a contrastive strategy for fusion.

    \item \texttt{LMGEC}~\cite{fettal2023simultaneous}: A simple linear model that simultaneously learns representations and clusters using a one-hop filter.

    \item \texttt{VGMGC}~\cite{chen2025variational}: Uses a variational graph generator to infer a consensus graph while modeling cross-view uncertainty.
\end{itemize}

The code for all baseline methods was obtained from the repositories provided by the original authors. Table~\ref{tab:links_code} summarizes the links to their implementations.

\subsection{Evaluation Metrics}\label{sec:eval-metrics}
The specific mathematical definitions of {\em Clustering Accuracy} (ACC), {\em Normalized Mutual Information} (NMI), {\em Adjusted Rand Index} (ARI), {\em F1-Score} (F1), and {\em Completeness Score} (CS) are as follows:
\begin{equation*}
ACC = \frac{\sum_{v_i\in V}{\mathbb{1}_{y_{i}=\textsf{map}(y^{\prime}_{i})}}}{n},
\end{equation*}
where $y^{\prime}_{i}$ and $y_{i}$ stand for the predicted and ground-truth cluster labels of node $v_i$, respectively, $\textsf{map}(y^{\prime}_{i})$ is the permutation function that maps each $y^{\prime}_{i}$ to the equivalent cluster label provided via the Hungarian algorithm \cite{kuhn1955hungarian}, and the value of $\mathbb{1}_{y_{i}=\textsf{map}(y^{\prime}_{i})}$ is 1 if $y_{i}=\textsf{map}(y^{\prime}_{i})$ and 0 otherwise,
\begin{equation*}
NMI = \frac{\sum_{i=1}^{K}\sum_{j=1}^{K}{|C^{\ast}_i\cap C_j|\cdot \log{\frac{n \cdot |C^{\ast}_i\cap C_j|}{|C^{\ast}_i|\cdot |C_j|}}}}{\sqrt{\left(\sum_{i=1}^K{|C^{\ast}_i|\log{\frac{|{C^{\ast}_i}|}{n}}}\right) \left(\sum_{j=1}^K{|C_j|\log{\frac{|{C_j}|}{n}}}\right)}},
\end{equation*}

\begin{small}
\begin{equation*}
ARI=\frac{\sum_{i=1}^K\sum_{j=1}^K{\binom {|C^{\ast}_i\cap C_j|}2}-\left(\sum_{i=1}^K{\binom {|C^{\ast}_i|}2}\cdot \sum_{j=1}^K{\binom {|C_j|} 2}\right)/{\binom {n}2}}{\frac{1}{2}\left(\sum_{i=1}^K{\binom {|C^{\ast}_i|}2}+ \sum_{j=1}^K{\binom {|C_j|}2} \right)- \left(\sum_{i=1}^K{\binom {|C^{\ast}_i|}2}\cdot \sum_{j=1}^K{\binom {|C_j|}2}\right)/{\binom {n}2} },
\end{equation*}
\end{small}
where $C^{\ast}$ represents the set of ground-truth clusters and $C$ is the set of predicted clusters. The F1-Score, which is the harmonic mean of pairwise precision and recall, is calculated based on the pairs of nodes:
\begin{equation*}
F1 = \frac{2 \sum_{i=1}^{K} \sum_{j=1}^{K} \binom{|C^{\ast}_i \cap C_j|}{2}}{\sum_{i=1}^{K} \binom{|C^{\ast}_i|}{2} + \sum_{j=1}^{K} \binom{|C_j|}{2}},
\end{equation*}
where the numerator accounts for the true positive pairs (pairs of nodes in the same cluster in both ground-truth and prediction), and the denominator is the sum of pairs in the ground-truth clusters and pairs in the predicted clusters. Finally, the Completeness Score measures if all members of a ground-truth class are assigned to the same predicted cluster, defined as $1 - H(C^{\ast}|C) / H(C^{\ast})$, which can be expanded to:
\begin{equation*}
CS = 1 - \frac{ \sum_{i=1}^{K} \sum_{j=1}^{K} \frac{|C^{\ast}_i \cap C_j|}{n} \log\left(\frac{|C^{\ast}_i \cap C_j|}{|C_j|}\right) }{ \sum_{i=1}^{K} \frac{|C^{\ast}_i|}{n} \log\left(\frac{|C^{\ast}_i|}{n}\right) },
\end{equation*}
where the numerator is the conditional entropy of the ground-truth clusters given the predicted clusters $H(C^{\ast}|C)$ and the denominator is the entropy of the ground-truth clusters $H(C^{\ast})$. For all these metrics, a higher value indicates a better clustering performance.

\subsection{Additional Parameter Analysis}\label{sec:appendix_params}
\begin{figure*}[ht]
\centering
\begin{small}
\begin{tikzpicture}
    \begin{axis}[
        hide axis,
        xmin=0, xmax=5, ymin=0, ymax=1,
        legend columns=5,
        legend style={
            at={(0.5,1.0)},
            anchor=south,
            draw=none,
            font=\small,
            /tikz/every even column/.append style={column sep=0.5cm}
        }
    ]
    \addlegendimage{line width=0.25mm,mark size=3pt,color=myblue2, mark=*}
    \addlegendentry{ACC}
    
   \addlegendimage{line width=0.25mm,mark size=2pt,color=cyan, mark=square*}
    \addlegendentry{NMI}

    \addlegendimage{line width=0.25mm,mark size=3pt,color=myorange, mark=triangle*}
    \addlegendentry{F1}

    \addlegendimage{line width=0.25mm,mark size=3pt,color=myred, mark=diamond*}
    \addlegendentry{ARI}

    \addlegendimage{line width=0.25mm,mark size=3pt,color=mygreen3, mark=pentagon*}
    \addlegendentry{CS}
    \end{axis}
\end{tikzpicture}

\begin{tikzpicture}
    \begin{axis}[
        title={Movies},
        ylabel={Metric Score ($\%$)},
        xmin=0, xmax=1,
        xtick={0,0.2,0.4,0.6,0.8,1},
        grid=major,
        width=0.26\textwidth,
        height=0.20\textwidth,
        clip=false,
    ]
    \addplot[line width=0.25mm,mark size=3pt,myblue2, mark=*] table[x=t, y expr=\thisrow{acc}*100] {
        t acc
        0   0.2536588292
        0.1 0.2565978887
        0.2 0.255878119
        0.3 0.260196737
        0.4 0.2573776392
        0.5 0.2581573896
        0.6 0.258037428
        0.7 0.2552183301
        0.8 0.2514395393
        0.9 0.2573776392
        1   0.2564779271
    };
    \addplot[line width=0.25mm,mark size=2pt,cyan, mark=square*] table[x=t, y expr=\thisrow{nmi}*100] {
        t nmi
        0   0.2226471905
        0.1 0.2225335019
        0.2 0.2239397552
        0.3 0.2245119529
        0.4 0.2235328794
        0.5 0.2230503642
        0.6 0.2235481592
        0.7 0.2215230816
        0.8 0.2203984813
        0.9 0.2243518458
        1   0.222252156
    };
    \addplot[line width=0.25mm,mark size=3pt,myorange, mark=triangle*] table[x=t, y expr=\thisrow{f1}*100] {
        t f1
        0   0.2022509733
        0.1 0.2055086682
        0.2 0.2074697476
        0.3 0.2091348041
        0.4 0.2068495909
        0.5 0.2069435936
        0.6 0.204425842
        0.7 0.2088999014
        0.8 0.2032409328
        0.9 0.2100131323
        1   0.2071900735
    };
    \addplot[line width=0.25mm,mark size=3pt,myred, mark=diamond*] table[x=t, y expr=\thisrow{ari}*100] {
        t ari
        0   0.07839145568
        0.1 0.07975190209
        0.2 0.08005812971
        0.3 0.08120155651
        0.4 0.07896138307
        0.5 0.08036940329
        0.6 0.08086783107
        0.7 0.07942161323
        0.8 0.08050580978
        0.9 0.08103169966
        1   0.08121085931
    };
    \addplot[line width=0.25mm,mark size=3pt,mygreen3, mark=pentagon*] table[x=t, y expr=\thisrow{cs}*100] {
        t cs
        0   0.2000612058
        0.1 0.1995760394
        0.2 0.2011389365
        0.3 0.2015199527
        0.4 0.2006225017
        0.5 0.200102122
        0.6 0.2010885051
        0.7 0.1987358498
        0.8 0.1976867126
        0.9 0.2014817469
        1   0.1992095039
    };
    \end{axis}
\end{tikzpicture}\hspace{5mm}
\begin{tikzpicture}
    \begin{axis}[
        title={Toys},
        xmin=0, xmax=1,
        xtick={0,0.2,0.4,0.6,0.8,1},
        grid=major,
        width=0.26\textwidth,
        height=0.20\textwidth,
        clip=false,
    ]
    \addplot[line width=0.25mm,mark size=3pt,myblue2, mark=*] table[x=t, y expr=\thisrow{acc}*100] {
        t acc
        0   0.5366996859
        0.1 0.5423532254
        0.2 0.5447692679
        0.3 0.5575259725
        0.4 0.5316743175
        0.5 0.5369412902
        0.6 0.5208987678
        0.7 0.5310461464
        0.8 0.5317709592
        0.9 0.5381493114
        1   0.5381976323
    };
    \addplot[line width=0.25mm,mark size=2pt,cyan, mark=square*] table[x=t, y expr=\thisrow{nmi}*100] {
        t nmi
        0   0.5146481006
        0.1 0.5143976605
        0.2 0.5176233807
        0.3 0.5236729838
        0.4 0.5181020459
        0.5 0.5157962055
        0.6 0.5152308699
        0.7 0.5264103487
        0.8 0.5276488984
        0.9 0.5281218101
        1   0.5188757356
    };
    \addplot[line width=0.25mm,mark size=3pt,myorange, mark=triangle*] table[x=t, y expr=\thisrow{f1}*100] {
        t f1
        0   0.4841969531
        0.1 0.4885466753
        0.2 0.5072778015
        0.3 0.5078897502
        0.4 0.4747673938
        0.5 0.4819984097
        0.6 0.4761880553
        0.7 0.4809404714
        0.8 0.4894087772
        0.9 0.4812611697
        1   0.4753122435
    };
    \addplot[line width=0.25mm,mark size=3pt,myred, mark=diamond*] table[x=t, y expr=\thisrow{ari}*100] {
        t ari
        0   0.364393756
        0.1 0.3709247809
        0.2 0.3680955735
        0.3 0.3853123111
        0.4 0.3717645932
        0.5 0.3704188977
        0.6 0.3543506929
        0.7 0.3770111
        0.8 0.3734718466
        0.9 0.3797822158
        1   0.3662875915
    };
    \addplot[line width=0.25mm,mark size=3pt,mygreen3, mark=pentagon*] table[x=t, y expr=\thisrow{cs}*100] {
        t cs
        0   0.5061683799
        0.1 0.5062635655
        0.2 0.5085038722
        0.3 0.514716908
        0.4 0.5086360023
        0.5 0.5063338906
        0.6 0.5050400781
        0.7 0.5157909633
        0.8 0.5176313878
        0.9 0.518529686
        1   0.5111117829
    };
    \end{axis}
\end{tikzpicture}\hspace{5mm}
\begin{tikzpicture}
    \begin{axis}[
        title={Grocery-S},
        xmin=0, xmax=1,
        xtick={0,0.2,0.4,0.6,0.8,1},
        grid=major,
        width=0.26\textwidth,
        height=0.20\textwidth,
        clip=false,
    ]
    \addplot[line width=0.25mm,mark size=3pt,myblue2, mark=*] table[x=t, y expr=\thisrow{acc}*100] {
        t acc
        0   0.5196204756
        0.1 0.5281129202
        0.2 0.5319784468
        0.3 0.5341454844
        0.4 0.5313927609
        0.5 0.528991449
        0.6 0.531802741
        0.7 0.5270001171
        0.8 0.5289328804
        0.9 0.507321073
        1   0.5200304557
    };
    \addplot[line width=0.25mm,mark size=2pt,cyan, mark=square*] table[x=t, y expr=\thisrow{nmi}*100] {
        t nmi
        0   0.5335901896
        0.1 0.5419121939
        0.2 0.5421037992
        0.3 0.540046226
        0.4 0.5399412349
        0.5 0.5394165328
        0.6 0.5416962663
        0.7 0.5491496084
        0.8 0.5428649272
        0.9 0.5341142224
        1   0.5332005154
    };
    \addplot[line width=0.25mm,mark size=3pt,myorange, mark=triangle*] table[x=t, y expr=\thisrow{f1}*100] {
        t f1
        0   0.4412363932
        0.1 0.4427298459
        0.2 0.4561992
        0.3 0.4541636729
        0.4 0.4447585625
        0.5 0.4535554143
        0.6 0.4530845322
        0.7 0.4545414221
        0.8 0.4462053112
        0.9 0.4208587126
        1   0.4311729557
    };
    \addplot[line width=0.25mm,mark size=3pt,myred, mark=diamond*] table[x=t, y expr=\thisrow{ari}*100] {
        t ari
        0   0.4043546176
        0.1 0.4111797339
        0.2 0.4098958421
        0.3 0.4136481461
        0.4 0.4077911592
        0.5 0.4066660481
        0.6 0.4089463132
        0.7 0.4162643774
        0.8 0.4010658324
        0.9 0.3933245242
        1   0.3911932543
    };
    \addplot[line width=0.25mm,mark size=3pt,mygreen3, mark=pentagon*] table[x=t, y expr=\thisrow{cs}*100] {
        t cs
        0   0.5180500709
        0.1 0.5269844297
        0.2 0.5261238291
        0.3 0.5239244447
        0.4 0.524232462
        0.5 0.5225780163
        0.6 0.5262163711
        0.7 0.533272203
        0.8 0.5266878648
        0.9 0.5178591576
        1   0.5168707779
    };
    \end{axis}
\end{tikzpicture}\hspace{5mm}
\begin{tikzpicture}
    \begin{axis}[
        title={Reddit-S},
        xmin=0, xmax=1,
        xtick={0,0.2,0.4,0.6,0.8,1},
        grid=major,
        width=0.26\textwidth,
        height=0.20\textwidth,
        clip=false,
    ]
    \addplot[line width=0.25mm,mark size=3pt,myblue2, mark=*] table[x=t, y expr=\thisrow{acc}*100] {
        t acc
        0   0.8916
        0.1 0.8927268152
        0.2 0.8821567887
        0.3 0.8966276582358123
        0.4 0.8662388323
        0.5 0.8805838681
        0.6 0.8525229646
        0.7 0.8524600478
        0.8 0.8080407701
        0.9 0.8631559079
        1   0.8342141689
    };
    \addplot[line width=0.25mm,mark size=2pt,cyan, mark=square*] table[x=t, y expr=\thisrow{nmi}*100] {
        t nmi
        0   0.8746
        0.1 0.8786078147
        0.2 0.8785207145
        0.3 0.8824435667422431
        0.4 0.8748601254
        0.5 0.8778350971
        0.6 0.8643331749
        0.7 0.8607301316
        0.8 0.8503859381
        0.9 0.8619395812
        1   0.8608137747
    };
    \addplot[line width=0.25mm,mark size=3pt,myorange, mark=triangle*] table[x=t, y expr=\thisrow{f1}*100] {
        t f1
        0   0.8142
        0.1 0.8153247506
        0.2 0.7736091484
        0.3 0.8197597706231031
        0.4 0.7608948481
        0.5 0.7730907117
        0.6 0.7504890121
        0.7 0.7514565306
        0.8 0.6523850447
        0.9 0.7533439242
        1   0.7038994417
    };
    \addplot[line width=0.25mm,mark size=3pt,myred, mark=diamond*] table[x=t, y expr=\thisrow{ari}*100] {
        t ari
        0   0.8771664525
        0.1 0.8699479853
        0.2 0.871815967
        0.3 0.8806704829824539
        0.4 0.858310264
        0.5 0.8713411978
        0.6 0.8463774997
        0.7 0.8425018446
        0.8 0.8087411335
        0.9 0.8475330566
        1   0.8363103747
    };
    \addplot[line width=0.25mm,mark size=3pt,mygreen3, mark=pentagon*] table[x=t, y expr=\thisrow{cs}*100] {
        t cs
        0   0.8792307336
        0.1 0.8754145111
        0.2 0.877783272
        0.3 0.8802166212082612
        0.4 0.8724544778
        0.5 0.8759132449
        0.6 0.861179002
        0.7 0.8566083351
        0.8 0.8527015164
        0.9 0.859349738
        1   0.8597672763
    };
    \end{axis}
\end{tikzpicture}\hspace{5mm}
\end{small}
\caption{Parameter analysis of $\theta$.}
\label{fig:params_theta}
\end{figure*}
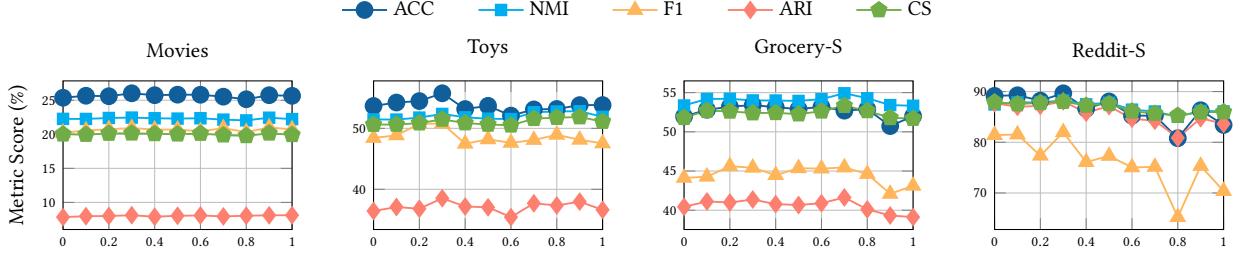

We evaluate the sensitivity to the threshold $\theta$ in our community contrastive loss. This parameter controls the fraction of samples within a pseudo-cluster selected as hard true samples for contrastive learning. A value of $\theta=0$ disables the loss, while $\theta=1$ removes the selective mechanism by treating all samples in the cluster as positives. The results, shown in Figure~\ref{fig:params_theta}, clearly demonstrate the benefit of our selective strategy. On most datasets, performance improves significantly when the loss is enabled (i.e., $\theta > 0$) and peaks around $\theta=0.3$. This suggests that selecting a smaller, confident subset of samples effectively filters noisy pairs, leading to higher-quality contrastive learning and more compact cluster representations.

\subsection{Visualization}

\begin{figure*}[ht]
    \centering
    \begin{small}
    \begin{tabularx}{\textwidth}{l *{6}{c}}
        & \texttt{S3GC} & \texttt{DMoN} & \texttt{EMVGC-LG} & \texttt{MCLGF} & \texttt{FastMICE} & \texttt{\algo{}} \\
        \rotatebox{90}{\centering Grocery-S} &
        {\centering\includegraphics[width=0.14\linewidth]{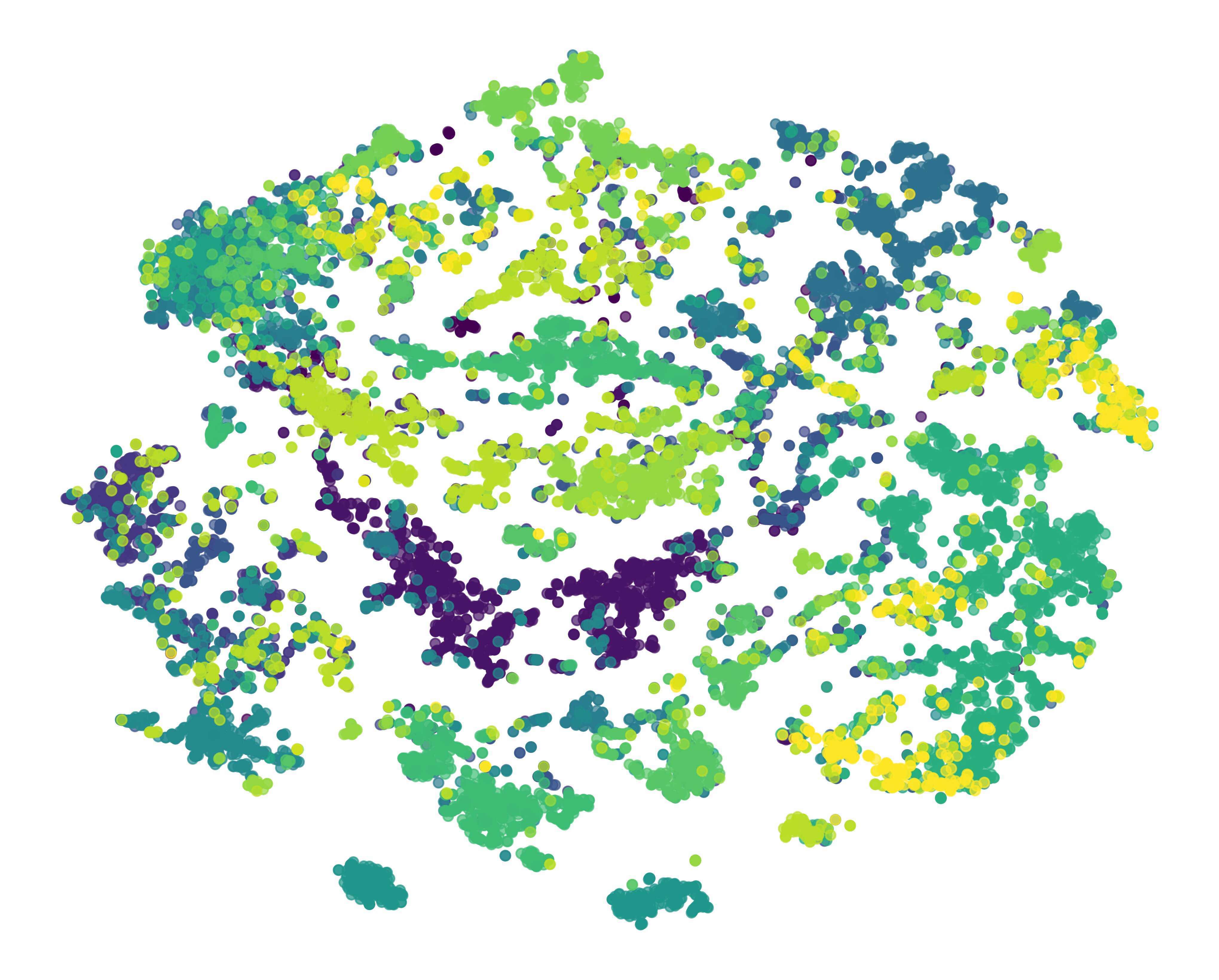}} &
        {\centering\includegraphics[width=0.14\linewidth]{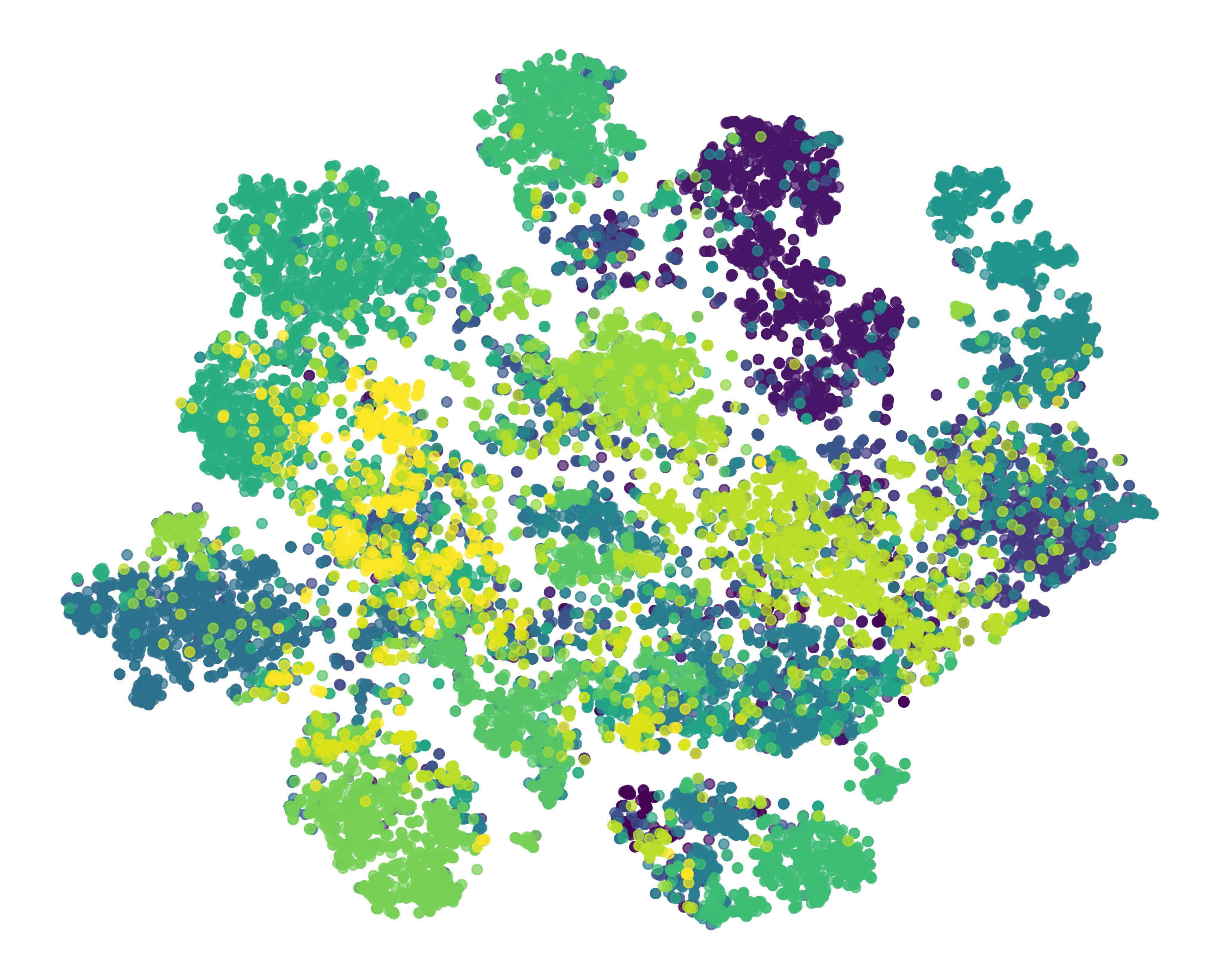}} &
        {\centering\includegraphics[width=0.14\linewidth]{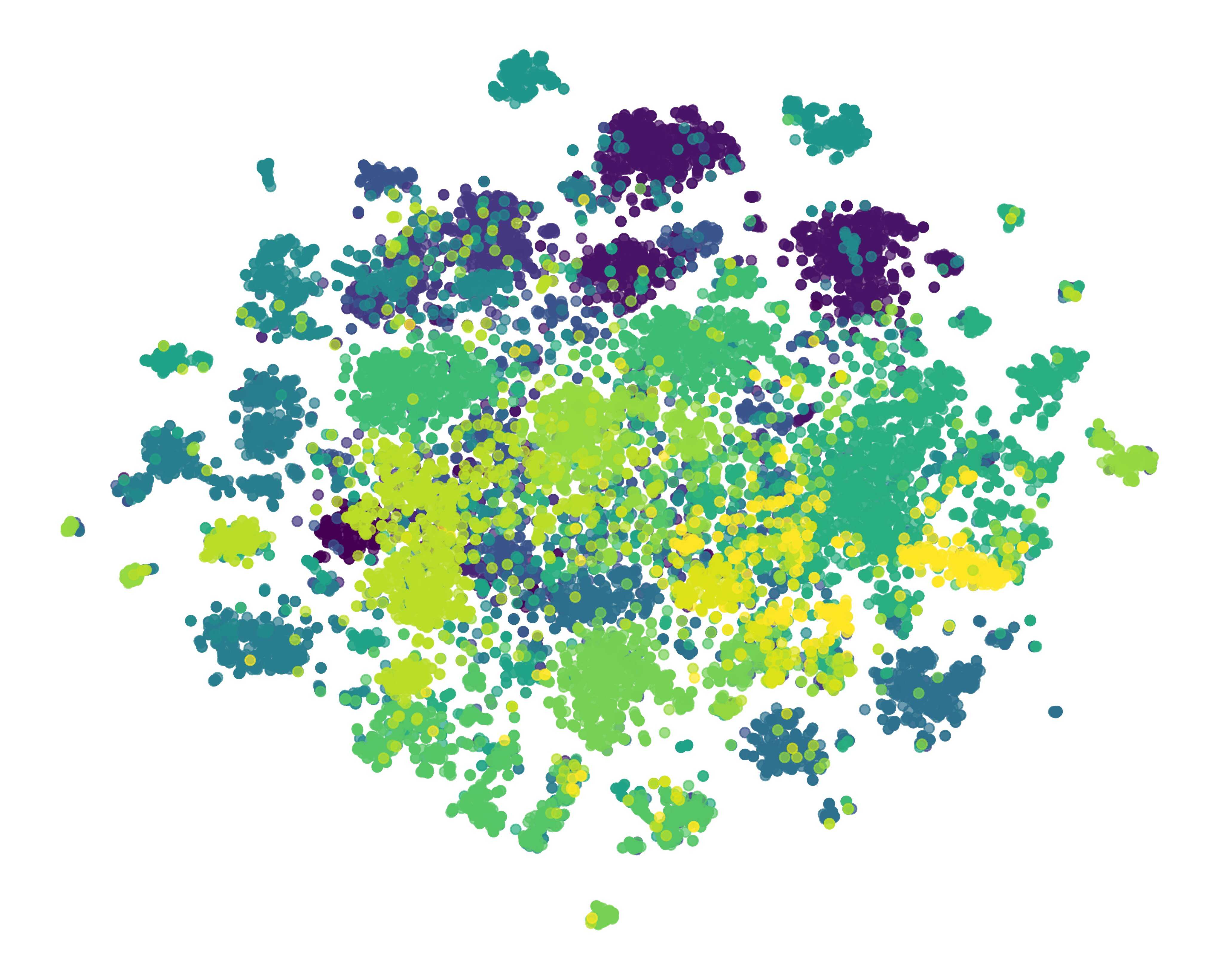}} &
        {\centering\includegraphics[width=0.14\linewidth]{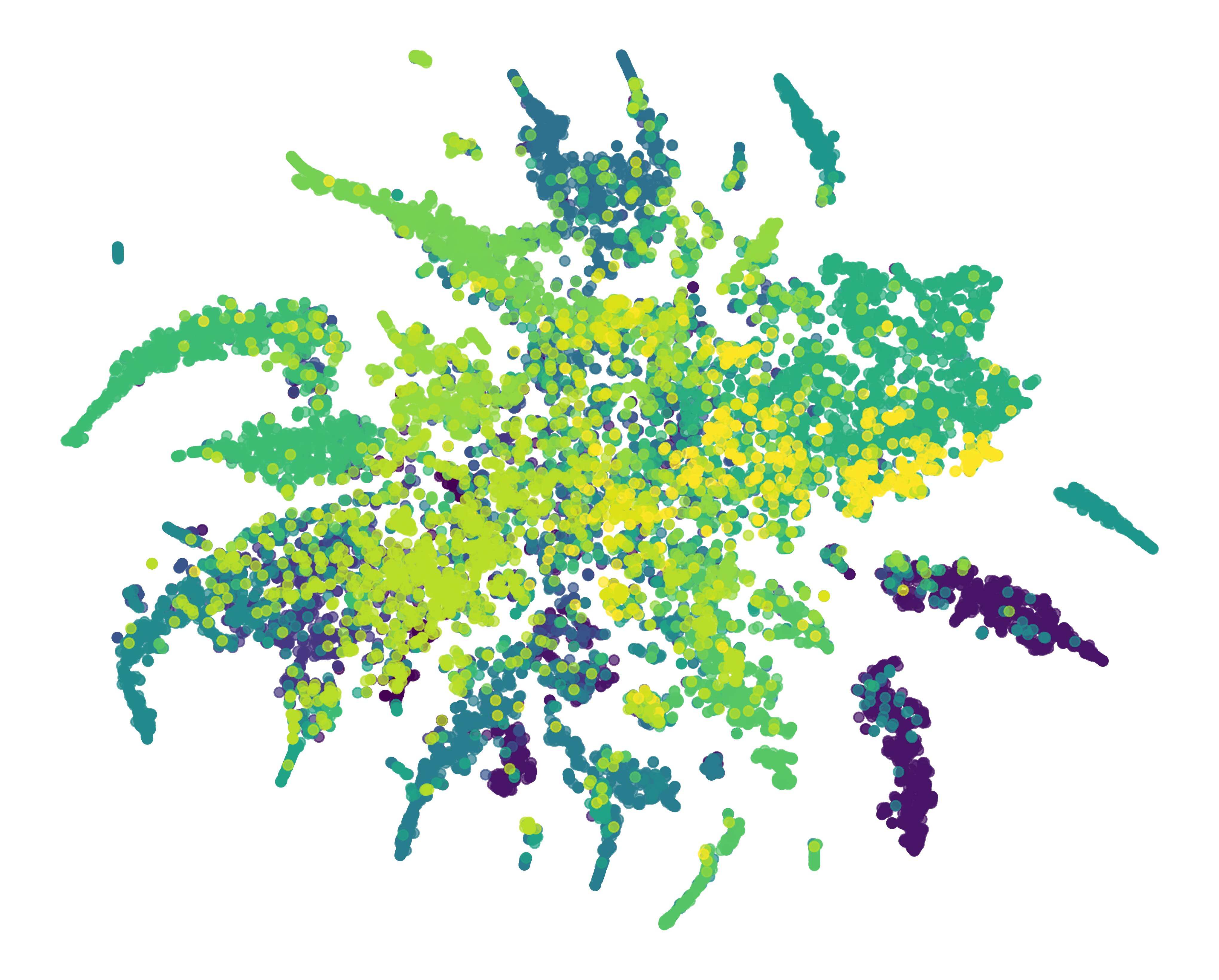}} &
        {\centering\includegraphics[width=0.14\linewidth]{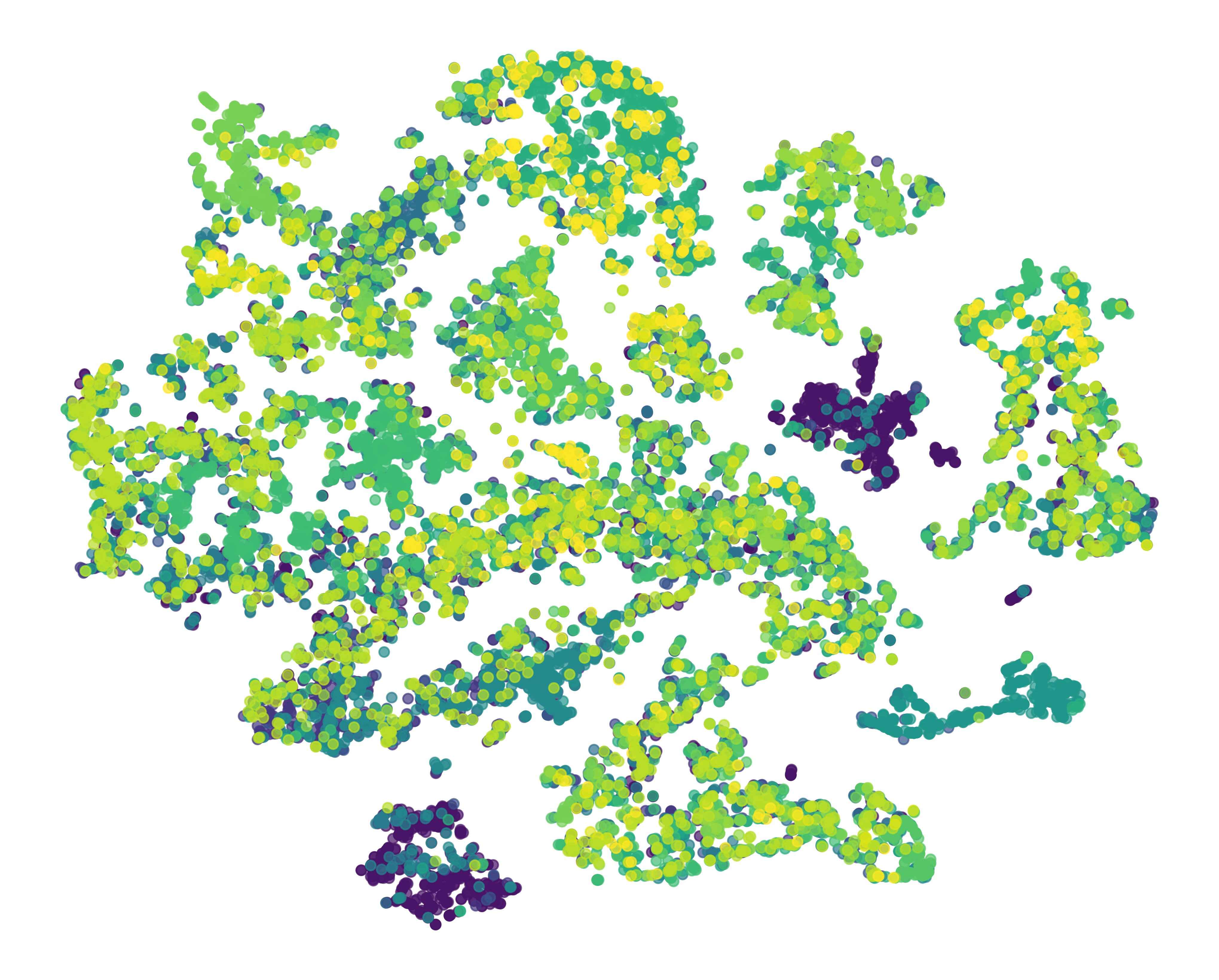}} &
        {\centering\includegraphics[width=0.14\linewidth]{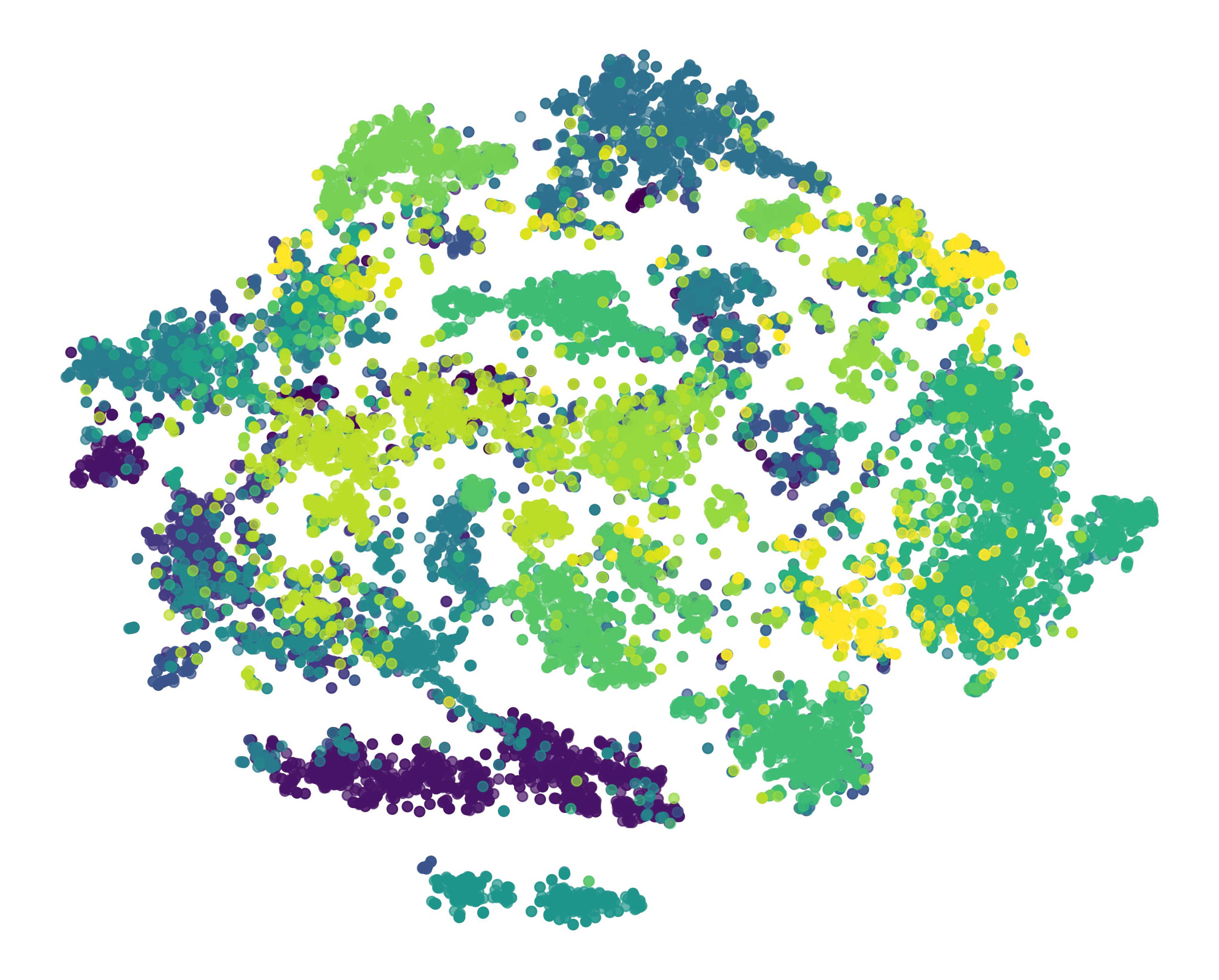}} \\
        \addlinespace[2ex]

        \rotatebox{90}{\centering Reddit-S} &
        {\centering\includegraphics[width=0.14\linewidth]{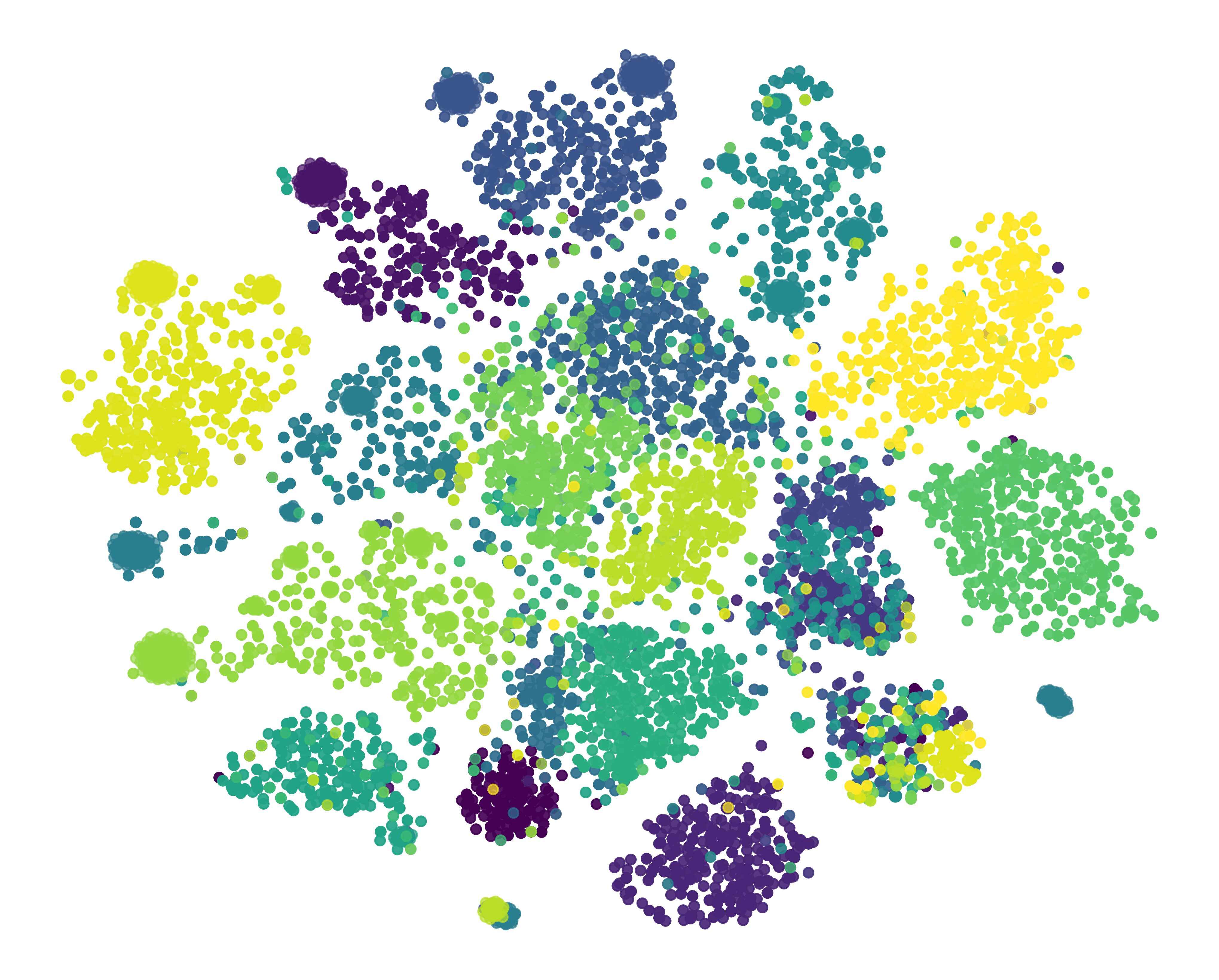}} &
        {\centering\includegraphics[width=0.14\linewidth]{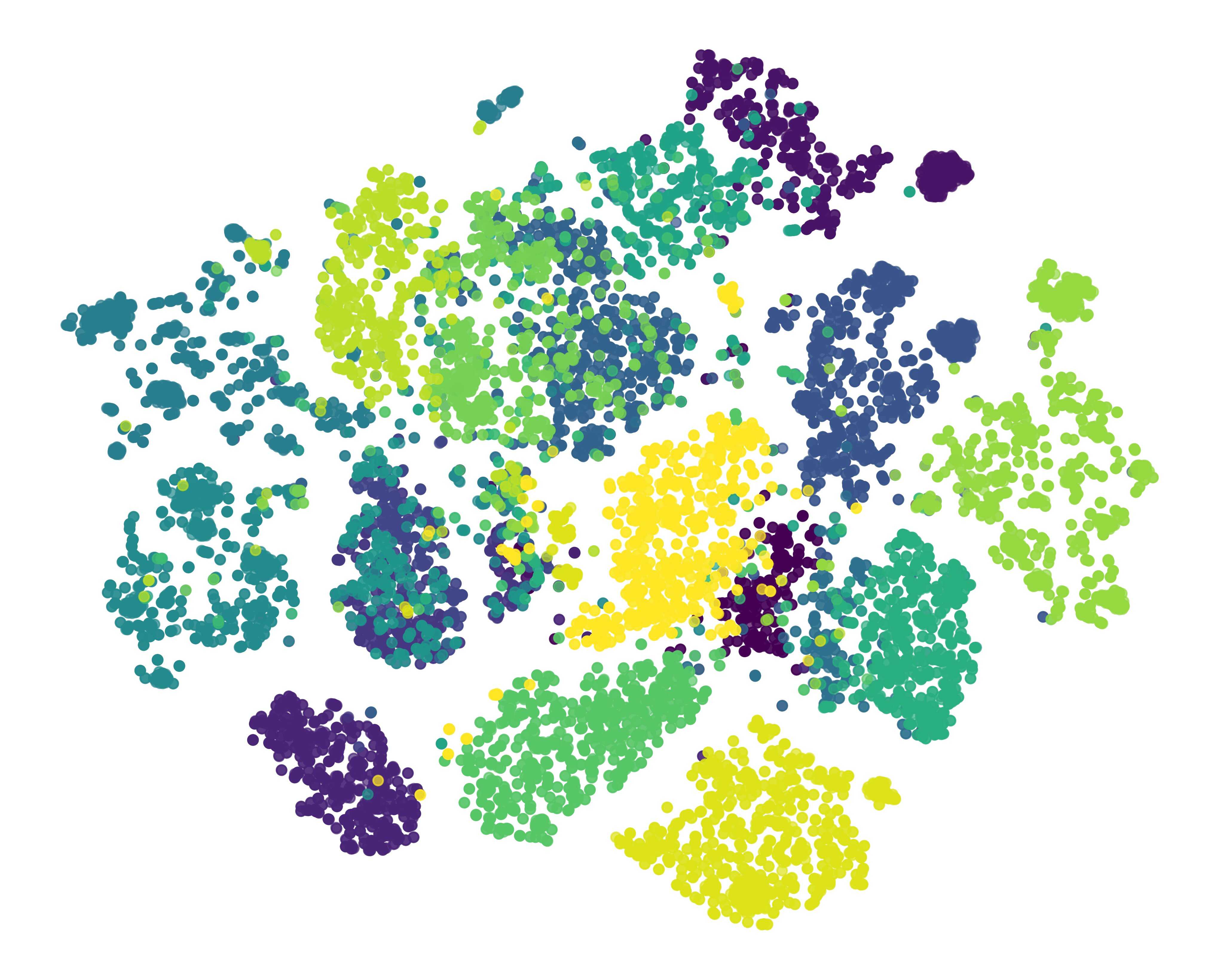}} &
        {\centering\includegraphics[width=0.14\linewidth]{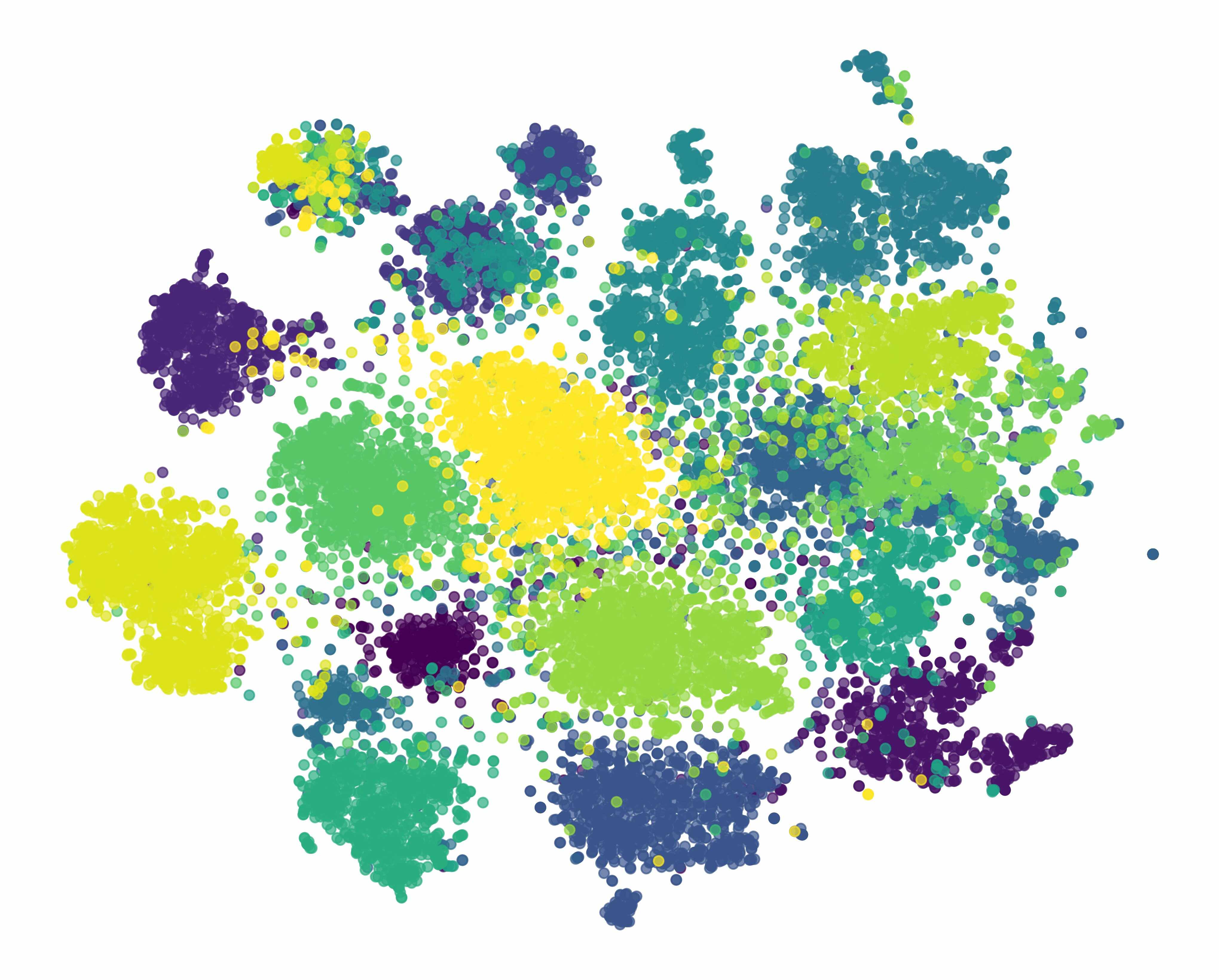}} &
        {\centering\includegraphics[width=0.14\linewidth]{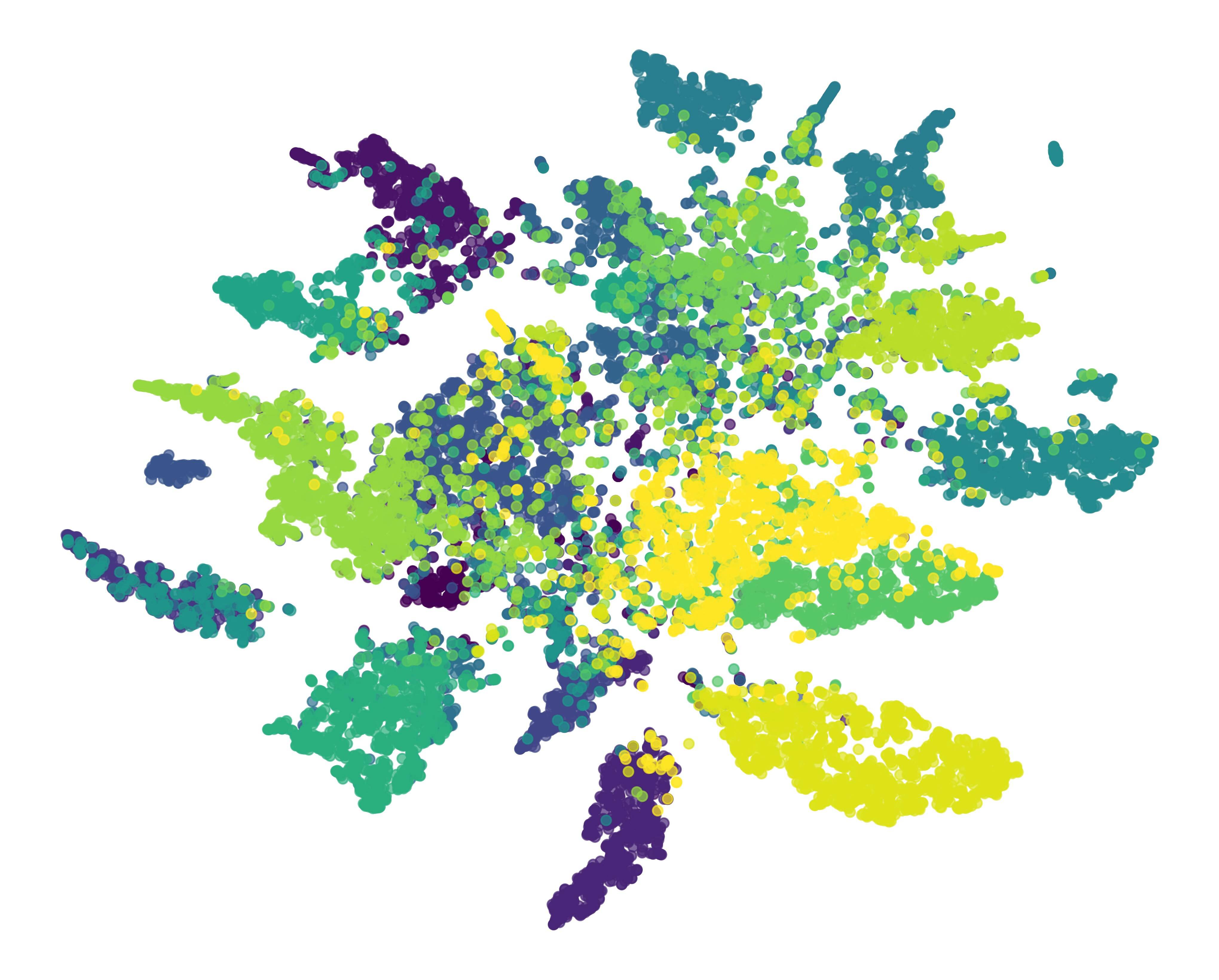}} &
        {\centering\includegraphics[width=0.14\linewidth]{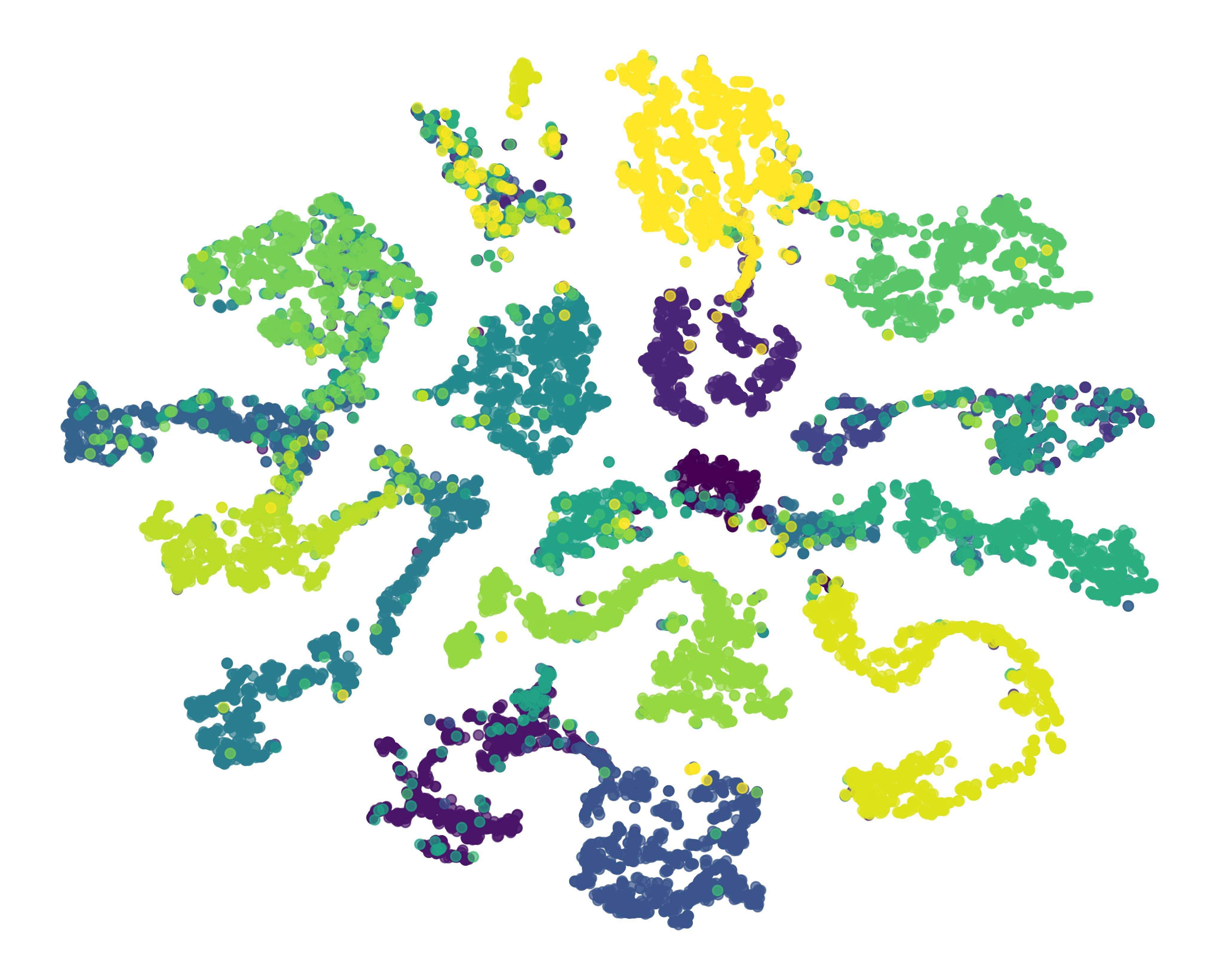}} &
        {\centering\includegraphics[width=0.14\linewidth]{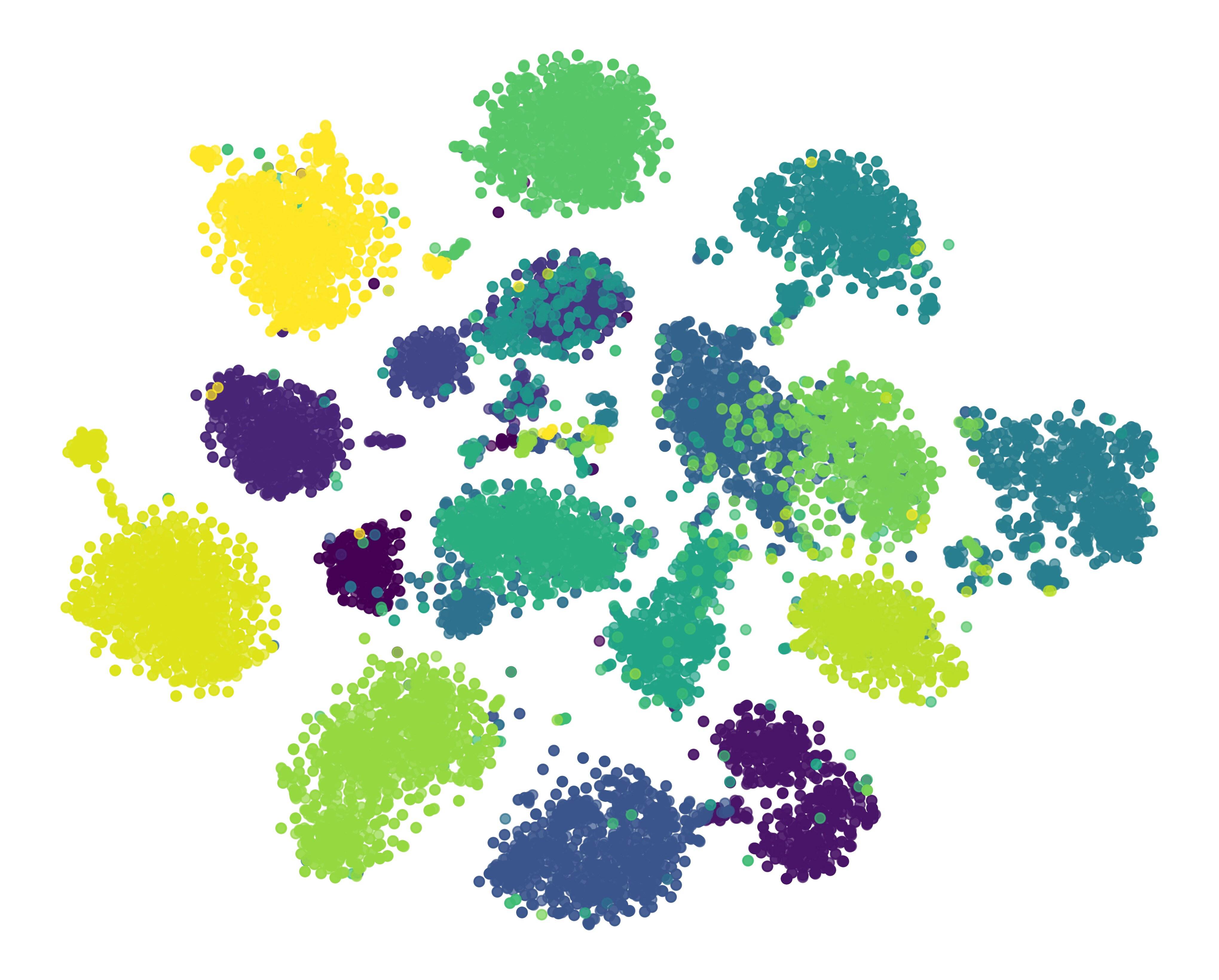}} \\

    \end{tabularx}
    \end{small}
    \caption{t-SNE visualization.}
    \label{fig:tsne}
\end{figure*}
To intuitively illustrate the effectiveness of our proposed method \algo{}, we employ t-SNE~\cite{van2008visualizing} to project the high-dimensional node embeddings learned by \algo{} and its five leading competitors, 
\texttt{S3GC}, \texttt{DMoN}, \texttt{EMVGC-LG}, \texttt{MCLGF}, and \texttt{FastMICE}, 
into a two-dimensional space. Figure~\ref{fig:tsne} presents the resulting visualizations on the \textit{Grocery-S} and \textit{Reddit-S} datasets. In these plots, each point corresponds to a node in the graph, and its color indicates its ground-truth cluster membership. As can be observed, the embeddings produced by \algo{} form more distinct and well-separated clusters. This clear separation between different ground-truth classes, compared to the more intermingled representations from the other methods, visually corroborates the superior clustering performance of our model and highlights its ability to learn more discriminative node representations.

\subsection{Hyperparameter Settings}
For each baseline method, we tune its hyperparameters by following the search strategy and space recommended in the official implementation. We report the results from the best-performing configuration to ensure a fair comparison.

For our proposed method, the final hyperparameter settings for each dataset are listed in Table~\ref{tab:hyperparams}. For all datasets, the hidden dimensionality $d$ is consistently set to 64. We employ the Adam optimizer for training, with a fixed weight decay of $1 \times 10^{-5}$. The remaining hyperparameters are optimized through grid search to determine the best configuration, with their ranges specified below:
\begin{itemize}
    \item \textbf{Learning rate:} We search within the set $\{10^{-2}, 5 \times 10^{-3}, 2 \times 10^{-3}, 10^{-3}, 5 \times 10^{-4}, 2 \times 10^{-4}, 10^{-4}, 5 \times 10^{-5}, 2 \times 10^{-5}, 10^{-5}\}$.
    \item \textbf{Number of message passing layers $T$:} The search space is $\{5, 10, 15, 20, 25, 30\}$.
    \item \textbf{Coefficients $\alpha$ and $\beta$:} Both parameters are selected from $\{0.01, 0.1, 1, 10, 100\}$.
    \item \textbf{Threshold $\theta$:} We explore values in $\{0.1, 0.2, \dots, 0.9\}$.
\end{itemize}

\begin{table}[ht]
\centering
\caption{Optimal hyperparameter settings.}
\label{tab:hyperparams}
\begin{small}
\begin{tabular}{lccccc}
\toprule
Dataset   & Learning Rate & $T$ & $\alpha$ & $\beta$ & $\theta$ \\
\midrule
\textit{Movies}    & $1 \times 10^{-3}$ & 10  & 1 & 1  & 0.3      \\
\textit{Toys}      & $1 \times 10^{-3}$ & 10  & 1 & 1  & 0.3      \\
\textit{Grocery-S} & $1 \times 10^{-3}$ & 10  & 1 & 1  & 0.3      \\
\textit{Grocery}   & $1 \times 10^{-3}$ & 10  & 1 & 1  & 0.1      \\
\textit{Reddit-S}  & $1 \times 10^{-4}$ & 10  & 1 & 1  & 0.3      \\
\textit{Reddit}    & $1 \times 10^{-2}$ & 10  & 1 & 1  & 0.1      \\
\textit{Photo}     & $1 \times 10^{-4}$ & 10  & 1 & 1  & 0.1      \\
\textit{Arts}      & $1 \times 10^{-3}$ & 10  & 1 & 1  & 0.2      \\
\bottomrule
\end{tabular}
\end{small}
\end{table}

\section{Theoretical Proofs}

\begin{proof}[\bf Proof of Lemma~\ref{lem:S-property}]
The proof consists of two parts. First, we prove that $\SM^{(i)}$ is positive semidefinite. Second, we prove that the dominant eigenvalue of $\frac{\beta}{(\beta+1)m}\sum_{i=1}^m\SM^{(i)}$ is not greater than $1$.

\stitle{$\SM^{(i)}$ is positive semidefinite}
Let us define a kernel matrix $\KM^{(i)} \in \mathbb{R}^{d\times d}$ where each element is given by the exponential dot product kernel:
\begin{equation}
    \KM^{(i)}_{j,\ell} = \exp{\left(\frac{\ZM_{\cdot,j}^{\top}\ZM_{\cdot,\ell}}{\sqrt{n}}\right)}.
\end{equation}
Since the function $k(\UM, \VM) = \exp(\frac{\UM^\top\VM}{\sqrt{n}})$ is a valid Mercer kernel~\cite{mercer1909xvi, scholkopf2002learning}, the resulting Gram matrix~\cite{gram1883ueber, hastie2009elements, bhatia2013matrix} $\KM^{(i)}$ over the set of vectors $\{\ZM_{\cdot,1}, \dots, \ZM_{\cdot,d}\}$ is positive semidefinite.

Next, let us define a diagonal matrix $\DM^{(i)} \in \mathbb{R}^{d\times d}$ whose diagonal entries are the row sums of $\KM^{(i)}$:
\begin{equation}
    \DM^{(i)}_{j,j} = \sum_{x=1}^d \KM^{(i)}_{j,x} = \sum_{x=1}^d \exp{\left(\frac{\ZM_{\cdot,j}^{\top}\ZM_{\cdot,x}}{\sqrt{n}}\right)}.
\end{equation}
Since the exponential function is always positive, all entries $\DM^{(i)}_{j,j}$ are positive, and thus $\DM^{(i)}$ is positive definite and invertible.

From the definition of $\SM^{(i)}$ in Eq.~\eqref{eq:Si}, we can express it as a symmetric normalization of $\KM^{(i)}$:
\begin{equation}
    \SM^{(i)} = (\DM^{(i)})^{-1/2} \KM^{(i)} (\DM^{(i)})^{-1/2}.
\end{equation}
This is a congruence transformation of the positive semidefinite matrix $\KM^{(i)}$. Since congruence transformations preserve positive semidefiniteness, $\SM^{(i)}$ is also positive semidefinite.

\stitle{The dominant eigenvalue is not greater than 1}
We first show that the dominant eigenvalue of each $\SM^{(i)}$ is exactly 1.
Note that since $\KM^{(i)}_{j,\ell} > 0$ and $\DM^{(i)}_{j,j} > 0$ for all $j, \ell$, every element of $\SM^{(i)}$ is strictly positive. Thus, $\SM^{(i)}$ is a positive matrix.

Let us define a vector $\VM \in \mathbb{R}^d$ with components $v_j = (\DM^{(i)}_{j,j})^{1/2} = \left(\sum_{k=1}^d \KM^{(i)}_{j,k}\right)^{1/2}$. Since all its components are positive, $\VM$ is a positive vector. Let's apply $\SM^{(i)}$ to $\VM$:
\begin{align}
    (\SM^{(i)}\VM)_j &= \sum_{\ell=1}^d \SM^{(i)}_{j,\ell} v_\ell \nonumber \\ 
    &= \sum_{\ell=1}^d \frac{\KM^{(i)}_{j,\ell}}{(\DM^{(i)}_{j,j})^{1/2} (\DM^{(i)}_{\ell,\ell})^{1/2}} \cdot (\DM^{(i)}_{\ell,\ell})^{1/2} \nonumber \\ 
    &= \frac{1}{(\DM^{(i)}_{j,j})^{1/2}} \sum_{\ell=1}^d \KM^{(i)}_{j,\ell} \nonumber \\ 
    &= \frac{\DM^{(i)}_{j,j}}{(\DM^{(i)}_{j,j})^{1/2}} = (\DM^{(i)}_{j,j})^{1/2} = v_j.
\end{align}
This shows that $\SM^{(i)}\VM = \VM$, which means that $\VM$ is an eigenvector of $\SM^{(i)}$ with a corresponding eigenvalue of 1.

According to the Perron-Frobenius theorem~\cite{Perron1907} for positive matrices, the largest eigenvalue is simple, positive, and corresponds to a strictly positive eigenvector. Since we have found a positive eigenvector $\VM$ with eigenvalue 1, this must be the dominant eigenvalue. Thus, $\lambda_{\max}(\SM^{(i)}) = 1$.

Now, let $\SM_{\text{sum}} = \frac{\beta}{(\beta+1)m}\sum_{i=1}^m\SM^{(i)}$. We want to find its dominant eigenvalue, $\lambda_{\max}(\SM_{\text{sum}})$.
\begin{equation}
    \lambda_{\max}(\SM_{\text{sum}}) = \lambda_{\max}\left(\frac{\beta}{(\beta+1)m}\sum_{i=1}^m\SM^{(i)}\right) = \frac{\beta}{(\beta+1)m} \lambda_{\max}\left(\sum_{i=1}^m\SM^{(i)}\right).
\end{equation}
By Weyl's inequality~\cite{weyl1912asymptotische, franklin2000matrix} for the eigenvalues of a sum of symmetric matrices, we have:
\begin{equation}
    \lambda_{\max}\left(\sum_{i=1}^m\SM^{(i)}\right) \le \sum_{i=1}^m \lambda_{\max}(\SM^{(i)}).
\end{equation}
Substituting this into our equation for $\lambda_{\max}(\SM_{\text{sum}})$:
\begin{equation}
    \lambda_{\max}(\SM_{\text{sum}}) \le \frac{\beta}{(\beta+1)m} \sum_{i=1}^m \lambda_{\max}(\SM^{(i)}).
\end{equation}
Since we have shown that $\lambda_{\max}(\SM^{(i)}) = 1$ for all $i$:
\begin{equation}
    \lambda_{\max}(\SM_{\text{sum}}) \le \frac{\beta}{(\beta+1)m} \sum_{i=1}^m 1 = \frac{\beta}{(\beta+1)m} \cdot m = \frac{\beta}{\beta+1}.
\end{equation}
Given that $\beta \ge 0$, we have $0 \le \frac{\beta}{\beta+1} < 1$. Therefore, the dominant eigenvalue of $\frac{\beta}{(\beta+1)m}\sum_{i=1}^m\SM^{(i)}$ is not greater than 1. This completes the proof.
\end{proof}

\begin{proof}[\bf Proof of Lemma~\ref{lem:dual-filtering}]
The optimization objective is given by:
\begin{equation}
\min_{\HM}{\mathcal{O}(\HM) = \|\HM - \ZM\|_F^2 + \alpha\Tr(\HM^{\top}\LM\HM) + \frac{\beta}{m}\sum_{i=1}^{m}{\Tr(\HM(\IM-\SM^{(i)})\HM^{\top})}}
\end{equation}
To find the minimum, we compute the gradient of $\mathcal{O}(\HM)$ with respect to $\HM$ and set it to zero. Using standard matrix calculus identities, we get:
\begin{equation}
\frac{\partial \mathcal{O}(\HM)}{\partial \HM} = 2(\HM - \ZM) + 2\alpha\LM\HM + 2\HM \left( \frac{\beta}{m}\sum_{i=1}^{m}{(\IM-\SM^{(i)})} \right) = 0.
\end{equation}
This Sylvester equation can be solved approximately by considering the proximal operators for the two regularization terms sequentially, which is a common approach in such filtering frameworks. First, we solve for the feature-side smoothing, and then for the graph-side smoothing.

\stitle{Feature-domain filtering}
We solve the sub-problem involving the feature Laplacians:
\begin{equation}
\min_{\HM'} {\|\ZM - \HM'\|_F^2 + \frac{\beta}{m}\sum_{i=1}^{m}{\Tr(\HM'(\IM-\SM^{(i)})(\HM')^{\top})}}.
\end{equation}
Setting the gradient with respect to $\HM'$ to zero gives:
$2(\HM' - \ZM) + 2\HM' \left( \frac{\beta}{m}\sum_{i=1}^{m}{(\IM-\SM^{(i)})} \right) = 0$, which can be simplified as $$\HM' \left( \IM + \frac{\beta}{m}\sum_{i=1}^{m}{(\IM-\SM^{(i)})} \right) = \ZM.$$
The intermediate solution is $\HM' = \ZM \left( \IM + \frac{\beta}{m}\sum_{i=1}^{m}{(\IM-\SM^{(i)})} \right)^{-1}$.

\stitle{Graph-domain filtering}
We use the result $\HM'$ as the input to the graph-based regularization sub-problem:
\begin{equation}
\min_{\HM} {\|\HM' - \HM\|_F^2 + \alpha\Tr(\HM^{\top}\LM\HM)}.
\end{equation}
Setting the gradient with respect to $\HM$ to zero gives:
$2(\HM - \HM') + 2\alpha\LM\HM = 0$, which simplifies to $(\IM + \alpha\LM)\HM = \HM'$.
The solution is $\HM = (\IM + \alpha\LM)^{-1} \HM'$.

\stitle{Combine and simplify}
By substituting the expression for $\HM'$ into the second solution, we obtain the combined closed-form solution:
\begin{equation}\label{eq:H_intermediate}
\HM = (\IM + \alpha\LM)^{-1} \ZM \left( \IM + \frac{\beta}{m}\sum_{i=1}^{m}{(\IM-\SM^{(i)})} \right)^{-1}.
\end{equation}
With the definition of the graph Laplacian $\LM = \IM - \NAM$:
\begin{align*}
\IM + \alpha\LM &= \IM + \alpha(\IM - \NAM) = (\alpha+1)\IM - \alpha\NAM. \\
\IM + \frac{\beta}{m}\sum_{i=1}^{m}{(\IM-\SM^{(i)})} &= \IM + \frac{\beta}{m}\sum_{i=1}^{m}{(\IM - \SM^{(i)})}\\
& = \IM + \frac{\beta}{m}(m\IM) - \frac{\beta}{m}\sum_{i=1}^{m}{\SM^{(i)}} \\
&= (\beta+1)\IM - \frac{\beta}{m}\sum_{i=1}^{m}{\SM^{(i)}}.
\end{align*}
Substituting these back into the solution for $\HM$ yields the expression from Eq.~\eqref{eq:H_intermediate}:
\begin{equation}
\HM = \left( (\alpha+1)\IM - \alpha\NAM \right)^{-1}\ZM \left( (\beta+1)\IM - \frac{\beta}{m}\sum_{i=1}^m\SM^{(i)} \right)^{-1}.
\end{equation}
\stitle{Neumann series expansion}
To reveal the filtering mechanism, we apply the Neumann series expansion in Theorem~\ref{lem:neu}, $(\IM - \NAM)^{-1} = \sum_{t=0}^\infty \NAM^t$, given that the dominant eigenvalue of $\NAM$ is not greater 1~\cite{chung1997spectral}. Thus, we rewrite the inverse terms:
\begin{align*}
\left( (\alpha+1)\IM - \alpha\NAM \right)^{-1} &= \left( (\alpha+1)\left(\IM - \frac{\alpha}{\alpha+1}\NAM\right) \right)^{-1} \\
&= \frac{1}{\alpha+1}\sum_{t=0}^\infty{\left(\frac{\alpha}{\alpha+1}\NAM\right)^t}.
\end{align*}
Similarly, with Lemma~\ref{lem:S-property}, we have
\begin{align*}
\left( (\beta+1)\IM - \frac{\beta}{m}\sum_{i=1}^m\SM^{(i)} \right)^{-1} &= \frac{1}{\beta+1}\left( \IM - \frac{\beta}{(\beta+1)m}\sum_{i=1}^m\SM^{(i)} \right)^{-1} \\
&= \frac{1}{\beta+1}\sum_{t=0}^\infty{\left(\frac{\beta}{(\beta+1)m}\sum_{i=1}^m{\SM^{(i)}}\right)^t}.
\end{align*}
Combining these two series expansions gives the final form of the solution:
\begin{equation*}
\HM = \frac{1}{(\alpha+1)(\beta+1)}\sum_{t=0}^\infty{\left(\frac{\alpha}{\alpha+1}\NAM\right)^t}{\ZM}\cdot \sum_{t=0}^\infty{\left(\frac{\beta}{(\beta+1)m}\sum_{i=1}^m{\SM^{(i)}}\right)^t}.
\end{equation*}
This completes the proof.
\end{proof}

\begin{proof}[\bf Proof of Theorem~\ref{lem:dual-low-pass}]
The solution to our optimization objective is given by:
\begin{equation}
\HM = (\IM + \alpha\LM)^{-1} \ZM \left( \IM + \frac{\beta}{m}\sum_{i=1}^{m}{(\IM-\SM^{(i)})} \right)^{-1}
\end{equation}
Let's analyze the filtering effect of the left and right matrix inverse terms separately.

\stitle{Node-Domain Filtering (Left-hand side)}
The node-domain filter is represented by the operator $\FM_{\text{node}} = (\IM + \alpha\LM)^{-1}$. Let the eigendecomposition of the graph Laplacian $\LM$ be $\LM = \VM\mathbf{\Lambda}\VM^{\top}$, where $\VM$ is the matrix of orthonormal eigenvectors and $\mathbf{\Lambda}$ is the diagonal matrix of corresponding non-negative eigenvalues, $0 = \lambda_1 \le \lambda_2 \le \dots \le \lambda_n$. These eigenvalues represent the frequencies of the graph signal.

We can express the filter operator in terms of this decomposition:
\begin{align*}
\FM_{\text{node}} &= (\VM\VM^{\top} + \alpha\VM\mathbf{\Lambda}\VM^{\top})^{-1} \\
&= (\VM(\IM + \alpha\mathbf{\Lambda})\VM^{\top})^{-1} \\
&= \VM(\IM + \alpha\mathbf{\Lambda})^{-1}\VM^{\top}
\end{align*}
The matrix $(\IM + \alpha\mathbf{\Lambda})^{-1}$ is a diagonal matrix, whose $k$-th diagonal entry is $h(\lambda_k) = \frac{1}{1 + \alpha\lambda_k}$. This function $h(\lambda_k)$ is the frequency response of the node-domain filter. Since $\alpha > 0$ and $\lambda_k \ge 0$, $h(\lambda_k)$ is a monotonically decreasing function of $\lambda_k$. It preserves the lowest frequency component ($\lambda_k=0$) with a gain of 1, while attenuating higher frequency components (where $\lambda_k$ is large). This is the characteristic behavior of a low-pass filter.

\stitle{Feature-Domain Filtering (Right-hand side)}
Similarly, let's define a consensus feature Laplacian $\bar{\LM}_{\text{feat}} = \frac{1}{m}\sum_{i=1}^{m}{(\IM-\SM^{(i)})}$. The feature-domain filter is $\FM_{\text{feat}} = (\IM + \beta\bar{\LM}_{\text{feat}})^{-1}$. Let the eigendecomposition of $\bar{\LM}_{\text{feat}}$ be $\bar{\LM}_{\text{feat}} = \UM\mathbf{\Omega}\UM^{\top}$, where $\UM$ is the matrix of eigenvectors and $\mathbf{\Omega}$ is the diagonal matrix of eigenvalues $\omega_j \ge 0$.

Following the same logic, we can write:
\begin{equation}
\FM_{\text{feat}} = \UM(\IM + \beta\mathbf{\Omega})^{-1}\UM^{\top}
\end{equation}
The frequency response of the feature-domain filter is given by the diagonal elements of $(\IM + \beta\mathbf{\Omega})^{-1}$, which are $g(\omega_j) = \frac{1}{1 + \beta\omega_j}$. This is also a low-pass filter, attenuating high-frequency components within the feature dimensions.

\stitle{Conclusion}
The complete operation $\HM = \FM_{\text{node}} \ZM \FM_{\text{feat}}$ can be seen as passing the input signal $\ZM$ through two distinct low-pass filters. The left-multiplication by $\FM_{\text{node}}$ smooths the representations across the nodes of the graph, removing high-frequency noise between neighboring nodes. The right-multiplication by $\FM_{\text{feat}}$ smooths the representations across the feature dimensions, removing noise by enforcing consistency based on the consensus feature similarity. Therefore, the method effectively denoises the embeddings by suppressing high-frequency components in both the node and feature domains.
\end{proof}

\begin{proof}[\bf Proof of Theorem~\ref{lem:approxi}]
The $T$-term truncated operator from the Neumann series expansion is $\FM^{(T)} = \frac{1}{\alpha+1}\sum_{t=0}^T{(\frac{\alpha}{\alpha+1}\NAM)^t}$. Its spectral response $h^{(T)}(\lambda_k)$ is given by applying it to the corresponding eigenvalue of $\NAM$, which is $\mu_k = 1-\lambda_k$. This yields a geometric series:
\begin{align*}
h^{(T)}(\lambda_k) & = \frac{1}{\alpha+1}\sum_{t=0}^T \left(\frac{\alpha(1-\lambda_k)}{\alpha+1}\right)^t \\
& = \frac{1}{\alpha+1} \frac{1 - \left(\frac{\alpha(1-\lambda_k)}{\alpha+1}\right)^{T+1}}{1 - \frac{\alpha(1-\lambda_k)}{\alpha+1}}.
\end{align*}
Simplifying the denominator gives:
\begin{equation}
1 - \frac{\alpha(1-\lambda_k)}{\alpha+1} = \frac{\alpha+1 - \alpha + \alpha\lambda_k}{\alpha+1} = \frac{1+\alpha\lambda_k}{\alpha+1}.
\end{equation}
Substituting this back, we obtain:
\begin{align*}
h^{(T)}(\lambda_k) &= \frac{1}{\alpha+1} \frac{1 - \left(\frac{\alpha(1-\lambda_k)}{\alpha+1}\right)^{T+1}}{\frac{1+\alpha\lambda_k}{\alpha+1}}\\
& = \frac{1}{1+\alpha\lambda_k} \left(1 - \left(\frac{\alpha(1-\lambda_k)}{\alpha+1}\right)^{T+1}\right).
\end{align*}
Recognizing that the ideal filter response is $h(\lambda_k) = (1+\alpha\lambda_k)^{-1}$,
the approximation error is given by:
\begin{align*}
|h^{(T)}(\lambda_k) - h(\lambda_k)| & = \left| -h(\lambda_k) \left(\frac{\alpha(1-\lambda_k)}{\alpha+1}\right)^{T+1} \right|\\
&= |h(\lambda_k)| \left|\frac{\alpha(1-\lambda_k)}{\alpha+1}\right|^{T+1}.
\end{align*}
Since $\alpha \ge 0$ and for the normalized Laplacian $\lambda_k \in [0, 2]$, the base of the exponent $\left|\frac{\alpha(1-\lambda_k)}{\alpha+1}\right|$ is strictly less than 1 for all non-trivial converging cases (if $\alpha=0$, the error is trivially 0). Thus, the error term decreases exponentially as $T \to \infty$.
\end{proof}

\section{Theoretical Discussions of Eq.~\eqref{eq:obj}}
\label{sec:additional_theoretical_analyses}
Building on the analyses in \cite{shahid2016low}, we provide further theoretical discussion of the dual graph filtering framework. To facilitate the analysis, we first define the spectral decompositions of the key operators and the learned representation. These decompositions form the basis for interpreting our method from four complementary perspectives.

Let the eigendecomposition of the node graph Laplacian be $\LM = \VM\mathbf{\Lambda}\VM^{\top}$, where $\VM$ is the matrix of eigenvectors and $\mathbf{\Lambda}$ is the diagonal matrix of corresponding eigenvalues. For the feature domain, we define the average feature Laplacian as $\bar{\LM}_{\text{feat}} = \frac{1}{m}\sum_{i=1}^{m}{(\IM-\SM^{(i)})}$, with its eigendecomposition given by $\bar{\LM}_{\text{feat}} = \UM\mathbf{\Omega}\UM^{\top}$. Finally, we denote the Singular Value Decomposition (SVD) of our learned representation matrix as $\HM = \PM \mathbf{\Sigma} \QM^{\top}$.

\stitle{Singular Value Penalization}
The closed-form solution for $\HM$, derived from Theorem~\ref{lem:dual-low-pass}, provides a clear interpretation of the filtering process. The solution is given by:
$$
\HM = \left[\VM(\IM+\alpha\mathbf{\Lambda})^{-1}\VM^{\top}\right] \ZM \left[\UM(\IM+\beta\mathbf{\Omega})^{-1}\UM^{\top}\right].
$$
This expression shows that the node-domain filter and feature-domain filter act on the left and right singular subspaces of the initial features $\ZM$, respectively. The frequency response functions, $h(\lambda_k)=(1+\alpha\lambda_k)^{-1}$ and $g(\omega_j)=(1+\beta\omega_j)^{-1}$, effectively shrink the singular values of $\ZM$ associated with high-frequency variations (i.e., large eigenvalues $\lambda_k$ or $\omega_j$) in either domain. This directly penalizes noisy spectral components to achieve a denoised representation.

\stitle{Penalization in the Spectral Graph Fourier Domain}
An alternative viewpoint comes from analyzing the regularization terms in the spectral domain. By leveraging the properties of the trace and the graph Laplacians, these terms can be expressed as:
$$
\alpha\cdot\Tr(\HM^{\top}\LM\HM) + \beta\cdot\Tr(\HM\bar{\LM}_{\text{feat}}\HM^{\top}) = \alpha \|\mathbf{\Lambda}^{\frac{1}{2}} \VM^\top\HM\|_F^2 + \beta {\|\mathbf{\Omega}^{\frac{1}{2}} (\HM\UM)^\top\|_F^2}.
$$
This formulation reveals that our optimization objective directly penalizes the energy of the representation $\HM$ in the graph Fourier domains. The penalty is weighted by the eigenvalues, which correspond to frequencies. Consequently, the optimization encourages the signal energy to concentrate in the lowest frequencies (i.e., smallest eigenvalues) of both the node and feature Laplacians, enforcing smoothness and structural coherence.

\stitle{Weighted Subspace Alignment}
A deeper analysis is possible by substituting the SVD of $\HM$ into the regularization terms. This recasts the objective as a weighted sum over the singular values of $\HM$ and their alignment with the graph eigenspaces:
$$
\sum_{k}^{\min \left\{n, d\right\}}\sigma_{k}^{2}\left(\alpha \sum_{j=1}^{n} \lambda_{j} (\mathbf{p}_{k}^{\top}\mathbf{v}_{j})^{2} + \beta \sum_{j=1}^{d} \omega_{j} (\mathbf{q}_{k}^{\top}\mathbf{u}_{j})^{2}\right),
$$
where $\sigma_k$ is the $k$-th singular value of $\HM$; $\mathbf{p}_k$ and $\mathbf{q}_k$ are the corresponding left and right singular vectors; and $\{\lambda_{j}, \mathbf{v}_{j}\}$ and $\{\omega_{j}, \mathbf{u}_{j}\}$ are the eigenvalue-eigenvector pairs for $\LM$ and $\bar{\LM}_{\text{feat}}$, respectively. This interpretation shows that minimizing the objective has two simultaneous effects: (1) it penalizes large singular values $\sigma_k$, and (2) it forces the left singular vectors $\mathbf{p}_k$ of $\HM$ to align with the low-frequency eigenvectors $\mathbf{v}_{j}$ of the node graph, and similarly aligns the right singular vectors $\mathbf{q}_k$ with the low-frequency eigenvectors $\mathbf{u}_{j}$ of the feature graph.

\stitle{Subspace Rotation}
Finally, the optimization in Eq.~\eqref{eq:obj} can be viewed as finding optimal subspaces spanned by the columns of $\PM$ and $\QM$ that minimize an alignment cost with the eigenspaces of the graph operators. Given a fixed singular value matrix $\mathbf{\Sigma}$, the problem becomes:
$$
(\PM^*, \QM^*) = \arg\min_{\PM, \QM} \left( \alpha \Tr(\mathbf{\Sigma} \PM^\top \LM \PM \mathbf{\Sigma}) + \beta \Tr(\mathbf{\Sigma} \QM^\top \bar{\LM}_{\text{feat}} \QM \mathbf{\Sigma}) \right).
$$
This perspective frames our method as a subspace rotation problem. The optimization process effectively rotates the principal left and right singular subspaces of the initial features $\ZM$ to find new subspaces for $\HM$. These new subspaces are chosen to be more coherent with the intrinsic low-frequency structures of the node and feature graphs, thereby filtering out noise that is inconsistent with the data's underlying structure.

\end{document}